\documentclass[letterpaper]{article} 
\usepackage{aaai2026}  
\usepackage{times}  
\usepackage{helvet}  
\usepackage{courier}  
\usepackage[hyphens]{url}  
\usepackage{graphicx} 
\usepackage{natbib}  
\usepackage{caption} 
\usepackage{algorithm}
\usepackage{algorithmic}
\usepackage{array}
\usepackage{graphicx}
\usepackage{makecell}
\usepackage{amssymb}
\usepackage{amsmath}

\usepackage[table]{xcolor}

\usepackage{booktabs}

\usepackage{multirow}

\usepackage{xcolor}

\usepackage{newfloat}
\usepackage{listings}
\DeclareCaptionStyle{ruled}{labelfont=normalfont,labelsep=colon,strut=off} 
\floatstyle{ruled}
\newfloat{listing}{tb}{lst}{}
\floatname{listing}{Listing}
\title{DTTNet: Improving Video Shadow Detection via Dark-Aware Guidance and Tokenized Temporal Modeling}
\author{
    Zhicheng Li\textsuperscript{\rm 1,}\textsuperscript{\rm 2}\equalcontrib, 
    Kunyang Sun\textsuperscript{\rm 1,}\textsuperscript{\rm 2}\equalcontrib, 
    Rui Yao\textsuperscript{\rm 1,}\textsuperscript{\rm 2}\thanks{Corresponding authors, ruiyao@cumt.edu.cn.}, 
    Hancheng Zhu\textsuperscript{\rm 1,}\textsuperscript{\rm 2}, 
    Fuyuan Hu\textsuperscript{\rm 3},
    Jiaqi Zhao\textsuperscript{\rm 1,}\textsuperscript{\rm 2}, 
    Zhiwen Shao\textsuperscript{\rm 1,}\textsuperscript{\rm 2}, 
    Yong Zhou\textsuperscript{\rm 1,}\textsuperscript{\rm 2} 
}
\affiliations{
    \textsuperscript{\rm 1}School of Computer Science and Technology / School of Artificial Intelligence, China University of Mining and Technology\\
    \textsuperscript{\rm 2}Mine Digitization Engineering Research Center of the Ministry of Education, China\\
    \textsuperscript{\rm 3}School of Electronic and Information Engineering, Suzhou University of Science and Technology, China\\

}

\usepackage{bibentry}

\begin{document}

\maketitle

\begin{abstract}
Video shadow detection confronts two entwined difficulties: distinguishing shadows from complex backgrounds and modeling dynamic shadow deformations under varying illumination. To address shadow-background ambiguity, we leverage linguistic priors through the proposed Vision-language Match Module (VMM) and a Dark-aware Semantic Block (DSB), extracting text-guided features to explicitly differentiate shadows from dark objects. Furthermore, we introduce adaptive mask reweighting to downweight penumbra regions during training and apply edge masks at the final decoder stage for better supervision. For temporal modeling of variable shadow shapes, we propose a Tokenized Temporal Block (TTB) that decouples spatiotemporal learning. TTB summarizes cross-frame shadow semantics into learnable temporal tokens, enabling efficient sequence encoding with minimal computation overhead. Comprehensive Experiments on multiple benchmark datasets demonstrate state-of-the-art accuracy and real-time inference efficiency. Codes are available at https://github.com/city-cheng/DTTNet.

\end{abstract}

\section{Introduction}
Shadows, ubiquitous in natural imagery and video sequences, offer critical cues for vision tasks including light source conditions~\cite{lalonde2014lighting}, object shapes~\cite{karsch2011rendering,shao2011shadow}, and depth relationships ~\cite{adams2022depth}. 
Conversely, ambiguous shadows can degrade performance in object detection~\cite{cucchiara2003detecting,hu2021revisiting}, illumination estimation~\cite{adams2021shedding}, and visual tracking~\cite{chen2021triple}. While image shadow detection focuses on spatial separation under complex lighting, video shadow detection additionally requires modeling temporal shadow deformations induced by dynamic illumination. This necessity for spatiotemporal dual modeling renders video shadow detection a substantially more challenging problem.

Recent advances in deep learning have significantly propelled video shadow detection. Unlike video object segmentation where semantics vary substantially, shadows maintain consistent characteristics—persistently occupying darker regions. Hence, one promising direction exploits inter-frame affinity to aggregate locally similar shadow features across video sequences. For instance, previous methods~\cite{chen2021triple,liu2023scotch} utilize intra-frame attention to aggregate temporally consistent features. However, frame-level feature fusion learns from predefined frame pairs, which not only produces redundant representations but also leads to fragmented temporal modeling and compromised representation coherence. To efficiently encoding temporal information, we are motivated to optimize learnable tokens to capture shadow dynamics and subsequently inject them into spatial features. This token-level fusion not only enhances the aggregation of multi-frame information but also effectively reduces computational overhead and parameter requirements. Consequently, we propose a Tokenized Temporal Block (TTB) that first encodes temporal contexts into compact tokens and then selectively transfers this knowledge to spatial pixels via a spatial matching mechanism.

\begin{figure}[t]
\centering
\centerline{\includegraphics[width=\linewidth]{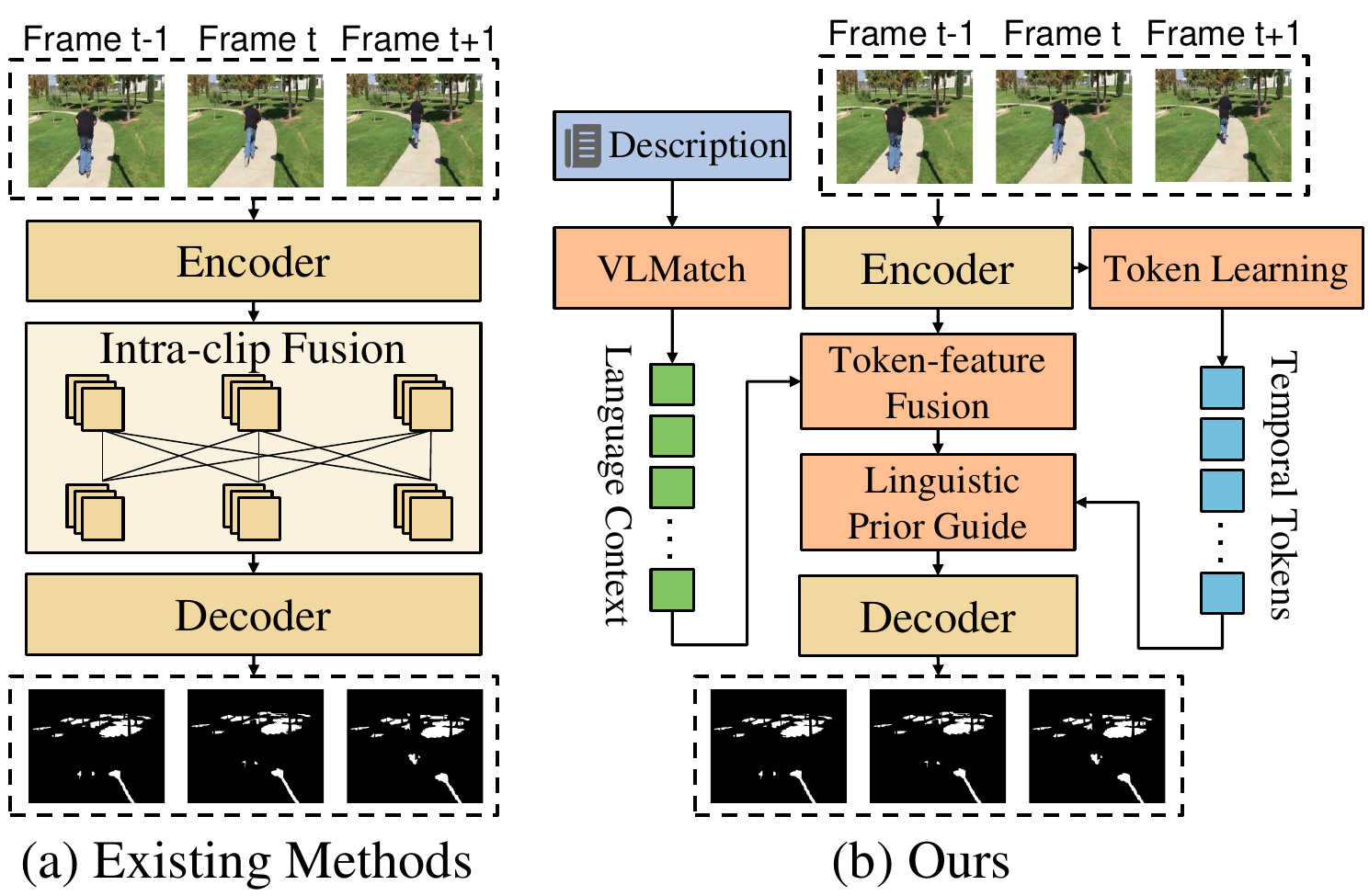}}
\caption{
Comparison between existing methods and ours. Despite using intra-clip fusion, we utilize token to learn temporal characteristics and introduce textual priors to guide the model to learn from dark regions.
}
\label{fig:performance}
\end{figure}

In terms of spatial modeling, a mainstream shadow detection paradigm exploits shadows' inherent low-luminance property by identifying dark regions. To achieve finer-grained shadow modeling, SSTINet~\cite{wei2024structure} decomposes shadow representations into structural and detailed components, serving targeted decoding of shadow regions. 
However, the lack of pixel-level annotations for dark regions confines vision-only shadow modeling to implicit representation learning, hindering explicit localization. To address this, we introduce textual priors to explicitly guide the model's attention towards darkness as shown in Fig.~\ref{fig:performance}. Specifically, to mitigate interference from non-shadow regions within dark areas, we propose distinct text descriptions for shadows and dark regions. 
 Leveraging CLIP's powerful zero-shot capabilities, our proposed Vision-language Match Module (VMM) then performs attention-driven cross-modal matching between these linguistic priors and image features, enhancing focus on both dark regions and shadows. Furthermore, we design the Dark-aware Semantic Block (DSB) to adaptively weight dark and shadow features.

While supervising each DSB output with the shadow mask, the inherent ambiguity of shadow edges (penumbra regions) presents a challenge: excessive focus on blurred boundaries during early training impedes learning of the primary shadow body. Therefore, unlike SSTINet~\cite{wei2024structure}, we apply a reweighting scheme to the supervision mask during training, assigning lower weights to penumbra regions. Additionally, edge masks are employed solely at the final decoder output, ensuring a balanced focus on both the main shadow structure and fine-grained details.
Our main contributions are as follows:
\begin{itemize}
    \item We introduce a novel framework, the Dark-aware and Temporal Tokenized Network(DTTNet), which integrates dark-aware linguistic guidance with tokenized temporal modeling to effectively capture spatial and temporal dependencies, significantly advancing the state-of-the-art in video shadow detection.
    \item We introduce a Tokenized Temporal Block designed to model temporal characteristics at the token level. By decoupling temporal modeling from spatial features, this module effectively captures shadow dynamics for efficient learning of coherent temporal properties.
    \item To resolve visual ambiguities, we leverage textual priors and introduce a Vision-language Match Module (VMM) along with a Dark-aware Semantic Block (DSB), specifically designed for shadow detection in dark regions. Furthermore, we explicitly decouple the penumbra region mask from the main shadow body to facilitate more effective model learning.
\end{itemize}

\section{Related Work}
\subsection{Image Shadow Detection} 
Image shadow detection has advanced through diverse innovations~\cite{khan2014automatic,hu2021revisiting,wang2018stacked,zhu2021mitigating}. Some research use various fusion modules to enhance the ability of the encoder. For example, BDRAR~\cite{zhu2018bidirectional} combines global and local context via a bidirectional pyramidal architecture and DSD~\cite{zheng2019distraction} learns distraction-aware features to reduce false positives through explicit modeling of ambiguous regions. Others try to model shadow and background separately. SDDNet~\cite{cong2023sddnet} decomposes shadow and background features with style-guided disentanglement to mitigate background color interference. Recently, 
\cite{guan2024delving} propose a dark-region recommendation module to enhance discrimination in low-intensity regions. 
Although effective on static images, these methods lack mechanisms to capture temporal dependencies, making them suboptimal for videos where shadows deform dynamically under varying illumination.

\begin{figure*}[t]
\centering
\includegraphics[width=1\linewidth]{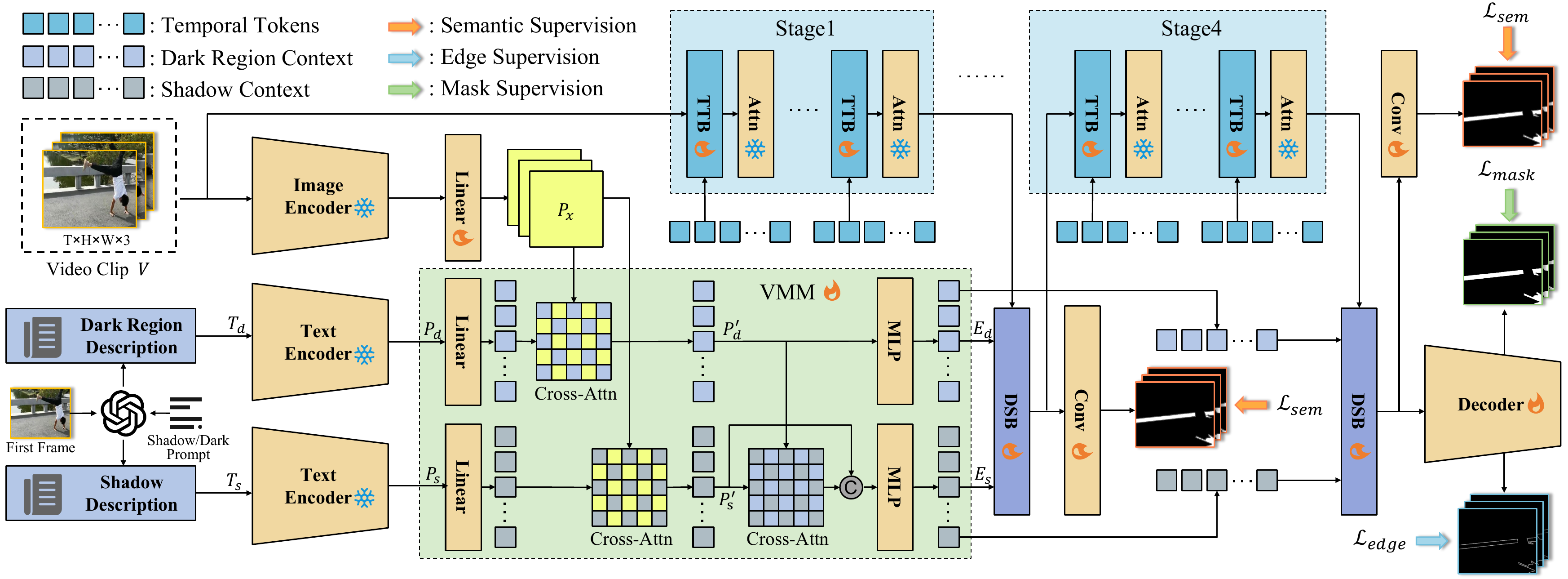}
\caption{Architecture of Dark-aware and Temporal Tokenized Network (DTTNet). DTTNet integrates dark-aware linguistic guidance with tokenized temporal modeling to effectively capture spatial and temporal dependencies. It consists of three novel modules: the Vision-language Match Module (VMM), Dark-aware Semantic Block and the Tokenized Temporal Block. We freeze most of the parameters and update only the parameter of decoder and proposed modules. }
\label{fig:overview}
\vspace{-3mm}
\end{figure*}

\subsection{Video Shadow Detection}  
Research on video shadow detection has gained momentum. \cite{chen2021triple} pioneered this field by constructing the first dedicated dataset and proposing a triple-flow network. Subsequent advancements have further expanded the landscape: SC-Cor~\cite{ding2022learning} enhances cross-frame feature coherence for shadow regions without relying on pixel-wise labels. SCOTCH\&SODA~\cite{liu2023scotch} introduced transformers into the task, leveraging trajectory attention to handle deformations and a contrastive loss to learn unified shadow representations. DAS~\cite{wang2023detect} employed SAM~\cite{kirillov2023segment} and designed a long short-term network for subsequent frame predictions. Recently, \cite{duan2024two} proposed a two-stage paradigm and the CVSD dataset to bring the task to complex scenarios. SSTINet~\cite{wei2024structure} deployed a structure-aware module focusing on edge-region distance relations. TBGDiff~\cite{zhou2024timeline} uses diffusion model with the guidance of boundary and long-term frames to generate shadow masks. Most existing methods overlook shadows' intrinsic properties, directly extracting regions from features, whereas our approach introduces linguistic priors to guide dark-region information extraction and uses tokens for temporal correlation learning, moving beyond multi-frame feature fusion.


\section{Methodology}
\subsection{Overview}

As shown in Fig.~\ref{fig:overview}, the proposed DTTNet leverages text priors for video shadow detection. For an input video clip $V$, a large multimodal model processes its first frame to generate descriptive texts for shadows $T_s$ and dark regions $T_d$ via a predefined template. The tuple ($V$, $T_s$, $T_d$) is fed into the pre-trained CLIP model for feature extraction. Then the proposed  VMM performs text-image matching, yielding visual-informed context for shadows and dark regions. Critically, at each encoder stage, this context is injected into our DSB to enhance encoder features with text priors, supervised by penumbra-aware mask for accuracy. Concurrently, a TTB within each encoder stage employs learnable tokens to summarize temporal information from preceding features. This tokenized temporal feature is fused via attention before feature propagation. The refined encoder features are finally decoded to output shadow masks $M$ for all frames.

\subsection{Vision-language Match Module}
Shadow regions often overlap with dark areas, causing ambiguity under complex lighting conditions when relying solely on visual cues. To address this challenge, we leverage textual priors to identify shadows and dark regions. For each video clip $V$, a vision-language large model is employed to generate structured descriptions—$T_s$ for shadows and $T_d$ for dark regions—based on a predefined template. The prompt used for generating these descriptions is:
\emph{``Describe the OBJECT, COUNT, POSITION, and DETAIL of all the shadow/dark areas in the image. When answering COUNT, do not use numbers just say single or multiple. Your answer must be like this: {COUNT} shadows/dark regions, shadow/dark region of {OBJECT}, {POSITION}, {DETAIL}."}

Exploiting CLIP's zero-shot power, we freeze the model parameters of CLIP and further feed both $T_s ,T_d$ and $V\in \mathbb{R}^{T\times H\times W\times 3}$ into CLIP to generate corresponding text features $P_{s}\in \mathbb{R}^{L_{s}\times C_{l}}, P_{d}\in \mathbb{R}^{L_{d}\times C_{l}}$ and image feature $ P_x\in \mathbb{R}^{T\times M\times C_{m}}$, where $L_{s}$, $L_{d}$ means length of shadow and dark region descriptions and $C_{l}$,$C_{m}$ refers to channel number of CLIP's text and image embedding. 
Our Vision-language Match Module then aligns these modalities, generating visual-informed contexts $E_s\in \mathbb{R}^{L_{s}\times C_e}$ and $E_d \in \mathbb{R}^{L_{d}\times C_e} $ for shadow and dark region respectively, where $C_e$ means the embedding channel number of DTTNet.
We project both modalities to a shared channel dimension via fully connected layers. For dark region context, cross-attention with $P_d$ as query and $P_x$ as key/value yields $P'_d$:
\begin{equation}
P'_d = \text{Attn}(P_d,P_x),
\end{equation} 
$P'_d$ is then processed by a multi-layer perceptron (MLP) to yield the refined dark region context $E_d$. 
\begin{equation}
    E_d = \text{MLP}(P'_d), 
\end{equation}
For shadow region context, exploiting dark regions' cues, we first apply cross-attention ($P_s$ as query, $P_x$ as key/value) for initial features $P'_s$. Then $P'_s$ is fused with $P'_d$ via another cross-attention.The concatenated outputs of both cross-attention operations are fed into a MLP to generate the final shadow context $E_s$:
\begin{equation}
    E_s = \text{MLP}(\text{Cat}(\text{Attn}(P_s,P_x),\text{Attn}(P'_s,P'_d))).
\end{equation}
Here $\text{Attn}(\cdot)$ denotes the standard attention mechanism, $\text{Cat}(\cdot)$ represents concatenation along the channel dimension, and 
the module $\text{MLP}(\cdot)$ consists of two sequential linear layers that first compress and then restore the channel dimensionality of the input feature representations.

\subsection{Dark-aware Semantic Block}
Compared to the extracted image features, text features contain more accurate high-level semantics. Therefore, we leverage shadow/dark region context (\textit{i.e.} $E_s$ and $E_d$) to perform semantic adjustment of the image features, rather than pixel-level classification (which is better suited for the decoder).
We introduce a DSB before feeding the outputs of each encoder stage to the decoder. The DSB enhances the semantic expressiveness of the features. Furthermore, penumbra-aware supervision is imposed on the DSB outputs to ensure semantic accuracy.

\begin{figure}[t]
\centering
\centerline{\includegraphics[width=\linewidth]{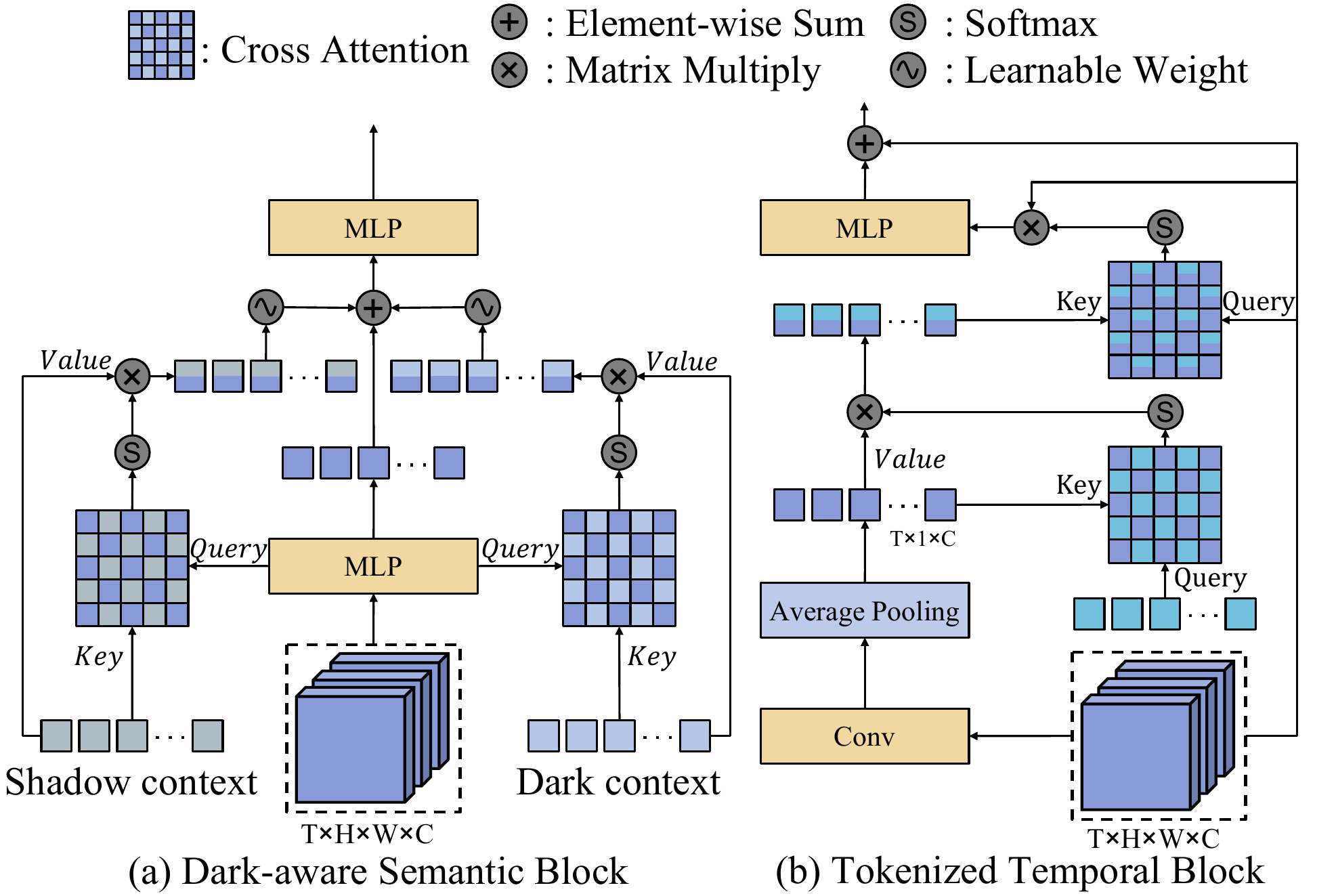}}
\caption{
Details of the proposed Dark-aware Semantic Block and Tokenized Temporal Block.
}
\label{fig:detail}
\end{figure}

Specifically, as shown in Fig.~\ref{fig:detail}(a), the encoder feature from the $i$-th stage, denoted as $X_i \in \mathbb{R}^{T\times \frac{H}{16}\times \frac{W}{16}\times C_{b}}$, undergoes processing to integrate shadow ($E_s$) and dark region ($E_d$) contexts. First, a fully connected layer compresses the channel dimension of $X_i$ to $C_e$ to align with the channel dimensionality of the context embeddings. Subsequently, cross-attention mechanisms individually inject $E_s$ and $E_d$ into the compressed feature, yielding shadow-enhanced feature $X_i^s$ and dark-region-enhanced feature $X_i^d$:
\begin{equation}
\begin{split}
X_i^s = \text{Attn}(\text{Linear}(X_i),E_s),\\
X_i^d = \text{Attn}(\text{Linear}(X_i),E_d),
\end{split}
\end{equation}
Here $\text{Linear}(\cdot)$ denotes the linear layer. 
These two enhanced features are then dynamically fused using learned weighting parameters (\textit{i.e.} $\alpha$ and $\beta$). The fused representation is passed through a $1\times1$ convolutional layer to generate a soft shadow mask $\tilde{M}_{i}$, which is supervised by the processed ground-truth shadow annotation $\hat{M}$. Concurrently, the fused representation is directly added to the original encoder feature $X_i$, resulting in the final enhanced feature output $X^{f}_i$:
\begin{equation}
    X^{f}_i = X_i + \alpha * X_i^s + \beta * X_i^d,
\end{equation}
where $\alpha$ and $\beta$ are learnable weights that allow the model to adapt the importance between shadow and dark regions.

\subsection{Tokenized Temporal Block}
Existing approaches primarily encode temporal information through frame-level feature fusion, neglecting learning coherent temporal representations. To address this, we introduce the Tokenized Temporal Block (TTB). This module employs a collection of learnable tokens to acquire a universal temporal representation applicable across all shadow videos after training. This representation enables more efficient extraction of consistent shadow features across consecutive frames.
As shown in Fig.~\ref{fig:overview}, we introduce a TTB before each layer of the encoder to capture temporal information prior to feature processing. Fig.~\ref{fig:detail}(b) shows the structure of TTB. For the encoder feature $X_{j-1}$ output from the $(j-1)$-th layer, its channel dimension is first aligned via a $1\times1$ convolutional layer. Subsequently, spatial information is compressed using average pooling, yielding a temporal feature $Z_{j} \in \mathbb{R}^{T\times 1\times C_e}$:
\begin{equation}
Z_{j} = \text{AvgPool}(\text{Conv}(X_{j-1})),
\end{equation}
where $\text{Conv}(\cdot)$ refers to the $1\times1$ convolution.
A set of learnable tokens $K_j\in \mathbb{R}^{L_{k}\times C_e}$ is then used as queries, with $Z_j$ serving as keys and values in a cross-attention operation, enabling the tokens to assimilate relevant temporal information from the temporal sequence:
\begin{equation}
K_{j} = \text{Attn}(K_j,Z_{j}).
\end{equation}
Following this, the original feature $X_{j-1}$ is spatially flattened to $N \times T \times C_e$ and undergoes cross-attention with the learned tokens. This process allows each frame to effectively incorporate information from the tokens while preserving its original spatial structure.
Finally, a $1\times1$ convolutional layer is used to restore the channel dimension, preparing the feature $X_j$ for processing in the $j$-th encoder layer:
\begin{equation}
X_{j} = \text{Conv}(\text{Attn}(K_j,X_{j-1})).
\end{equation}
\subsection{Decoder}


For features extracted at each encoder stage, we first apply a MLP to model channel-wise dependencies. The resulting features from all stages are then integrated through a convolutional block comprising sequential convolutional, batch normalization, ReLU activation, and convolutional layer. As outlined in Fig.~\ref{fig:overview}, we design a decoder that gradually upsamples features from the compressed feature size of $\frac{H}{16} \times \frac{W}{16}$ to $\frac{H}{4} \times \frac{W}{4}$. To mitigate information loss and noise amplification during upsampling, 
we perform convolutions before interpolations and employ DySample~\cite{liu2023learning} as the upsampling method. 
After two successive upsampling stages, the decoder employs an additional convolutional layer to produce preliminary mask features. Subsequently, two parallel $1\times1$ convolutional layers generate the final shadow mask prediction $\tilde{M}_{s}$ and edge mask prediction $\tilde{M}_{e}$ from the above mask features.

\begin{table*}[t]
\centering
\setlength{\tabcolsep}{2mm}
\resizebox{0.85\textwidth}{!}{
\begin{tabular}{c|c|c|cccc|cc}

\toprule
\multicolumn{3}{c|}{{Methods}}                                                                                    & \multicolumn{6}{c}{{Evaluation Metrics}}                                                                       \\ 

\midrule 
Tasks    & Techniques    & Year   & MAE $\downarrow$   & $\textrm{F}_{\beta} \uparrow$  & IoU $\uparrow$   & BER $\downarrow$  & S-BER $\downarrow$ & N-BER $\downarrow$ \\ 

\midrule
\multirow{4}{*}{IOS} & FPN~\cite{lin2017feature}                  & 2017       & 0.044 & 0.707 & 0.512 & 19.49 & 36.59 & 2.40 \\
                     & $\mathrm{R}^3$Net~\cite{deng2018r3net}     & 2018       & 0.043 & 0.710 & 0.502 & 20.40 & 37.37 & 3.55 \\ 
                     & Segformer~\cite{xie2021segformer}          & 2021       & 0.030 & 0.773 & 0.601 & 11.56 & 21.39 & 1.73 \\ 
                     & DDP~\cite{ji2023ddp}                       & 2023       & 0.038 & 0.771 & 0.608 & 10.74 & 18.90 & 2.57 \\ 
                     
                     \midrule 
\multirow{5}{*}{VOS} & STM~\cite{oh2019video}                     & 2019       & 0.069 & 0.598 & 0.408 & 25.69 & 47.44 & 3.95 \\ 
                     & COSNet~\cite{lu2019see}                    & 2019       & 0.040 & 0.706 & 0.515 & 20.51 & 39.22 & 1.79 \\  
                     & FEELVOS~\cite{voigtlaender2019feelvos}     & 2019       & 0.043 & 0.710 & 0.512 & 19.76 & 37.27 & 2.26 \\  
                     & STCN~\cite{cheng2021rethinking}            & 2021       & 0.048 & 0.684 & 0.528 & 12.42 & 21.36 & 3.48 \\   
                     & Pix2Seq~\cite{chen2023generalist}          & 2023       & 0.034 & 0.775 & 0.618 & 10.63 & 19.13 & 2.14 \\  
                     
                     \midrule 
\multirow{6}{*}{ISD} & BDRAR~\cite{zhu2018bidirectional}          & 2018       & 0.050 & 0.695 & 0.484 & 21.30 & 40.28 & 2.32 \\ 
                     & DSD~\cite{zheng2019distraction}            & 2019       & 0.044 & 0.702 & 0.519 & 19.89 & 37.88 & 1.89 \\ 
                     & MTMT~\cite{chen2020multi}                  & 2020       & 0.043 & 0.729 & 0.517 & 20.29 & 38.71 & 1.86 \\ 
                     & FSD~\cite{hu2021revisiting}                & 2021       & 0.057 & 0.671 & 0.486 & 20.57 & 38.06 & 3.06 \\ 
                     & SDDNet~\cite{cong2023sddnet}               & 2023       & 0.040 & 0.754 & 0.548 & 14.05 & 26.10 & 1.61 \\ 
                     & SILT~\cite{yang2023silt}                   & 2023       & 0.031 & 0.796 & 0.606 & 12.80 & 24.29 & 1.29 \\  
                     
                     \midrule 
\multirow{7}{*}{VSD} & TVSD~\cite{chen2021triple}                 & 2021       & 0.033 & 0.757 & 0.565 & 17.70 & 33.96 & 1.44 \\ 
                     & STICT~\cite{lu2022video}                   & 2022       & 0.046 & 0.702 & 0.545 & 16.60 & 29.58 & 3.59 \\ 
                     & SC-Cor~\cite{ding2022learning}             & 2022       & 0.042 & 0.762 & 0.615 & 13.61 & 24.31 & 2.91 \\
                     & SCOTCH \& SODA~\cite{liu2023scotch}             & 2023       & 0.029 & 0.793 & 0.640 & 9.06 & 16.26 & 1.44 \\  
                     & DAS~\cite{wang2023detect}                  & 2023       & 0.034 & 0.754 & 0.575 & 12.58 & 23.60 & 1.57 \\ 
                     & TBGDiff~\cite{zhou2024timeline}            & 2024       & 0.023 & 0.797 & 0.667 & 8.58 & 16.00 & 1.15 \\  
                     & TSVSD~\cite{duan2024two}                & 2024       & 0.027 & 0.801 & 0.684 & 8.96  & -     & -    \\  
                     & SSTINet~\cite{wei2024structure}            & 2024       & \underline{0.017} &\textbf{0.866} & \textbf{0.746} & \underline{6.48} & \underline{12.32} & \underline{0.65} \\ 
                     \rowcolor{gray!17}&  DTTNet (Ours)                              & -          & \textbf{0.016} & \underline{0.849} & \underline{0.718} & \textbf{6.45} & \textbf{12.29} & \textbf{0.61} \\

    \bottomrule 
\end{tabular}
}
\caption{Quantitative comparisons between our proposed method and SOTA methods on the ViSha~\cite{chen2021triple} dataset. "↑" denotes the higher the value, the better the performance, and "↓" means the lower the value, the better the performance. We compare with recent methods from Image Object Segmentation (IOS), Image Shadow Detection (ISD), Video Object Segmentation (VOS), and Video Shadow Detection (VSD). The best values are highlighted in \textbf{bold}, while the second best values are \underline{underlined}.}
\label{tab:res}
\end{table*}

\subsection{Loss function}
Our loss function consists of three components: penumbra-aware semantic loss $\mathcal{L}_{sem}$, shadow edge loss $\mathcal{L}_{edge}$, and shadow mask loss  $\mathcal{L}_{mask}$.
Given that encoder features at low resolutions prioritize semantic accuracy over pixel-level classification fidelity, we formulate the semantic loss as a regression task using mean absolute error to supervise the features of the DSB. To address ambiguity in shadow boundaries (penumbra regions), we attenuate edge values in proportion to their distance from the center while preserving values of main regions. This strategy focuses the DSB on capturing definitive shadow structures while mitigating interference from uncertain boundaries, implemented as:
\begin{equation}
     \hat{M}(u,v)= 
\begin{cases} 
1, &  \text{if }~\text{Eros}(M(u,v)) > 0 \\
\text{Dist}(M(u,v)), &  \text{else}
\end{cases}
\end{equation}
\begin{equation}
    \mathcal{L}_{sem} = \sum_{i}L_{mse}({\tilde{M}_{i}},\hat{M}),
\end{equation}
where $(u, v)$ denotes the location on the reprocessed mask $\hat{M}$ and $L_{mse}$ refers to the Mean Squared Error loss. $\tilde{M}_{i}$ is the auxiliary output of DSB in the $i$-th stage.~$\text{Eros}(\cdot)$ refers to the morphological erosion operation with a kernel size of 3.~$\text{Dist}(\cdot)$ represents the distance transform function. 
To supervise the final shadow mask, we employ a dual-branch supervision strategy where the original shadow mask and extracted edge components serve as distinct supervisory signals for the decoder's output features. For edge components, we calculate edge loss as follows:
\begin{equation}
    M_{e} = \mathbb{I}(1 - \hat{M}),
\end{equation}
\begin{equation}
    \mathcal{L}_{edge} = L_{bce}(\tilde{M}_{e},M_{e}) + L_{dice}(\tilde{M}_{e},M_{e}),
\end{equation}
where $M_{e}$ is defined as the ground truth edge mask and $\mathbb{I}(\cdot)$ means instruction function. Meanwhile, we calculate mask loss for shadow mask:
\begin{equation}
    \mathcal{L}_{mask} = L_{bce}(\tilde{M}_{s},M) + L_{dice}(\tilde{M}_{s},M),
\end{equation}
where $L_{bce}$ means Binary Cross-Entropy with Logits loss and $ L_{dice}$ refers to the Dice loss~\cite{milletari2016v}.

The final loss function is composed of the weighted sum of these three losses:
\begin{equation}
    \mathcal{L} =\lambda_1 * \mathcal{L}_{sem} + \lambda_2 * \mathcal{L}_{edge}+ \lambda_3 * \mathcal{L}_{mask},
\end{equation}
$\lambda_1$, $\lambda_2$ and $\lambda_3$ are hyper-parameters that balance the losses. 
\section{Experiments}

\subsection{Dataset and Evaluation Metrics}

\subsubsection{Datasets.}
The proposed DTTNet is validated on the Visha~\cite{chen2021triple} and CVSD~\cite{duan2024two} datasets. Following previous studies~\cite{wang2023detect,ding2022learning,liu2023scotch}, we employ the Visha dataset to evaluate the model’s performance. Visha comprises 120 videos with diverse content and varying lengths, with more than half of the clips derived from standard video tracking benchmarks. In addition, we conduct experiments on the more challenging CVSD dataset~\cite{duan2024two}, which was recently introduced and includes 196 video clips and a total of 19,757 frames featuring complex shadow patterns.

\subsubsection{Evaluation Metrics.}
To facilitate a fair and thorough performance comparison, we adopt the methodology established in prior research and compute six distinct evaluation metrics. These include Mean Absolute Error (MAE), F-measure, and Intersection over Union (IoU), as well as Balanced Error Rate (BER). Additionally, we incorporate the S-BER score, which is specifically designed for shadow regions, and the N-BER score tailored to non-shadow regions.

\subsection{Implementation Details}
The proposed model is implemented using the MMSegmentation codebase~\cite{contributors2020mmsegmentation}. The backbone network of DTTNet is the pre-trained DINOv2~\cite{oquab2023dinov2}. All parameters of the backbone are kept frozen, while the trainable parameters are exclusively derived from four new components: the Vision-language Match Module, Tokenized Temporal Block, Dark-aware Semantic Block, and the decoder. For optimizing these parameters, the AdamW is employed, configured with the learning rate of $5*10^{-5}$ and the weight decay of 0.01. Training is conducted with a batch size of 2 and each batch includes 5 frames of the video. 
We set $\lambda_1$, $\lambda_2$, and $\lambda_3$ to 1, 0.5, and 1, respectively.
All experiments are carried out at a resolution of 512×512, and Random Horizontal Flip is adopted in the training phase.

\begin{table}[t]
\centering
\setlength{\tabcolsep}{1.9mm}
\resizebox{1.01\linewidth}{!}{
\begin{tabular}{c|c|cccc}
\toprule
\multicolumn{2}{c|}{{Methods}} & \multicolumn{4}{c}{{Evaluation Metrics}}                                                                   \\ 
\midrule
Techniques    & Year   & MAE $\downarrow$   & $\textrm{F}_{\beta} \uparrow$  & IoU $\uparrow$   & BER $\downarrow$  \\ 
\midrule
TVSD            & 2021       & 0.099 & 0.539 & 0.369 & 27.28 \\ 
STICT              & 2022       & 0.073 & 0.608 & 0.447 & 23.27 \\ 
SC-Cor        & 2022       & 0.070 & 0.573 & 0.476 & 19.94 \\
SCOTCH \& SODA   & 2023       & 0.082 & 0.585 & 0.426 & 23.27 \\  
DAS             & 2023       & 0.087 & 0.561 & 0.435 & \underline{19.15} \\ 
TBGDiff       & 2024       & 0.057 &  \underline{0.663} & 0.445 & 23.01 \\  
TSVSD              & 2024       & \underline{0.046} &0.638 &  \underline{0.515} &  \textbf{18.32} \\ 
\rowcolor{gray!17} 
DTTNet (Ours)                                  & -          & \textbf{0.042} & \textbf{0.766} & \textbf{0.548} & 23.32 \\ 
\bottomrule
\end{tabular}
}
\caption{Quantitative comparisons between our proposed method and SOTA methods on CVSD~\cite{duan2024two} dataset. }
\label{tab:res2}
\end{table}

\subsection{Comparison With State-of-the-art Methods}
\subsubsection{Compared Methods.}
We compare our network with 21 methods across related tasks,
including IOS methods: FPN~\cite{lin2017feature}, $R^3$Net~\cite{deng2018r3net}, Segformer~\cite{xie2021segformer}, DDP~\cite{ji2023ddp}; VOS methods: STM~\cite{oh2019video}, COSNet~\cite{lu2019see}, FEELVOS~\cite{voigtlaender2019feelvos}, STCN~\cite{cheng2021rethinking}, Pix2Seq~\cite{chen2023generalist}; ISD methods BDRAR~\cite{zhu2018bidirectional}, DSD~\cite{zheng2019distraction}, MTMT~\cite{chen2020multi}, FSD~\cite{hu2021revisiting}, SDDNet~\cite{cong2023sddnet}, SILT~\cite{yang2023silt}; and video shadow detection (VSD) TVSD~\cite{chen2021triple}, STICT~\cite{lu2022video}, SC-Cor~\cite{ding2022learning}, SCOTCH \& SODA~\cite{liu2023scotch}, DAS~\cite{wang2023detect}, TBGDiff~\cite{zhou2024timeline}, TSVSD~\cite{duan2024two} and SSTINet~\cite{wei2024structure}. 
\vspace{-2mm}

\subsubsection{Quantitative Comparisons.}
Quantitative results for our approach alongside other methods are presented in Table~\ref{tab:res}. Since ISD methods are purpose-built for shadow detection, they outperform VOS and IOS networks in this task. Nevertheless, these ISD methods lack the integration of temporal information, resulting in inferior performance compared to VSD techniques. Among all existing approaches, SSTINet~\cite{wei2024structure} delivers the strongest results, yet our proposed method exceeds current state-of-the-art techniques in terms of MAE, BER, S-BER, and N-BER metrics. Specifically, our method reduces the MAE from 0.017 to 0.016, lowers the BER from 6.48 to 6.41, improves the S-BER from 12.32 to 12.29, and decreases the N-BER from 0.65 to 0.53. Additionally, on the CVSD dataset~\cite{duan2024two}—which has not been included the evaluation scope of most existing studies—our approach attains the top performance in MAE, $F_{\beta}$, and IoU metrics, with detailed results provided in Table~\ref{tab:res2}.

\begin{table}[t]
\centering
\setlength{\tabcolsep}{2mm}
\resizebox{1.01\linewidth}{!}{
\begin{tabular}{c|cccccc}
\toprule
Methods & Params(M) & FPS & IoU$\uparrow$ & BER$\downarrow$\\
\midrule
SCOTCH \& SODA & 211.8 & 11.4 & 0.640 & 9.07 \\
DAS & 101.3 & 12.1 & 0.667 & 8.58 \\
TBGDiff & \underline{102.3} & \underline{13.5} & 0.640 & 9.07 \\
SSTINet & 338.3 & 6.06 & {\bf 0.746} & \underline{6.48} \\
\rowcolor{gray!17}
DTTNet (Ours) & {\bf 46.6 (502.1)} & {\bf 26.63} & \underline{0.714} & {\bf 6.45} \\
\bottomrule
\end{tabular}
}
\caption{Parameters and efficiency on ViSha~\cite{chen2021triple} dataset. Since we freeze most of the parameters, the total number of parameters is provided in parentheses.}
\label{tab:ef}
\end{table}

\begin{table*}[t]
\setlength{\tabcolsep}{3.2mm}
\centering
\begin{tabular}{c|c|c|c|cccc|cccc}
\toprule
\multirow{2}{*}{Index} & \multicolumn{3}{c|}{{Components}} & \multicolumn{4}{c|}{{ViSha~\cite{chen2021triple}}} & \multicolumn{4}{c}{{CVSD~\cite{duan2024two}}} \\ 
\cmidrule{2-12}
                        & VMM & DSB & TTB & MAE$\downarrow$ & $F_{\beta}\uparrow$ & IoU$\uparrow$ & BER$\downarrow$ & MAE$\downarrow$ & $F_{\beta}\uparrow$ & IoU$\uparrow$ & BER$\downarrow$\\ 
\midrule
(a)&-&-&-& 0.027 & 0.775 & 0.640 & 10.01 & 0.048 & 0.720 & 0.501 & 25.85\\
(b)& - &$\checkmark$ &-& 0.023 & 0.807 & 0.671 & 8.47 & 0.045 & 0.731 & 0.522 & 24.77\\
(c)& $\checkmark$ &$\checkmark$&-& 0.021 & 0.817 & 0.686 & 7.98 & 0.044 & 0.754 & 0.534 & {\bf 22.61}\\
(d)& -&-&$\checkmark$& 0.017 & 0.833 & 0.712 & 7.27 & 0.043 & 0.761 & 0.545 & 23.68\\
\rowcolor{gray!17}
(e)& $\checkmark$&$\checkmark$&$\checkmark$& {\bf 0.016} & {\bf 0.849} & {\bf 0.718} & {\bf 6.45} & {\bf 0.042} & {\bf 0.766} & {\bf 0.548} & 23.32\\
\bottomrule
\end{tabular}
\caption{Ablation study on components of DTTNet. The best values are highlighted in bold.}
\label{tab:ablation}
\end{table*}

\subsubsection{Qualitative Comparisons.}
We present in Fig~\ref{fig:res1} the shadow masks generated by DTTNet and state-of-the-art methods (see \textbf{supplementary file} for more comparisons).  The first two rows illustrate scenarios with interference from other dark regions: we observe that existing methods struggle to focus on shadow regions and are consistently distracted by dark areas, whereas our method can minimize such interference to accurately localize shadows. For instance, in the first row, dark streaks formed at the road edge above tree shade cause other methods to mistakenly classify them as shadows. Furthermore, the black baffle surrounding the sandpit in the second row also disrupts other approaches, while DTTNet effectively identifies shadow regions while excluding these distractors—demonstrating that our network does not simply equate dark regions with shadows. In the third row, the gray rough ground texture is highly misleading: TBGDiff~\cite{zhou2024timeline} fails to classify it as background, whereas our results exhibit the least noise. The last row shows a blurry high-speed captured scene where the athlete’s shadow is nearly indistinguishable from the track and occupies a tiny area, leading other methods to miss the shadow entirely. In contrast, our network successfully discriminates the blurry shadow with the highest accuracy, showcasing its segmentation performance.

\begin{figure*}[t]
\centering
\includegraphics[width=0.1\textwidth]{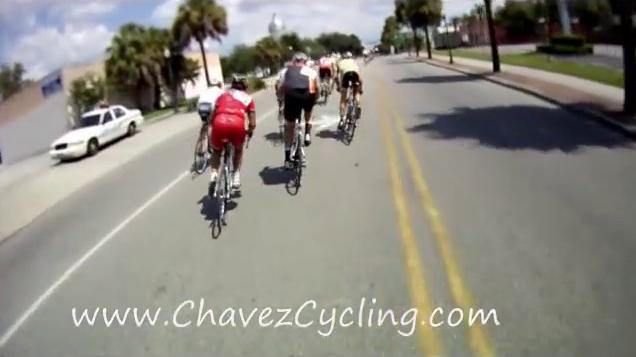}
\includegraphics[width=0.1\textwidth]{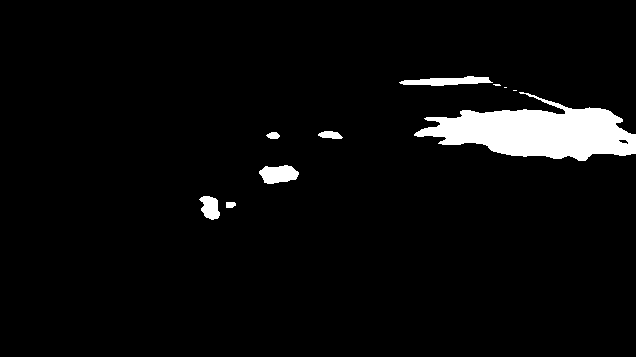}
\includegraphics[width=0.1\textwidth]{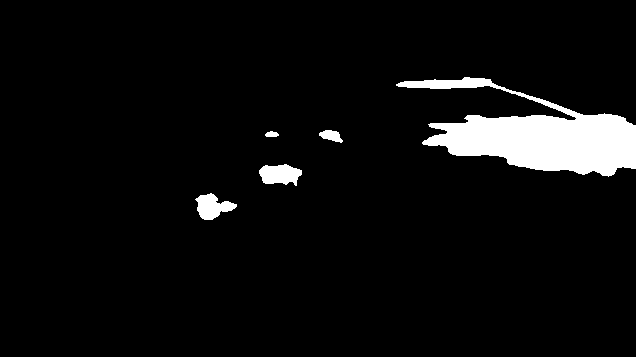}
\includegraphics[width=0.1\textwidth]{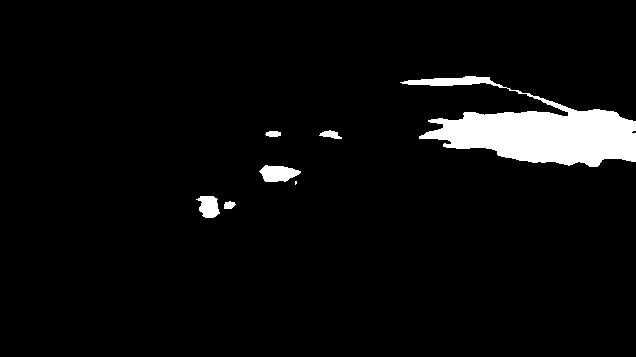}
\includegraphics[width=0.1\textwidth]{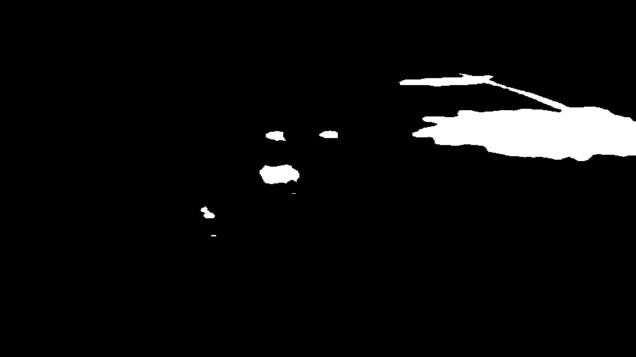}
\includegraphics[width=0.1\textwidth]{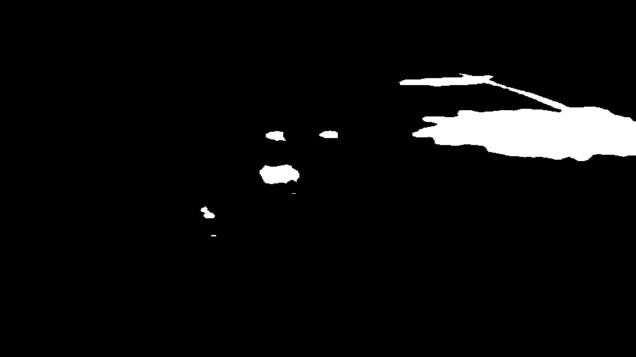}
\includegraphics[width=0.1\textwidth]{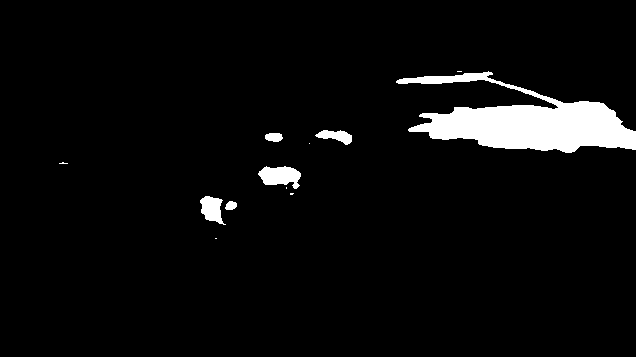}
\includegraphics[width=0.1\textwidth]{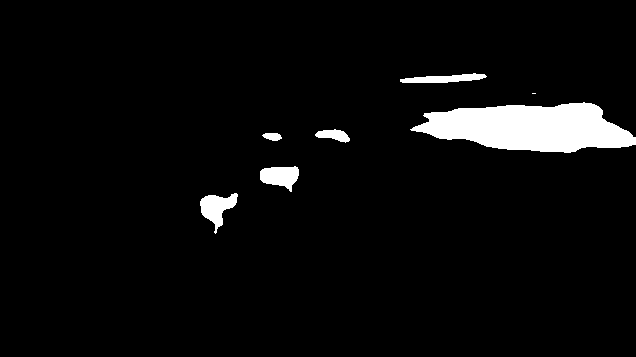}
\includegraphics[width=0.1\textwidth]{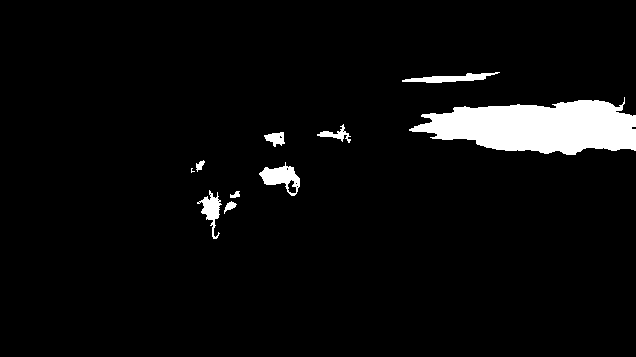}

\includegraphics[width=0.1\textwidth]{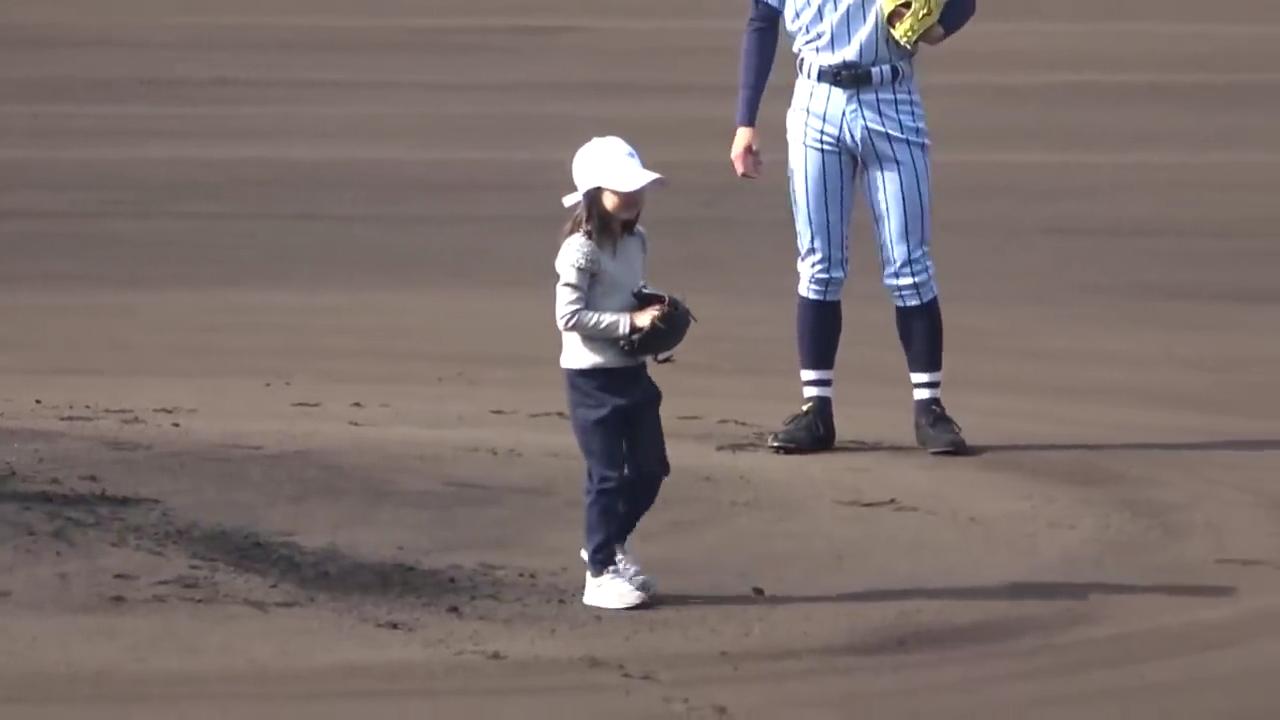}
\includegraphics[width=0.1\textwidth]{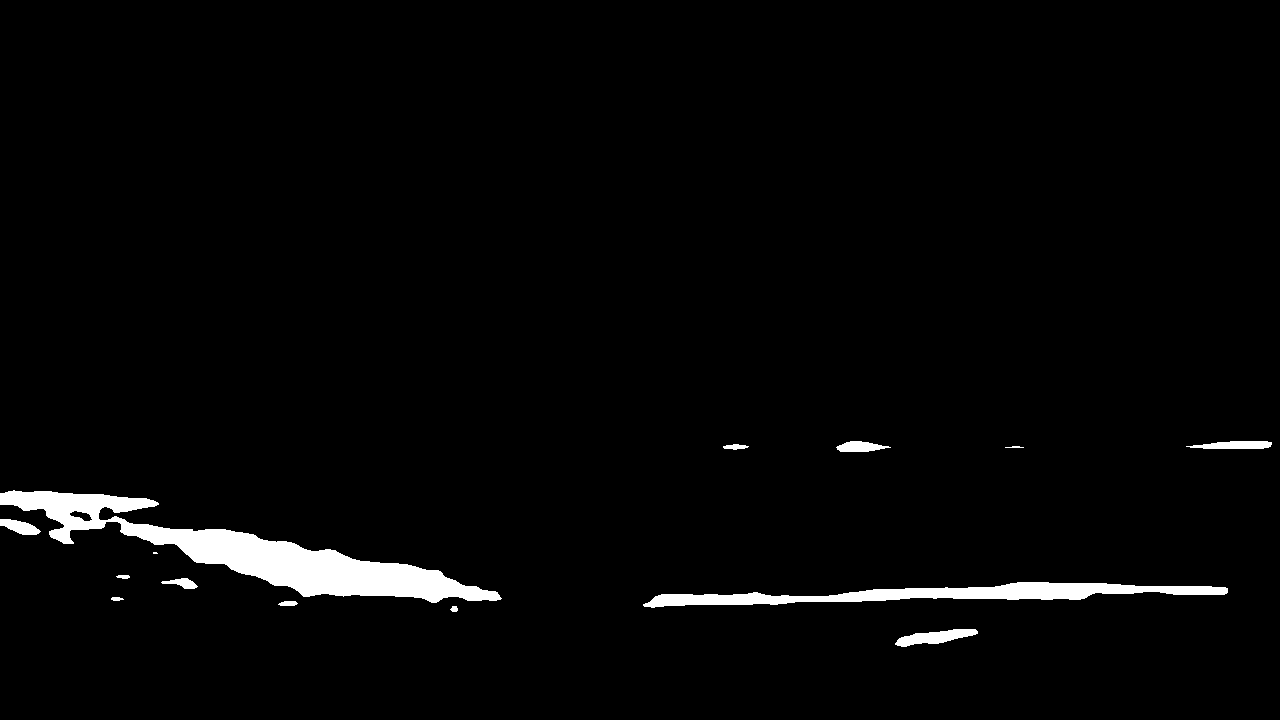}
\includegraphics[width=0.1\textwidth]{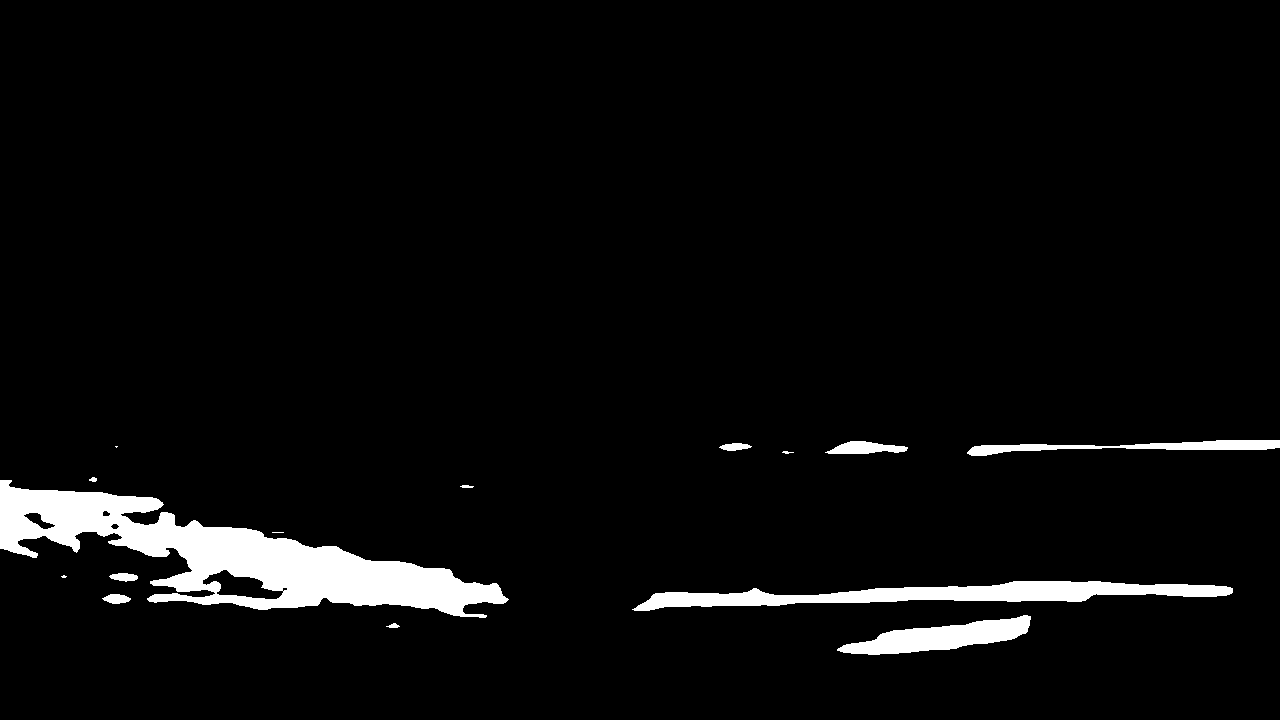}
\includegraphics[width=0.1\textwidth]{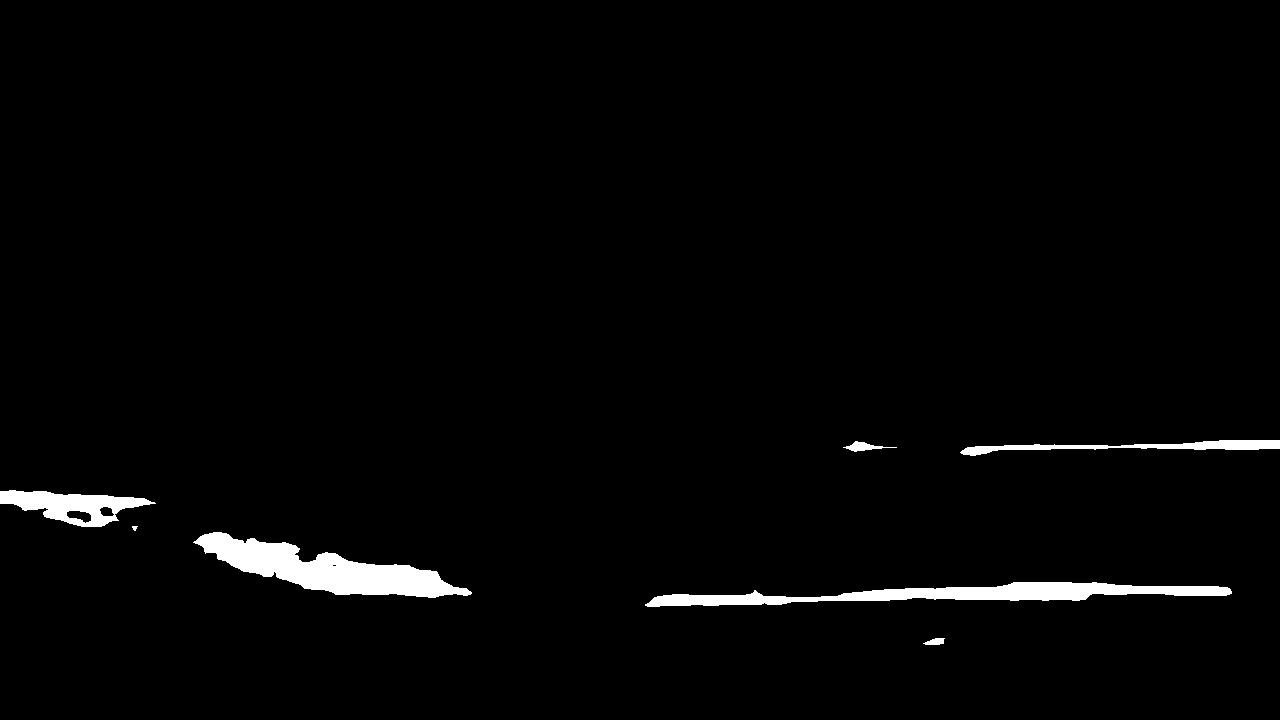}
\includegraphics[width=0.1\textwidth]{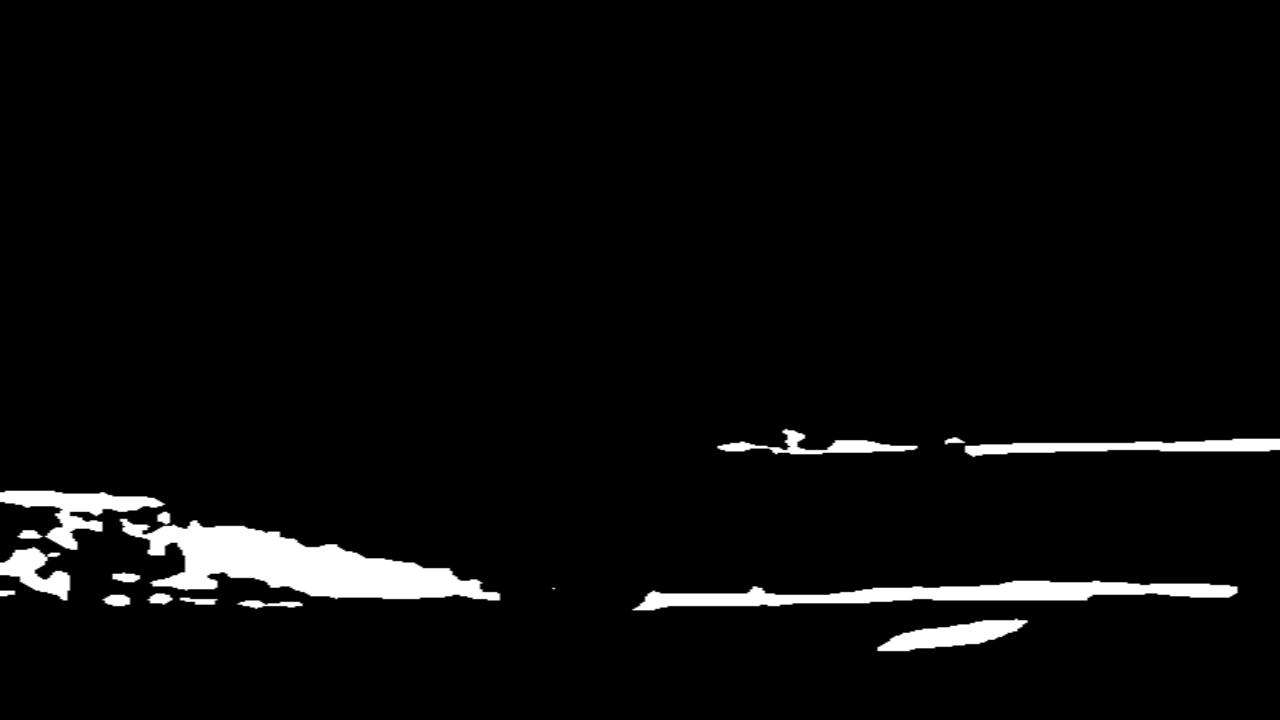}
\includegraphics[width=0.1\textwidth]{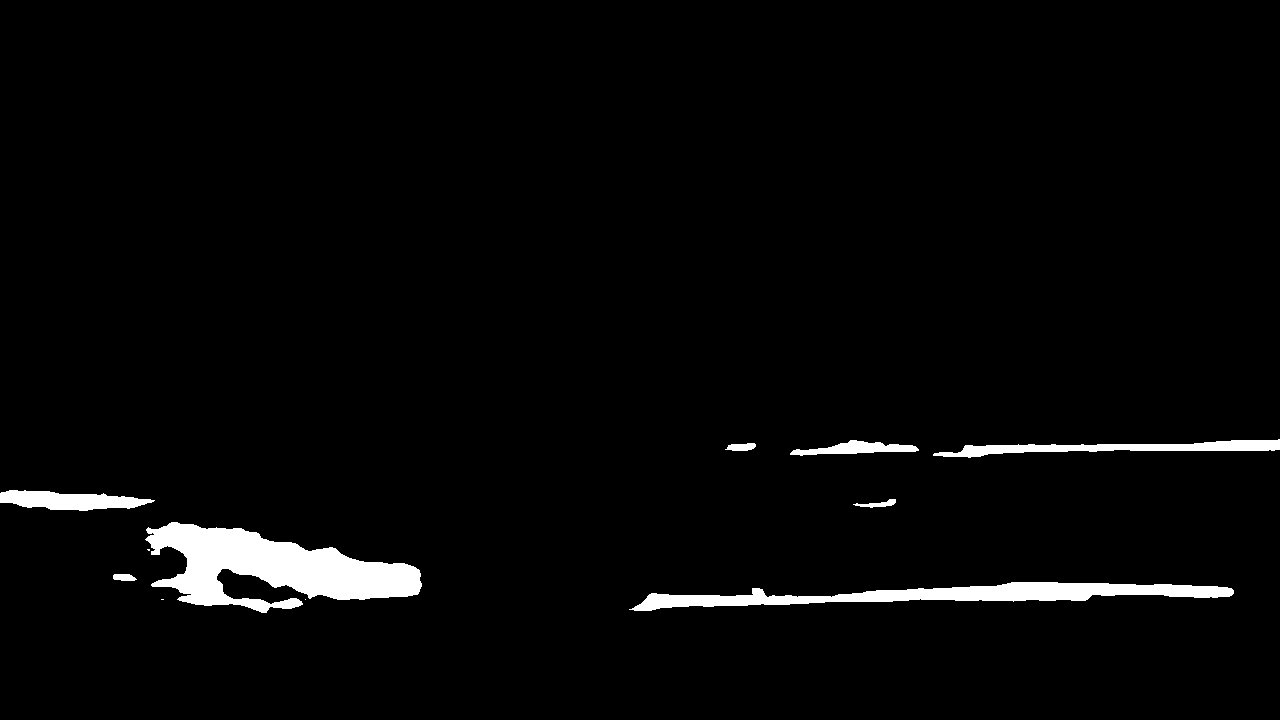}
\includegraphics[width=0.1\textwidth]{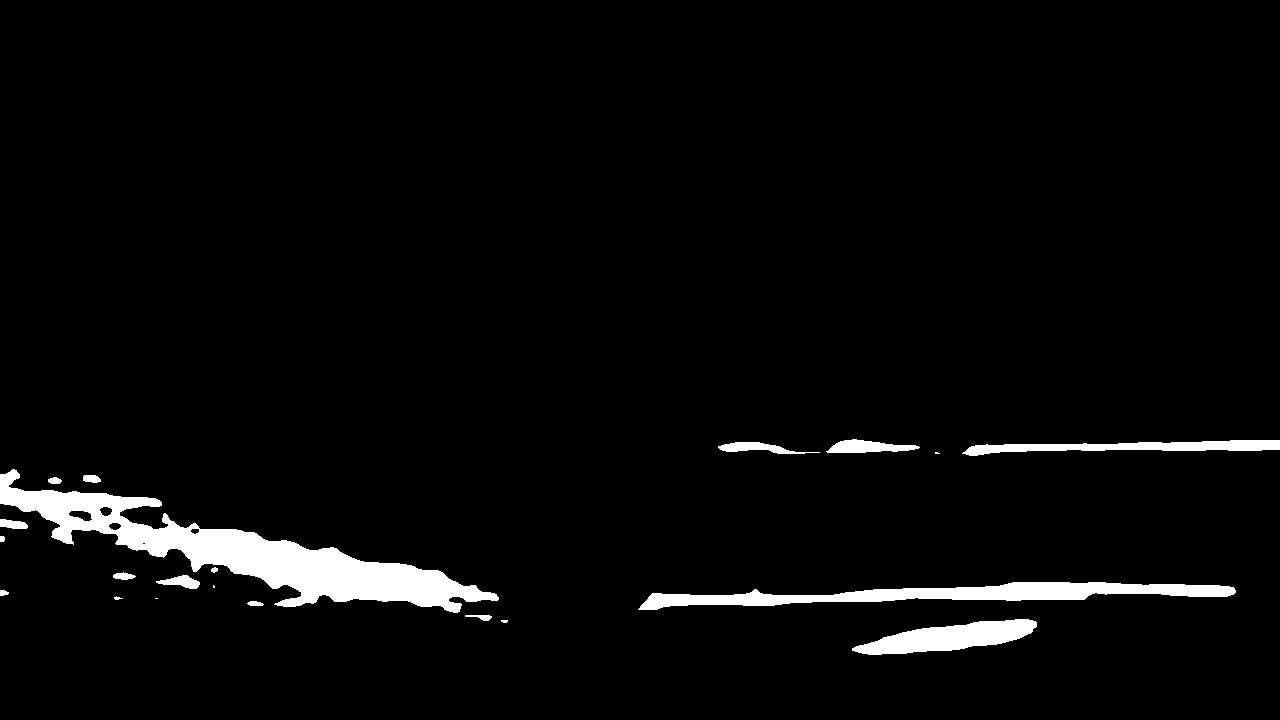}
\includegraphics[width=0.1\textwidth]{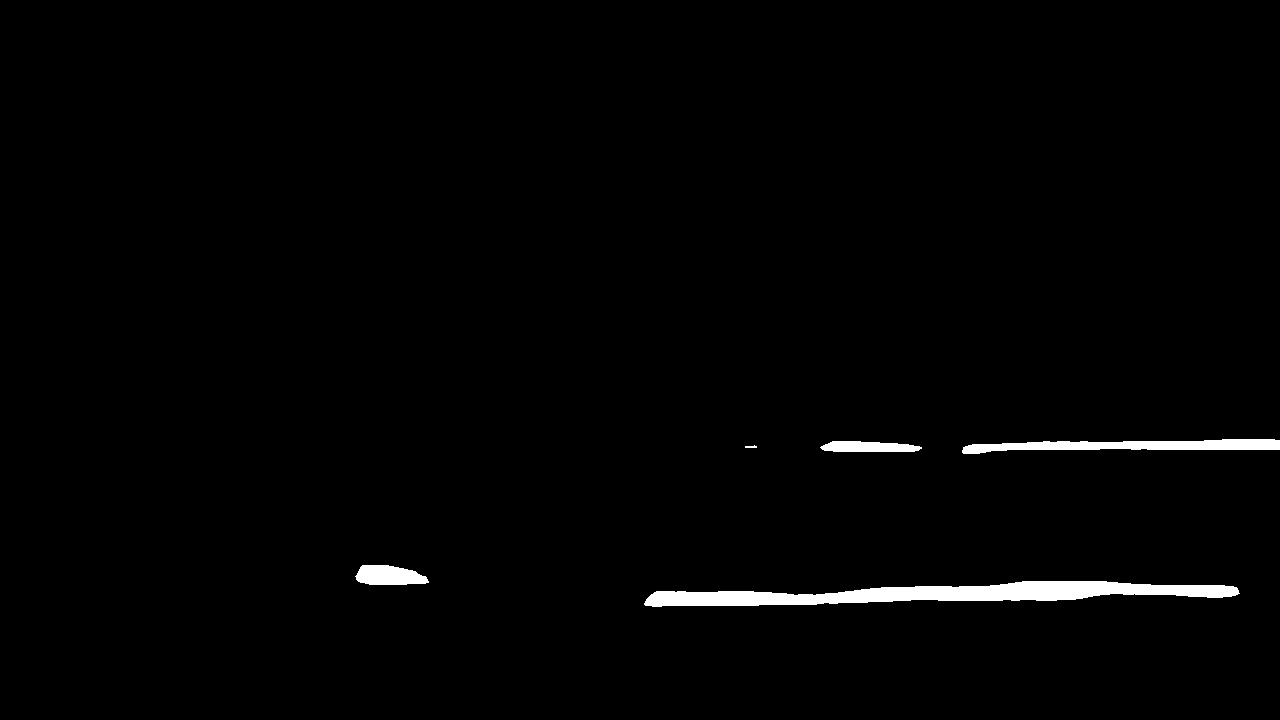}
\includegraphics[width=0.1\textwidth]{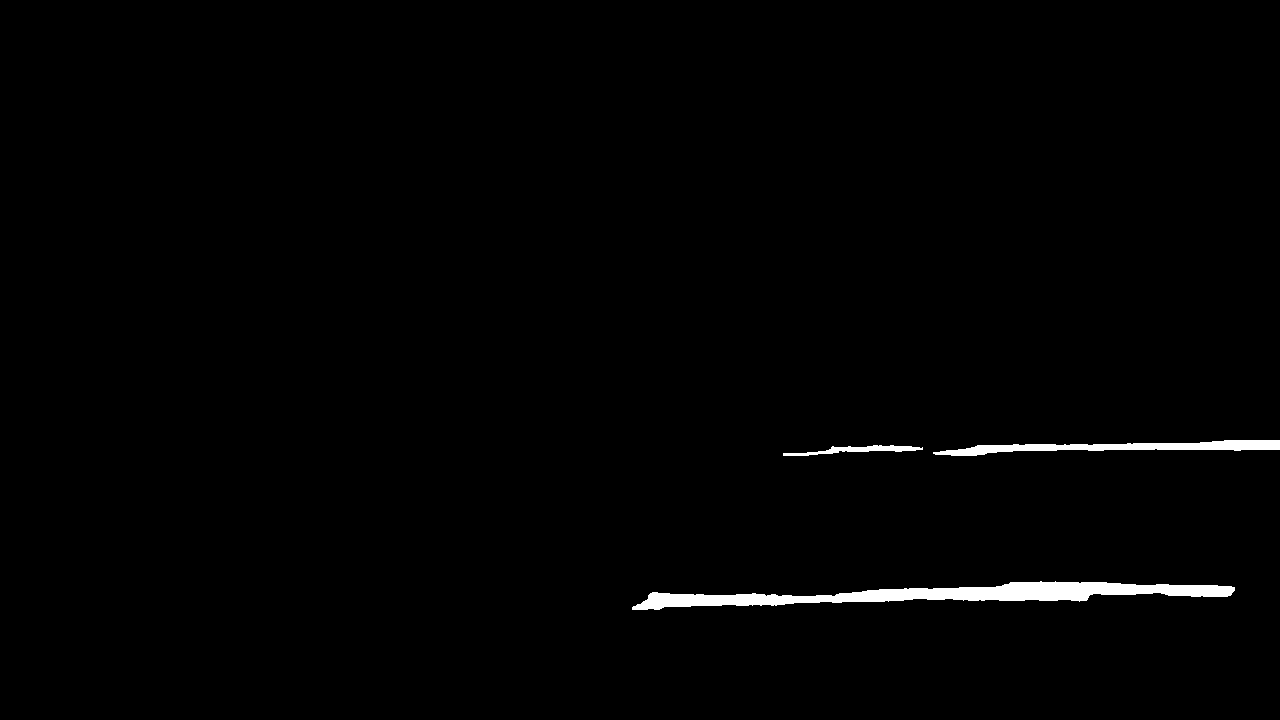}

\includegraphics[width=0.1\textwidth]{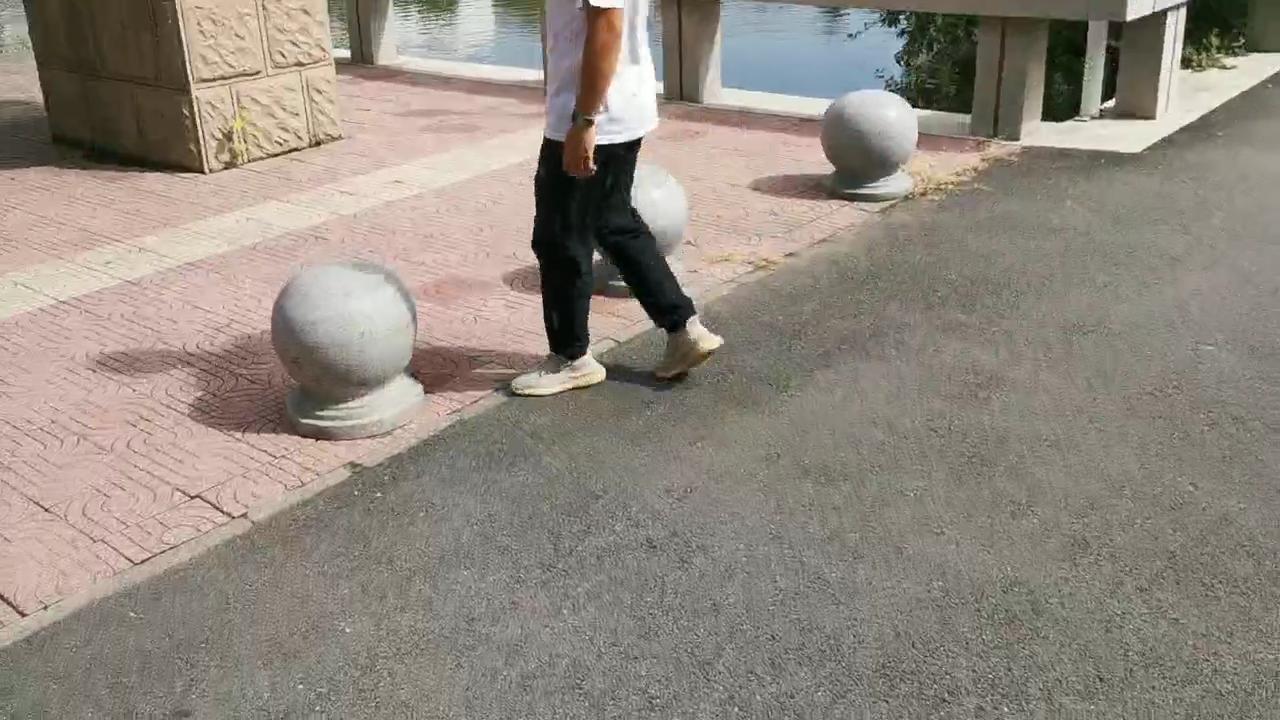}
\includegraphics[width=0.1\textwidth]{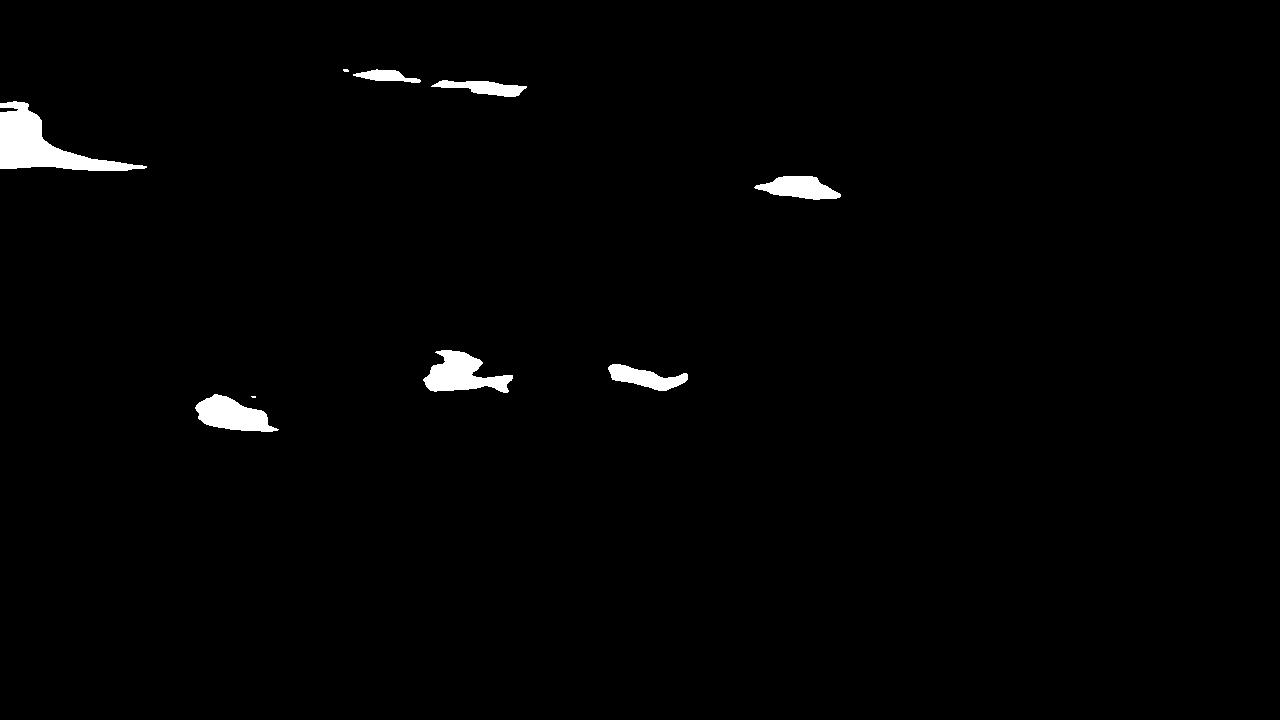}
\includegraphics[width=0.1\textwidth]{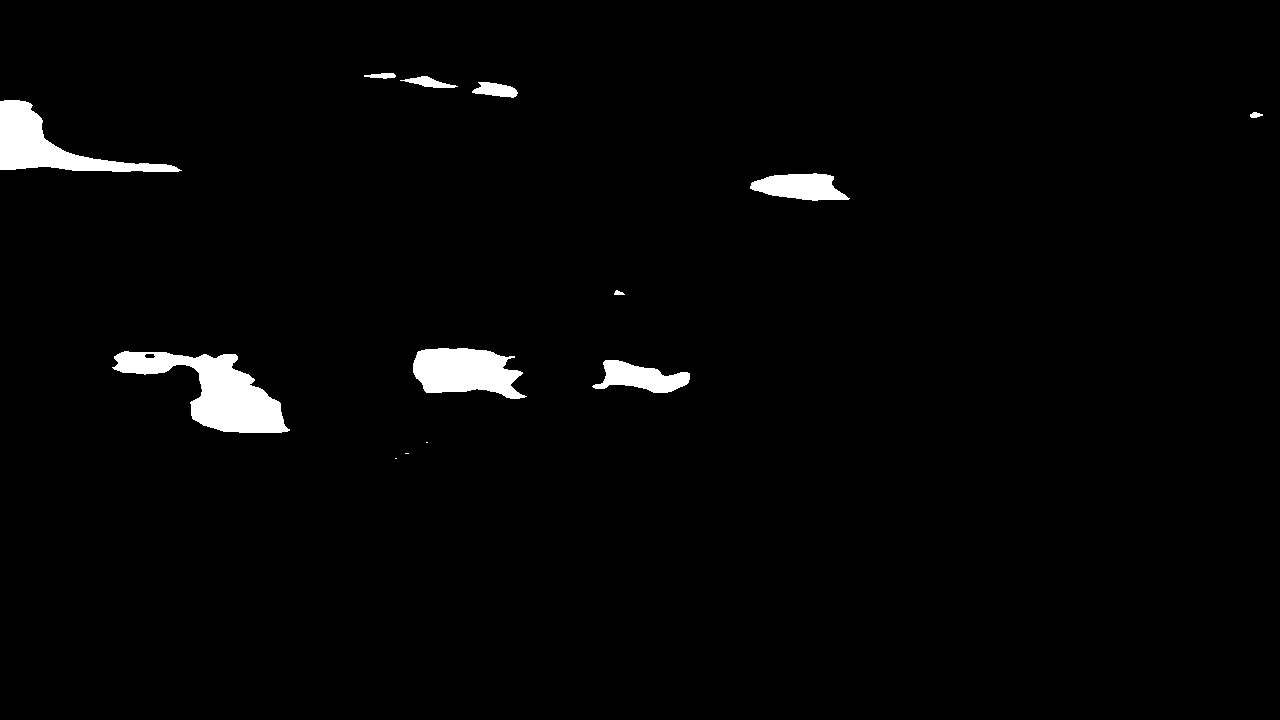}
\includegraphics[width=0.1\textwidth]{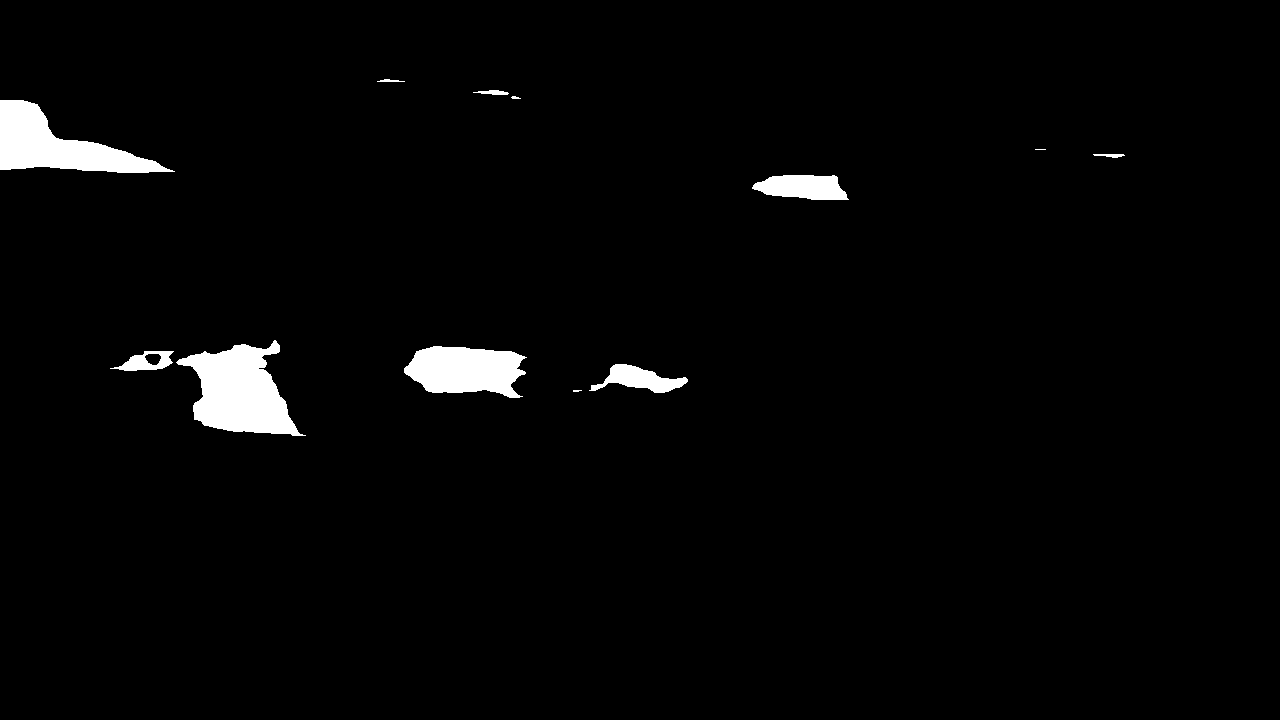}
\includegraphics[width=0.1\textwidth]{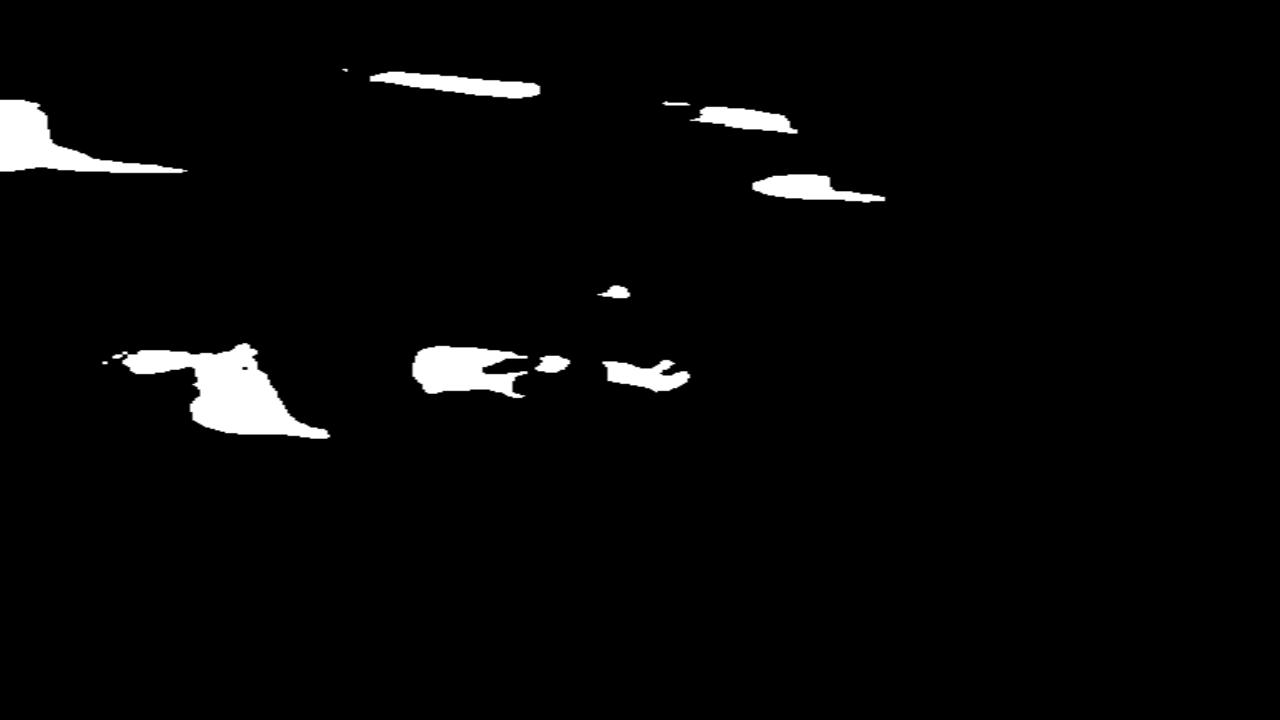}
\includegraphics[width=0.1\textwidth]{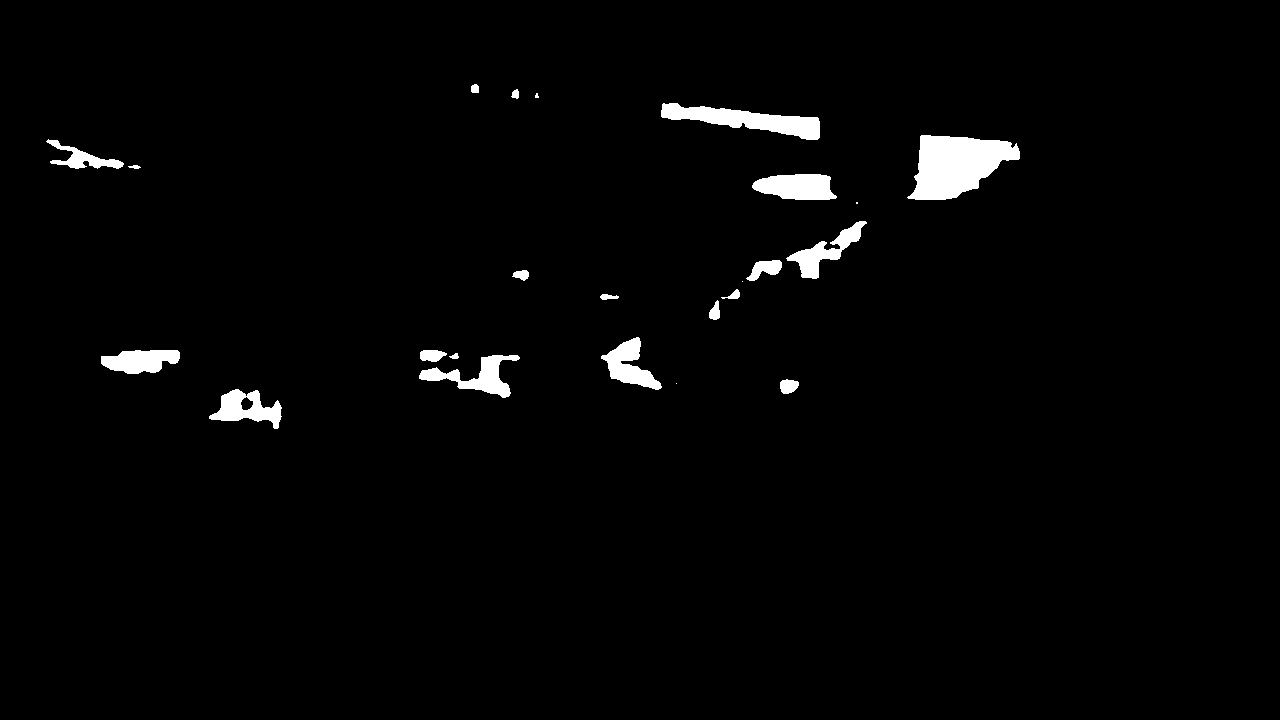}
\includegraphics[width=0.1\textwidth]{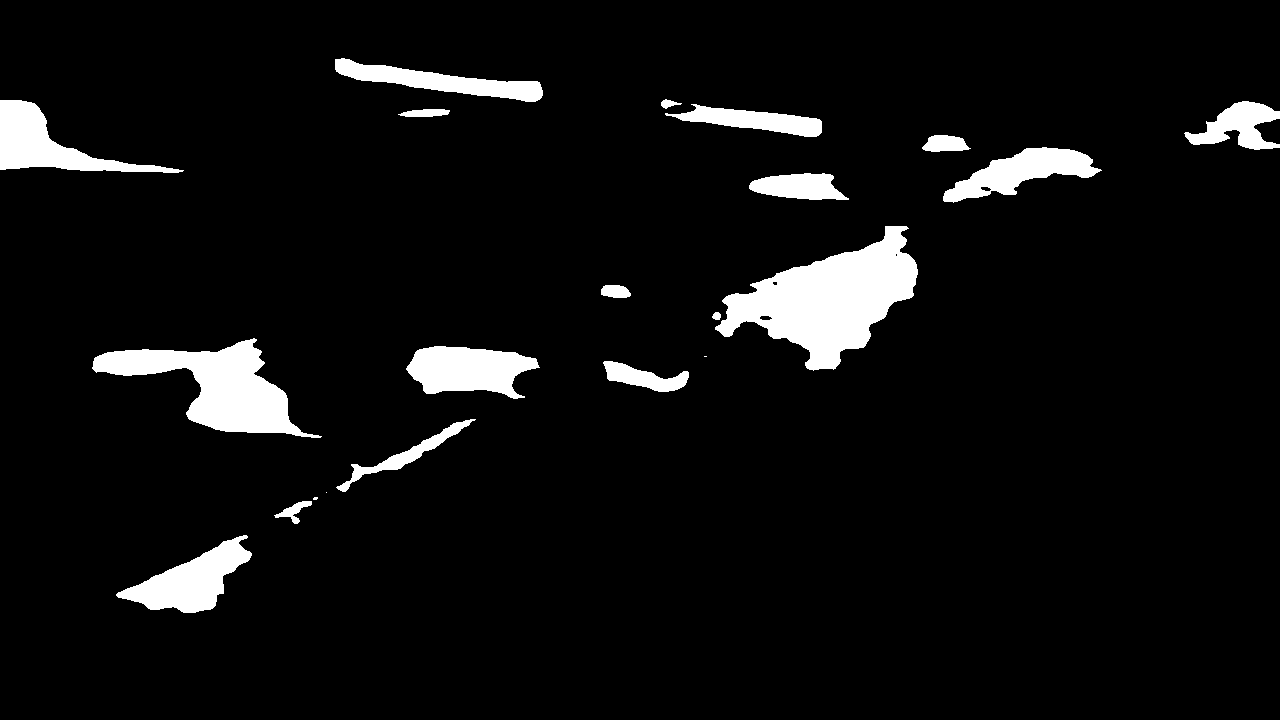}
\includegraphics[width=0.1\textwidth]{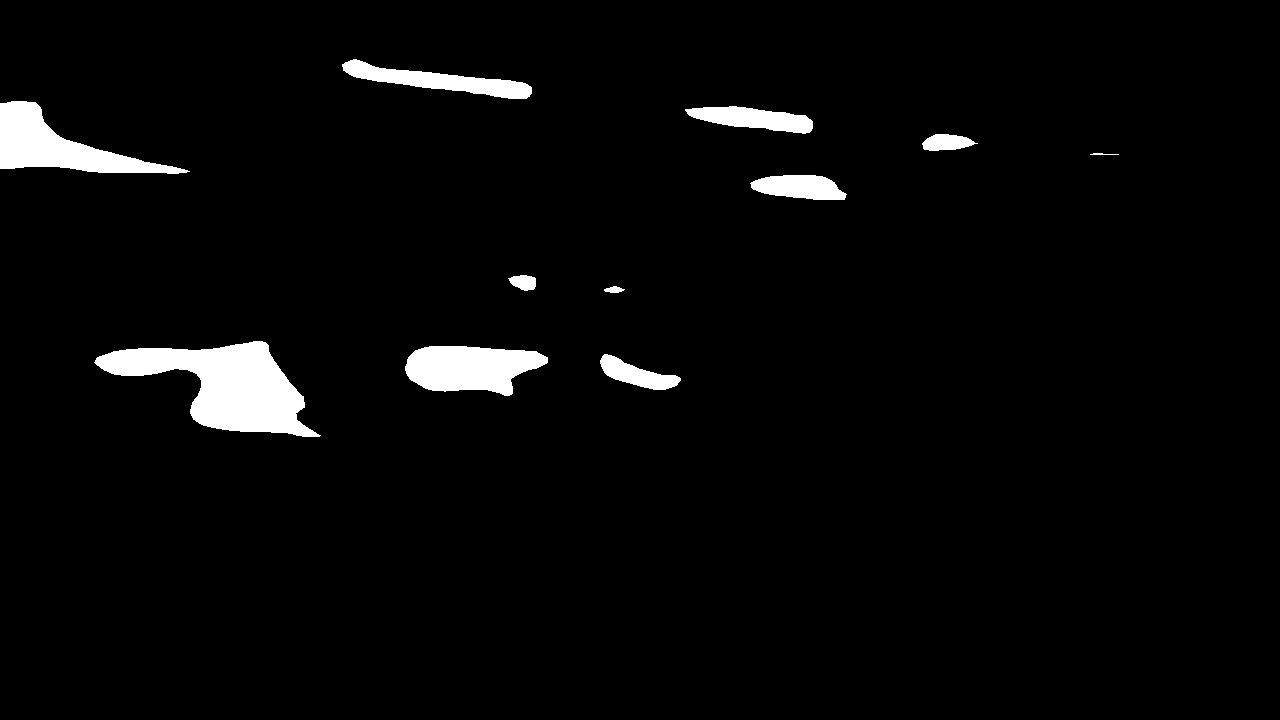}
\includegraphics[width=0.1\textwidth]{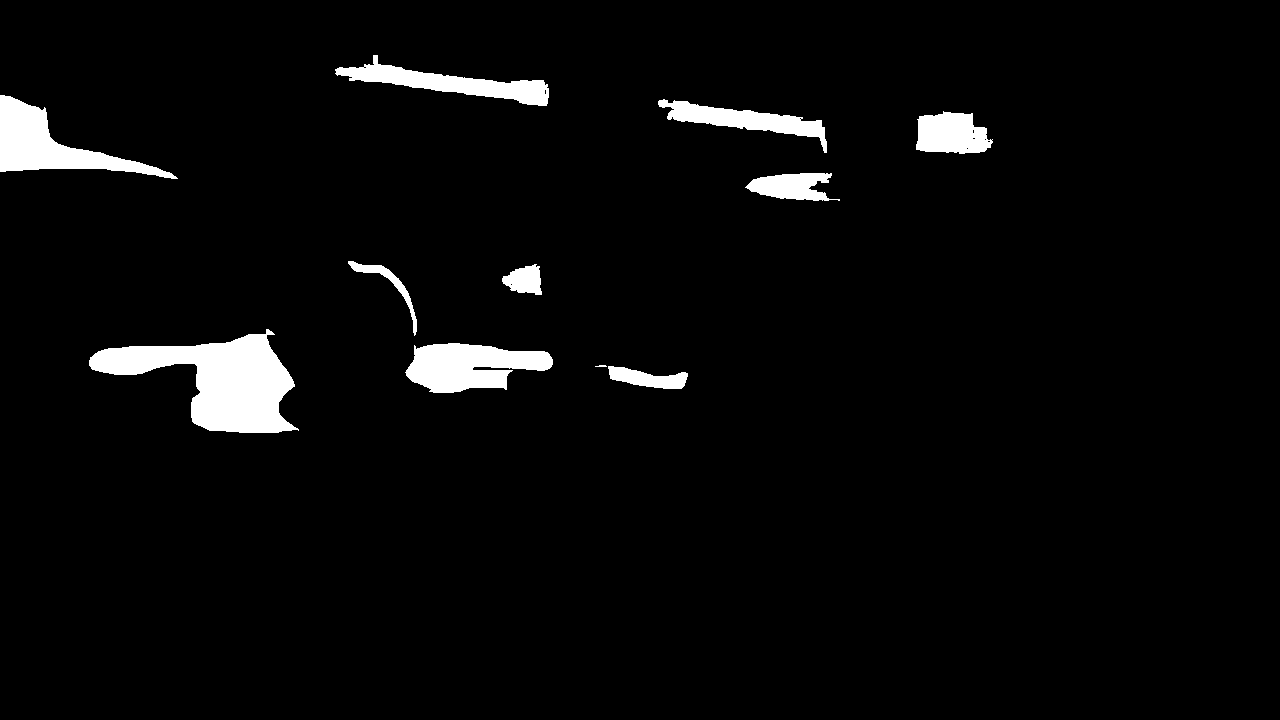}

\includegraphics[width=0.1\textwidth]{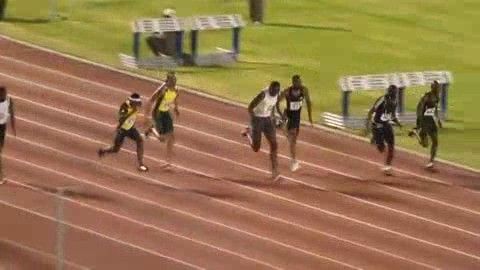}
\includegraphics[width=0.1\textwidth]{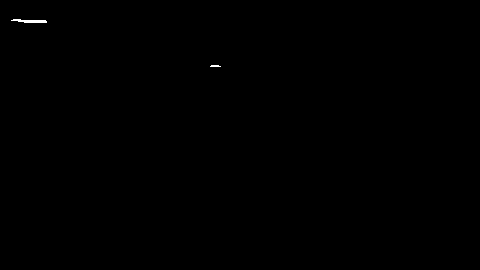}
\includegraphics[width=0.1\textwidth]{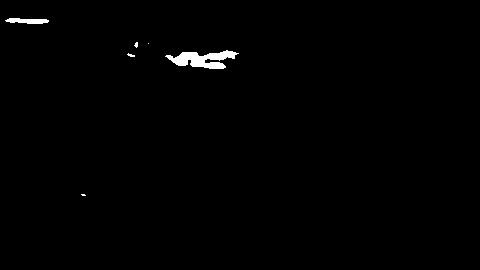}
\includegraphics[width=0.1\textwidth]{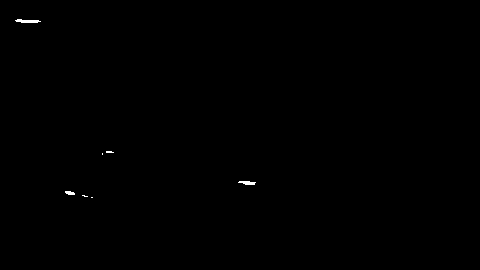}
\includegraphics[width=0.1\textwidth]{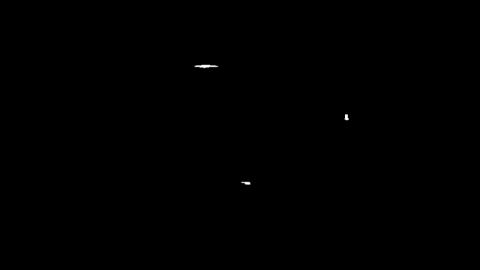}
\includegraphics[width=0.1\textwidth]{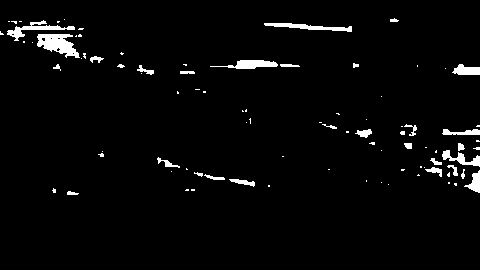}
\includegraphics[width=0.1\textwidth]{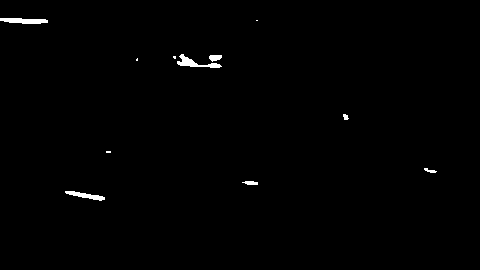}
\includegraphics[width=0.1\textwidth]{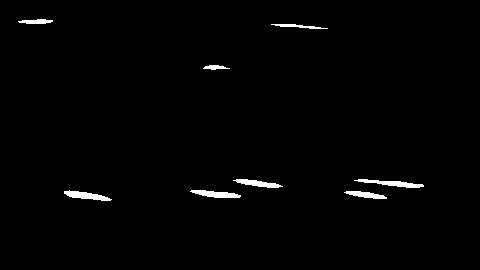}
\includegraphics[width=0.1\textwidth]{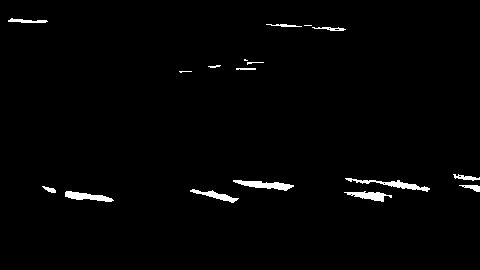}


\begin{minipage}{0.1\textwidth}
\centering
\footnotesize(a) Input
\end{minipage}
\begin{minipage}{0.1\textwidth}
\centering
\footnotesize(b) DDP
\end{minipage}
\begin{minipage}{0.1\textwidth}
\centering
\footnotesize(c) Pix2Seq
\end{minipage}
\begin{minipage}{0.1\textwidth}
\centering
\footnotesize(d) SILT
\end{minipage}
\begin{minipage}{0.1\textwidth}
\centering
\footnotesize(e) SCOTCH \& SODA
\end{minipage}
\begin{minipage}{0.1\textwidth}
\centering
\footnotesize(f) DAS
\end{minipage}
\begin{minipage}{0.1\textwidth}
\centering
\footnotesize(g) TBGDiff
\end{minipage}
\begin{minipage}{0.1\textwidth}
\centering
\footnotesize(h) Ours
\end{minipage}
\begin{minipage}{0.1\textwidth}
\centering
\footnotesize(i) GT
\end{minipage}
\caption{Qualitative comparison results of state-of-the-art methods. In comparison to other methods, our results exhibit less noise, and the predictions for shadow boundaries are more accurate. (b-d) are the best methods in IOS, ISD, and VOS in Table~\ref{tab:res}, and (e-g) are the latest networks in VSD.}
\label{fig:res1}
\end{figure*}

\subsection{Ablation Studies}

In this section, we first conduct ablation studies on the ViSha and CVSD dataset to demonstrate the effectiveness of our proposed modules (see \textbf{supplementary file} for more losses and tokenized temporal modeling ablations). We remove all proposed modules and retaining only the decoder for generating masks to build the baseline model. The experimental results of the baseline model are presented in Table~\ref{tab:ablation}(a). 

\subsubsection{Comparisons on Efficiency and Parameter}
As shown in Table~\ref{tab:ef}, we compare the parameter count and speed of recent video shadow detection networks on the Visha dataset. 
Since we freeze most of the parameters, the total number of parameters is provided in parentheses, with the number of learnable parameters outside the parentheses. DTTNet is able to reach the state-of-the-art performance and the faster inference speed. Notably, when using Automatic mixed precision~\cite{micikevicius2017mixed}, it can get real-time efficiency up to 26.63 fps.

\subsubsection{Effectiveness of Vision-language Match Module}
The Vision-language Match Module (VMM) is designed to align CLIP's text and image features, bridging the semantic gap between textual priors and visual content for shadow and dark regions. As shown in Table \ref{tab:ablation}(b), removing VMM leads to a significant performance drop: IoU decreases by 0.015 and F1-score drops by 0.01 compared to the the model with both VMM and DSB. This validates the necessity of VMM in leveraging CLIP's zero-shot capability to introduce semantic priors. The degradation confirms the critical role of VMM in enabling cross-modal knowledge transfer. Without VMM, the model loses the ability to effectively fuse text features ($P_{ts}, P_{td}$) with image features ($P_x$), resulting in misalignment between semantic guidance and visual content. The cross-attention mechanism of VMM can match text features to corresponding visual information, thereby reducing the differences between modalities and enabling the DSB to better fuse text priors into the visual features of the encoder.

\subsubsection{Effectiveness of Dark-aware Semantic Block}
The Dark-aware Semantic Block (DSB) integrates textual context into visual features and employs penumbra-aware supervision to prioritize shadow semantic learning. It works with VMM to help the model to learn more shadow information from linguistic priors. Removing DSB and VMM from the full model results in a 0.82 BER increase and 0.016 F1-score drop, as Table \ref{tab:ablation}(c) shows, demonstrating its key role in refining semantic representation. Since the DSB is designed to acquire semantics from priors rather than refine the original image features for primary pixels and the decreased supervision in the edge (the penumbra area), its impact on the IoU score is relatively minor.

\subsubsection{Effectiveness of Tokenized Temporal Block.}
The Tokenized Temporal Block (TTB) decouples temporal modeling from spatial features via learnable tokens, enabling efficient aggregation of cross-frame shadow patterns. As shown in Table \ref{tab:ablation}(d), removing TTB results in a significant performance drop, with a 0.032 IoU decrease and 3.9\% F1-score reduction. This highlights the importance of temporal feature modeling in Video Shadow Detection. Without TTB, the baseline model relies on naive frame-wise feature concatenation, which struggles to capture temporal correlations. When adding TTB to baseline model, the performance can be enhanced greatly. TTB's token-based summarization effectively distills temporal invariants into compact tokens, which are then spatially injected to enhance frame-level features. As it avoids redundant cross-frame attention operations, the computation cost is relatively small. As Table~\ref{tab:ef} shows, although we employ TTB in every layer of the encoder, the model can get over 26.63 fps in a single RTX 3090 when using automatic mixed precision. 

\section{Conclusion}

In this paper, we present DTTNet, a novel framework that integrates dark-aware linguistic guidance with tokenized temporal modeling. Specifically, the Tokenized Temporal Block (TTB) efficiently capture shadow dynamics via compact tokens to enhance cross-frame coherence. The Vision-language Match Module (VMM) leverages textual priors to explicitly guide attention toward shadows within dark regions, complemented by the Dark-aware Semantic Block (DSB) for adaptive feature weighting. Additionally, we reweight the penumbra regions for supervision to help the network focus on shadow body in the early stage. Experiments show that each component is effective and our approach can reach state-of-the-art results.

\section*{Acknowledgments}
This work was supported in part by the National Natural Science Foundation of China under Grant Nos. 62172417, 62272461, 62472424 and 62476189.

\bibliography{aaai2026}

\begin{thebibliography}{39}
\providecommand{\natexlab}[1]{#1}

\bibitem[{Adams et~al.(2022)Adams, Stefanucci, Creem-Regehr, and Bodenheimer}]{adams2022depth}
Adams, H.; Stefanucci, J.; Creem-Regehr, S.; and Bodenheimer, B. 2022.
\newblock Depth perception in augmented reality: The effects of display, shadow, and position.
\newblock In \emph{2022 IEEE conference on virtual reality and 3D user interfaces (VR)}, 792--801. IEEE.

\bibitem[{Adams et~al.(2021)Adams, Stefanucci, Creem-Regehr, Pointon, Thompson, and Bodenheimer}]{adams2021shedding}
Adams, H.; Stefanucci, J.; Creem-Regehr, S.; Pointon, G.; Thompson, W.; and Bodenheimer, B. 2021.
\newblock Shedding light on cast shadows: An investigation of perceived ground contact in ar and vr.
\newblock \emph{IEEE transactions on visualization and computer graphics}, 28(12): 4624--4639.

\bibitem[{Chen et~al.(2023)Chen, Li, Saxena, Hinton, and Fleet}]{chen2023generalist}
Chen, T.; Li, L.; Saxena, S.; Hinton, G.; and Fleet, D.~J. 2023.
\newblock A generalist framework for panoptic segmentation of images and videos.
\newblock In \emph{ICCV}, 909--919.

\bibitem[{Chen et~al.(2021)Chen, Wan, Zhu, Shen, Fu, Liu, and Qin}]{chen2021triple}
Chen, Z.; Wan, L.; Zhu, L.; Shen, J.; Fu, H.; Liu, W.; and Qin, J. 2021.
\newblock Triple-cooperative video shadow detection.
\newblock In \emph{CVPR}, 2715--2724.

\bibitem[{Chen et~al.(2020)Chen, Zhu, Wan, Wang, Feng, and Heng}]{chen2020multi}
Chen, Z.; Zhu, L.; Wan, L.; Wang, S.; Feng, W.; and Heng, P.-A. 2020.
\newblock A multi-task mean teacher for semi-supervised shadow detection.
\newblock In \emph{CVPR}, 5611--5620.

\bibitem[{Cheng, Tai, and Tang(2021)}]{cheng2021rethinking}
Cheng, H.~K.; Tai, Y.-W.; and Tang, C.-K. 2021.
\newblock Rethinking space-time networks with improved memory coverage for efficient video object segmentation.
\newblock \emph{NeurIPS}, 34: 11781--11794.

\bibitem[{Cong et~al.(2023)Cong, Guan, Chen, Zhang, Zhao, and Kwong}]{cong2023sddnet}
Cong, R.; Guan, Y.; Chen, J.; Zhang, W.; Zhao, Y.; and Kwong, S. 2023.
\newblock Sddnet: Style-guided dual-layer disentanglement network for shadow detection.
\newblock In \emph{ACM MM}, 1202--1211.

\bibitem[{Contributors(2020)}]{contributors2020mmsegmentation}
Contributors, M. 2020.
\newblock MMSegmentation: Openmmlab semantic segmentation toolbox and benchmark.

\bibitem[{Cucchiara et~al.(2003)Cucchiara, Grana, Piccardi, and Prati}]{cucchiara2003detecting}
Cucchiara, R.; Grana, C.; Piccardi, M.; and Prati, A. 2003.
\newblock Detecting moving objects, ghosts, and shadows in video streams.
\newblock \emph{IEEE transactions on pattern analysis and machine intelligence}, 25(10): 1337--1342.

\bibitem[{Deng et~al.(2018)Deng, Hu, Zhu, Xu, Qin, Han, and Heng}]{deng2018r3net}
Deng, Z.; Hu, X.; Zhu, L.; Xu, X.; Qin, J.; Han, G.; and Heng, P.-A. 2018.
\newblock R3net: Recurrent residual refinement network for saliency detection.
\newblock In \emph{IJCAI}, volume 684690. AAAI Press Menlo Park, CA, USA.

\bibitem[{Ding et~al.(2022)Ding, Yang, Hu, and Li}]{ding2022learning}
Ding, X.; Yang, J.; Hu, X.; and Li, X. 2022.
\newblock Learning shadow correspondence for video shadow detection.
\newblock In \emph{ECCV}, 705--722. Springer.

\bibitem[{Duan et~al.(2024)Duan, Cao, Zhu, Fu, Wang, Zhang, and Li}]{duan2024two}
Duan, X.; Cao, Y.; Zhu, L.; Fu, G.; Wang, X.; Zhang, R.; and Li, P. 2024.
\newblock Two-stage video shadow detection via temporal-spatial adaption.
\newblock In \emph{European Conference on Computer Vision}, 196--214. Springer.

\bibitem[{Guan, Xu, and Lau(2024)}]{guan2024delving}
Guan, H.; Xu, K.; and Lau, R.~W. 2024.
\newblock Delving into dark regions for robust shadow detection.
\newblock \emph{arXiv preprint arXiv:2402.13631}.

\bibitem[{Hu et~al.(2021)Hu, Wang, Fu, Jiang, Wang, and Heng}]{hu2021revisiting}
Hu, X.; Wang, T.; Fu, C.-W.; Jiang, Y.; Wang, Q.; and Heng, P.-A. 2021.
\newblock Revisiting shadow detection: A new benchmark dataset for complex world.
\newblock \emph{IEEE TIP}, 30: 1925--1934.

\bibitem[{Ji et~al.(2023)Ji, Chen, Xie, Hong, Liu, Liu, Lu, Li, and Luo}]{ji2023ddp}
Ji, Y.; Chen, Z.; Xie, E.; Hong, L.; Liu, X.; Liu, Z.; Lu, T.; Li, Z.; and Luo, P. 2023.
\newblock Ddp: Diffusion model for dense visual prediction.
\newblock In \emph{ICCV}, 21741--21752.

\bibitem[{Karsch et~al.(2011)Karsch, Hedau, Forsyth, and Hoiem}]{karsch2011rendering}
Karsch, K.; Hedau, V.; Forsyth, D.; and Hoiem, D. 2011.
\newblock Rendering synthetic objects into legacy photographs.
\newblock \emph{ACM Transactions on graphics (TOG)}, 30(6): 1--12.

\bibitem[{Khan et~al.(2014)Khan, Bennamoun, Sohel, and Togneri}]{khan2014automatic}
Khan, S.~H.; Bennamoun, M.; Sohel, F.; and Togneri, R. 2014.
\newblock Automatic feature learning for robust shadow detection.
\newblock In \emph{CVPR}, 1939--1946. IEEE.

\bibitem[{Kirillov et~al.(2023)Kirillov, Mintun, Ravi, Mao, Rolland, Gustafson, Xiao, Whitehead, Berg, Lo et~al.}]{kirillov2023segment}
Kirillov, A.; Mintun, E.; Ravi, N.; Mao, H.; Rolland, C.; Gustafson, L.; Xiao, T.; Whitehead, S.; Berg, A.~C.; Lo, W.-Y.; et~al. 2023.
\newblock Segment anything.
\newblock In \emph{ICCV}, 4015--4026.

\bibitem[{Lalonde and Matthews(2014)}]{lalonde2014lighting}
Lalonde, J.-F.; and Matthews, I. 2014.
\newblock Lighting estimation in outdoor image collections.
\newblock In \emph{2014 2nd international conference on 3D vision}, volume~1, 131--138. IEEE.

\bibitem[{Lin et~al.(2017)Lin, Doll{\'a}r, Girshick, He, Hariharan, and Belongie}]{lin2017feature}
Lin, T.-Y.; Doll{\'a}r, P.; Girshick, R.; He, K.; Hariharan, B.; and Belongie, S. 2017.
\newblock Feature pyramid networks for object detection.
\newblock In \emph{CVPR}, 2117--2125.

\bibitem[{Liu et~al.(2023{\natexlab{a}})Liu, Prost, Zhu, Papadakis, Li{\`o}, Sch{\"o}nlieb, and Aviles-Rivero}]{liu2023scotch}
Liu, L.; Prost, J.; Zhu, L.; Papadakis, N.; Li{\`o}, P.; Sch{\"o}nlieb, C.-B.; and Aviles-Rivero, A.~I. 2023{\natexlab{a}}.
\newblock Scotch and soda: A transformer video shadow detection framework.
\newblock In \emph{CVPR}, 10449--10458.

\bibitem[{Liu et~al.(2023{\natexlab{b}})Liu, Lu, Fu, and Cao}]{liu2023learning}
Liu, W.; Lu, H.; Fu, H.; and Cao, Z. 2023{\natexlab{b}}.
\newblock Learning to upsample by learning to sample.
\newblock In \emph{ICCV}, 6027--6037.

\bibitem[{Lu et~al.(2022)Lu, Cao, Liu, Long, Chen, Zhou, Yang, and Xiao}]{lu2022video}
Lu, X.; Cao, Y.; Liu, S.; Long, C.; Chen, Z.; Zhou, X.; Yang, Y.; and Xiao, C. 2022.
\newblock Video shadow detection via spatio-temporal interpolation consistency training.
\newblock In \emph{CVPR}, 3116--3125.

\bibitem[{Lu et~al.(2019)Lu, Wang, Ma, Shen, Shao, and Porikli}]{lu2019see}
Lu, X.; Wang, W.; Ma, C.; Shen, J.; Shao, L.; and Porikli, F. 2019.
\newblock See more, know more: Unsupervised video object segmentation with co-attention siamese networks.
\newblock In \emph{CVPR}, 3623--3632.

\bibitem[{Micikevicius et~al.(2017)Micikevicius, Narang, Alben, Diamos, Elsen, Garcia, Ginsburg, Houston, Kuchaiev, Venkatesh et~al.}]{micikevicius2017mixed}
Micikevicius, P.; Narang, S.; Alben, J.; Diamos, G.; Elsen, E.; Garcia, D.; Ginsburg, B.; Houston, M.; Kuchaiev, O.; Venkatesh, G.; et~al. 2017.
\newblock Mixed precision training.
\newblock \emph{arXiv preprint arXiv:1710.03740}.

\bibitem[{Milletari, Navab, and Ahmadi(2016)}]{milletari2016v}
Milletari, F.; Navab, N.; and Ahmadi, S.-A. 2016.
\newblock V-net: Fully convolutional neural networks for volumetric medical image segmentation.
\newblock In \emph{2016 fourth international conference on 3D vision (3DV)}, 565--571. Ieee.

\bibitem[{Oh et~al.(2019)Oh, Lee, Xu, and Kim}]{oh2019video}
Oh, S.~W.; Lee, J.-Y.; Xu, N.; and Kim, S.~J. 2019.
\newblock Video object segmentation using space-time memory networks.
\newblock In \emph{ICCV}, 9226--9235.

\bibitem[{Oquab et~al.(2023)Oquab, Darcet, Moutakanni, Vo, Szafraniec, Khalidov, Fernandez, Haziza, Massa, El-Nouby et~al.}]{oquab2023dinov2}
Oquab, M.; Darcet, T.; Moutakanni, T.; Vo, H.; Szafraniec, M.; Khalidov, V.; Fernandez, P.; Haziza, D.; Massa, F.; El-Nouby, A.; et~al. 2023.
\newblock Dinov2: Learning robust visual features without supervision.
\newblock \emph{arXiv preprint arXiv:2304.07193}.

\bibitem[{Shao, Taff, and Walsh(2011)}]{shao2011shadow}
Shao, Y.; Taff, G.~N.; and Walsh, S.~J. 2011.
\newblock Shadow detection and building-height estimation using IKONOS data.
\newblock \emph{International journal of remote sensing}, 32(22): 6929--6944.

\bibitem[{Voigtlaender et~al.(2019)Voigtlaender, Chai, Schroff, Adam, Leibe, and Chen}]{voigtlaender2019feelvos}
Voigtlaender, P.; Chai, Y.; Schroff, F.; Adam, H.; Leibe, B.; and Chen, L.-C. 2019.
\newblock Feelvos: Fast end-to-end embedding learning for video object segmentation.
\newblock In \emph{CVPR}, 9481--9490.

\bibitem[{Wang, Li, and Yang(2018)}]{wang2018stacked}
Wang, J.; Li, X.; and Yang, J. 2018.
\newblock Stacked conditional generative adversarial networks for jointly learning shadow detection and shadow removal.
\newblock In \emph{CVPR}, 1788--1797.

\bibitem[{Wang et~al.(2023)Wang, Zhou, Mao, and Li}]{wang2023detect}
Wang, Y.; Zhou, W.; Mao, Y.; and Li, H. 2023.
\newblock Detect any shadow: Segment anything for video shadow detection.
\newblock \emph{IEEE TCSVT}.

\bibitem[{Wei et~al.(2024)Wei, Xing, Liao, Zhang, and Liu}]{wei2024structure}
Wei, H.; Xing, G.; Liao, J.; Zhang, Y.; and Liu, Y. 2024.
\newblock Structure-aware spatial-temporal interaction network for video shadow detection.
\newblock In \emph{Proceedings of the Thirty-Third International Joint Conference on Artificial Intelligence, IJCAI-24, International Joint Conferences on Artificial Intelligence Organization}, 1425--1433.

\bibitem[{Xie et~al.(2021)Xie, Wang, Yu, Anandkumar, Alvarez, and Luo}]{xie2021segformer}
Xie, E.; Wang, W.; Yu, Z.; Anandkumar, A.; Alvarez, J.~M.; and Luo, P. 2021.
\newblock SegFormer: Simple and efficient design for semantic segmentation with transformers.
\newblock \emph{NeurIPS}, 34: 12077--12090.

\bibitem[{Yang et~al.(2023)Yang, Wang, Hu, and Fu}]{yang2023silt}
Yang, H.; Wang, T.; Hu, X.; and Fu, C.-W. 2023.
\newblock SILT: Shadow-aware Iterative Label Tuning for Learning to Detect Shadows from Noisy Labels.
\newblock In \emph{ICCV}, 12687--12698.

\bibitem[{Zheng et~al.(2019)Zheng, Qiao, Cao, and Lau}]{zheng2019distraction}
Zheng, Q.; Qiao, X.; Cao, Y.; and Lau, R.~W. 2019.
\newblock Distraction-aware shadow detection.
\newblock In \emph{CVPR}, 5167--5176.

\bibitem[{Zhou et~al.(2024)Zhou, Wang, Ye, Xing, Ma, Li, Wang, and Zhu}]{zhou2024timeline}
Zhou, H.; Wang, H.; Ye, T.; Xing, Z.; Ma, J.; Li, P.; Wang, Q.; and Zhu, L. 2024.
\newblock Timeline and Boundary Guided Diffusion Network for Video Shadow Detection.
\newblock In \emph{ACM MM}, 166--175.

\bibitem[{Zhu et~al.(2018)Zhu, Deng, Hu, Fu, Xu, Qin, and Heng}]{zhu2018bidirectional}
Zhu, L.; Deng, Z.; Hu, X.; Fu, C.-W.; Xu, X.; Qin, J.; and Heng, P.-A. 2018.
\newblock Bidirectional feature pyramid network with recurrent attention residual modules for shadow detection.
\newblock In \emph{ECCV}, 121--136.

\bibitem[{Zhu et~al.(2021)Zhu, Xu, Ke, and Lau}]{zhu2021mitigating}
Zhu, L.; Xu, K.; Ke, Z.; and Lau, R.~W. 2021.
\newblock Mitigating intensity bias in shadow detection via feature decomposition and reweighting.
\newblock In \emph{Proceedings of the IEEE/CVF International Conference on Computer Vision}, 4702--4711.

\end{thebibliography}

\end{document}


\section{Appendix}

To provide more comprehensive description of DTTNet, we include some supplementary information in this file.
\begin{itemize}
    \item Additional ablation studies.
    \item More qualitative comparisons.
    \item Core algorithm implementation.
    \item Video of the results of the proposed method.
\end{itemize}

\section{Ablation Studies}
\label{sec:ablation}
\subsection{Temporal Modeling Strategies}
Table~\ref{tab:pixel} presents results under different temporal modeling approaches. We conducted comparative experiments by replacing our proposed TTB with a layer of spatiotemporal self-attention for pixel-level temporal modeling and by removing the temporal module of DTTNet. It can be observed that spatiotemporal self-attention fails to effectively extract temporal information, leading to a performance degradation. Meanwhile, due to the quadratic complexity of self-attention, even with only one self-attention layer used per block, the training memory consumption increases significantly, from about 12, 000MB to 17, 000MB. Instead, the token-level fusion not only enhances the aggregation of multi-frame information but also effectively reduces computational overhead.
\begin{table}[h]
\setlength{\tabcolsep}{3mm}
\centering
\begin{tabular}{c|cccc}
\toprule
Methods & MAE$\downarrow$ & $F_{\beta}\uparrow$ & IoU$\uparrow$ & BER$\downarrow$ \\ 
\midrule
- & 0.021 & 0.817 & 0.686 & 7.98\\
Pixel & 0.019 & 0.833 & 0.685 & 8.32\\
\rowcolor{gray!17}
Tokenized & \textbf{0.016} & \textbf{0.849} & \textbf{0.718} & \textbf{6.45}\\
\bottomrule
\end{tabular}
\caption{Ablation study on pixel and tokenized temporal modeling. Experiments are conducted on the ViSha dataset.}
\label{tab:pixel}
\end{table}

\subsection{Loss functions} We conducted ablation experiments on three types of loss functions, with the results shown in Table~\ref{tab:loss}. We compared mask loss, edge loss, and semantic loss used in DTTNet. When all losses are used together, the model can achieve the best performance. When semantic loss is not used for supervision, a drop in performance can be observed. This indicates that raw text prior features cannot be directly obtained from the pre-trained CLIP model. Instead, the VLM and DSB require certain guidance to extract relevant features from text and images, thereby better enhancing the shadow regions. Furthermore, we find that it is necessary to supervise the network using edge masks. This helps the decoder handle shadow edges more meticulously, thereby improving the IoU score and making the detection more accurate.
\begin{table}[hbt]
\centering
\resizebox{1\linewidth}{!}{
\begin{tabular}{ccc|cccc}
\toprule
$\mathcal{L}_{mask}$ & $\mathcal{L}_{edge}$ & $\mathcal{L}_{sem}$ & MAE$\downarrow$ & $F_{\beta}\uparrow$ & IoU$\uparrow$ & BER$\downarrow$\\
\midrule
$\checkmark$ & - & - & 0.020 & 0.825 & 0.673 & 8.96 \\
$\checkmark$ & $\checkmark$ & - & 0.018 & 0.825 & 0.687 & 8.40 \\
\rowcolor{gray!17}
$\checkmark$ & $\checkmark$ & $\checkmark$ & \textbf{0.016} & \textbf{0.849} & \textbf{0.718} & \textbf{6.45} \\
\bottomrule
\end{tabular}
}
\caption{Ablation results of different loss functions. Experiments are conducted on the ViSha~\cite{chen2021triple} dataset.}
\label{tab:loss}
\end{table}

\begin{figure}[t]
\centering
\includegraphics[width=0.95\linewidth]{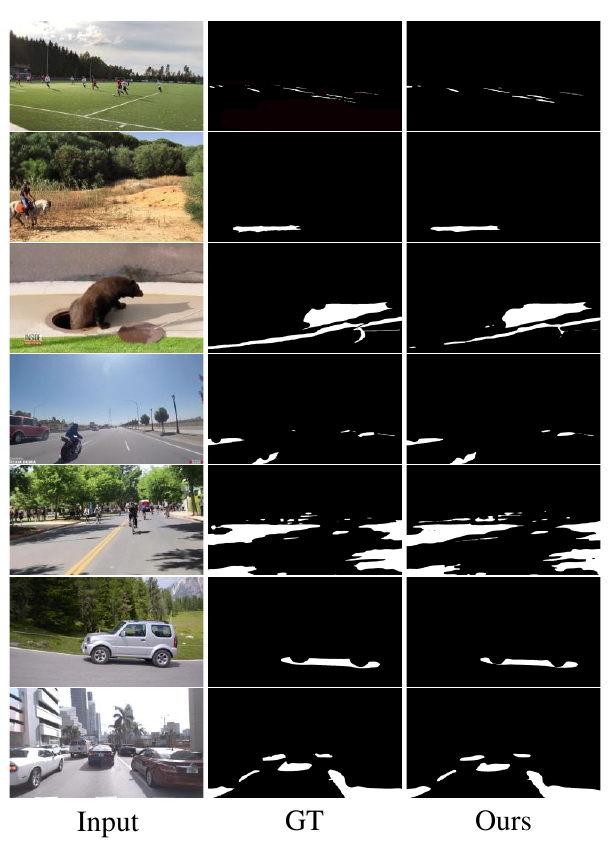}
\caption{Qualitative results of DTTNet on the CVSD~\cite{duan2024two}, a more complex dataset.}
\label{fig:longterm1}
\end{figure}

\section{More Qualitative Comparisons}
\label{sec:results}
\begin{figure*}[!t]
\centering

\includegraphics[width=0.1\textwidth]{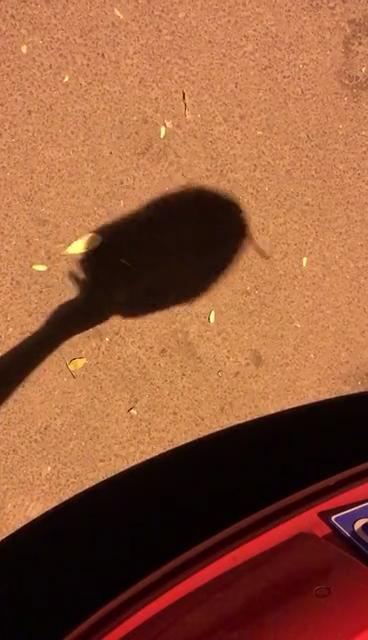}
\includegraphics[width=0.1\textwidth]{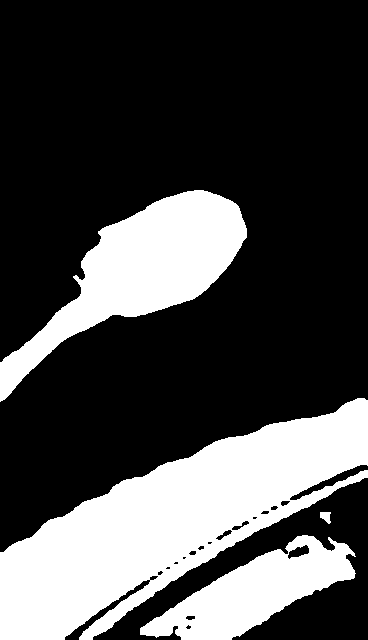}
\includegraphics[width=0.1\textwidth]{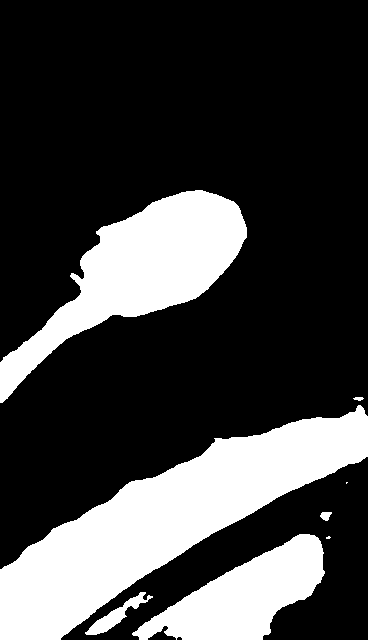}
\includegraphics[width=0.1\textwidth]{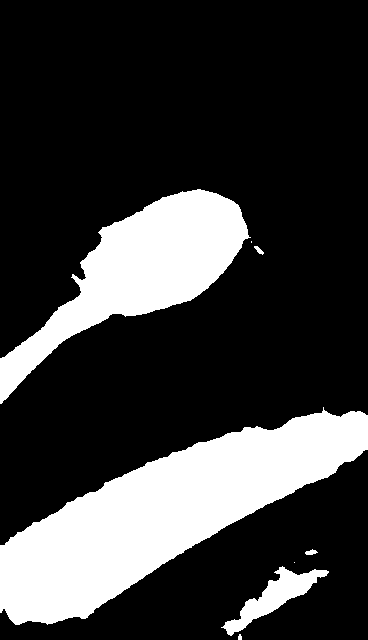}
\includegraphics[width=0.1\textwidth]{pictures/scotch&soda/cap_waving.jpg}
\includegraphics[width=0.1\textwidth]{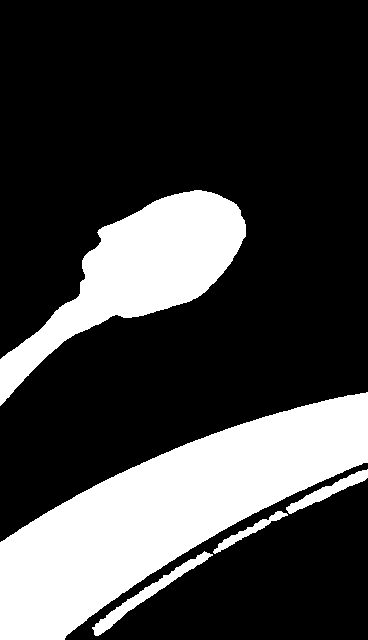}
\includegraphics[width=0.1\textwidth]{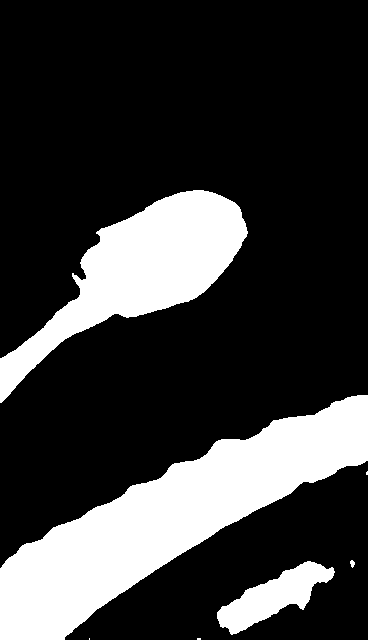}
\includegraphics[width=0.1\textwidth]{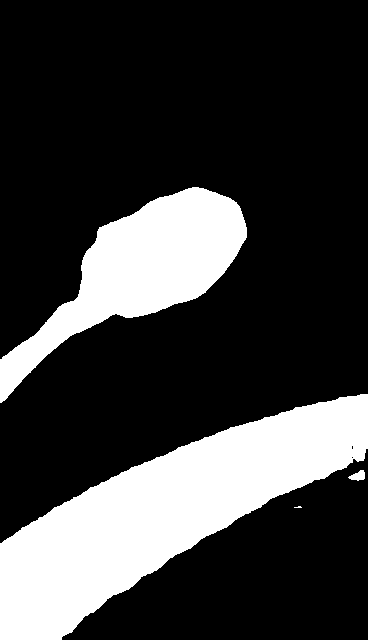}
\includegraphics[width=0.1\textwidth]{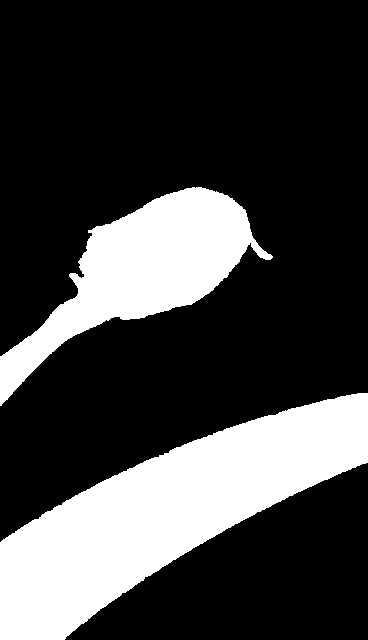}

\includegraphics[width=0.1\textwidth]{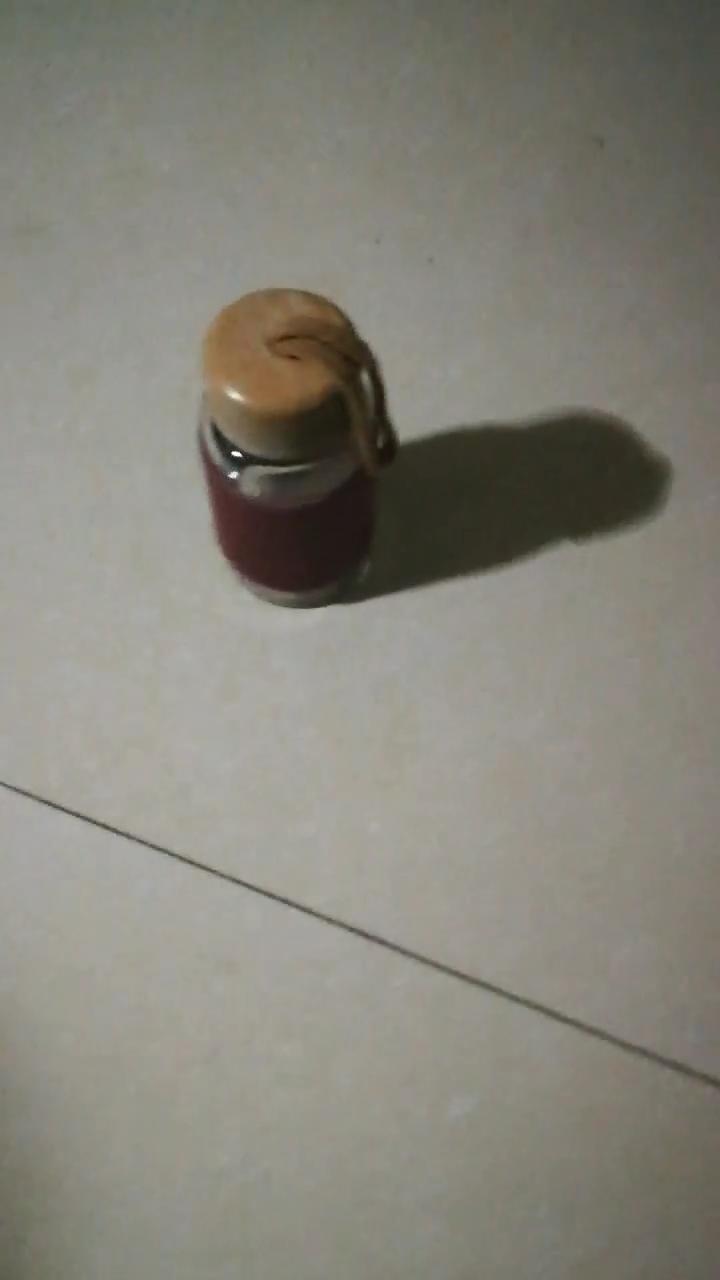}
    \includegraphics[width=0.1\textwidth]{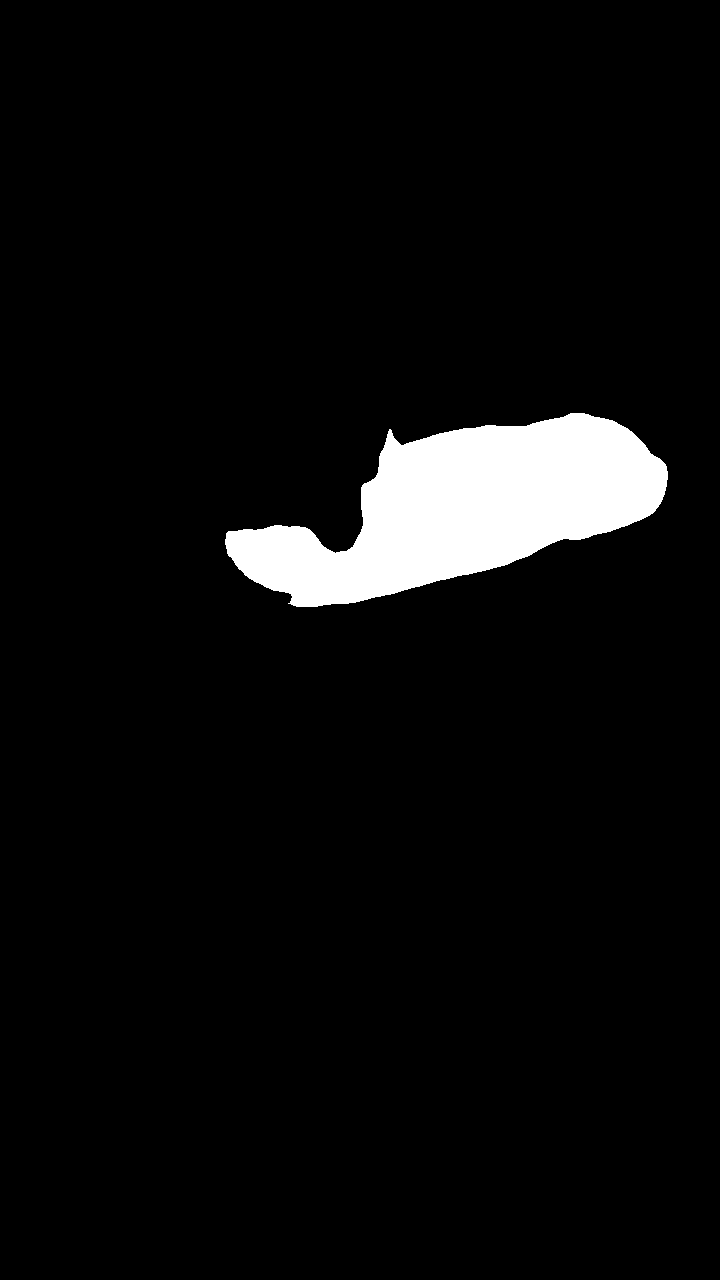}
\includegraphics[width=0.1\textwidth]{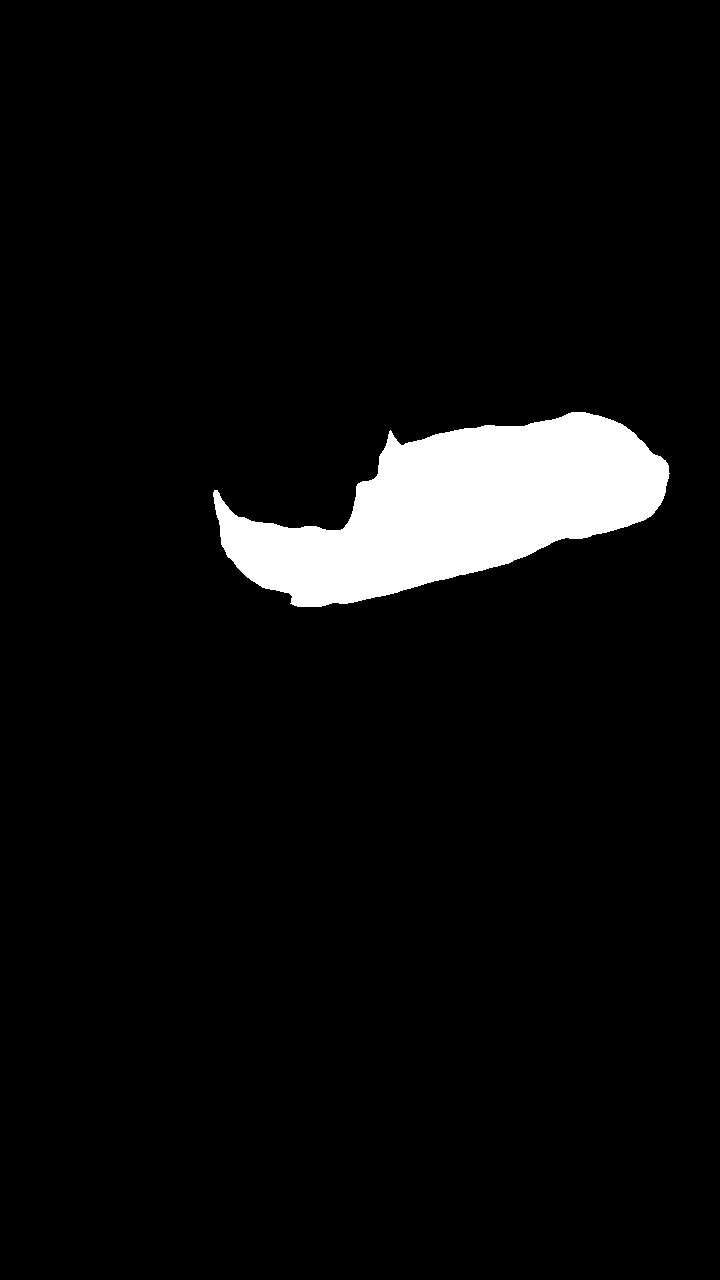}
\includegraphics[width=0.1\textwidth]{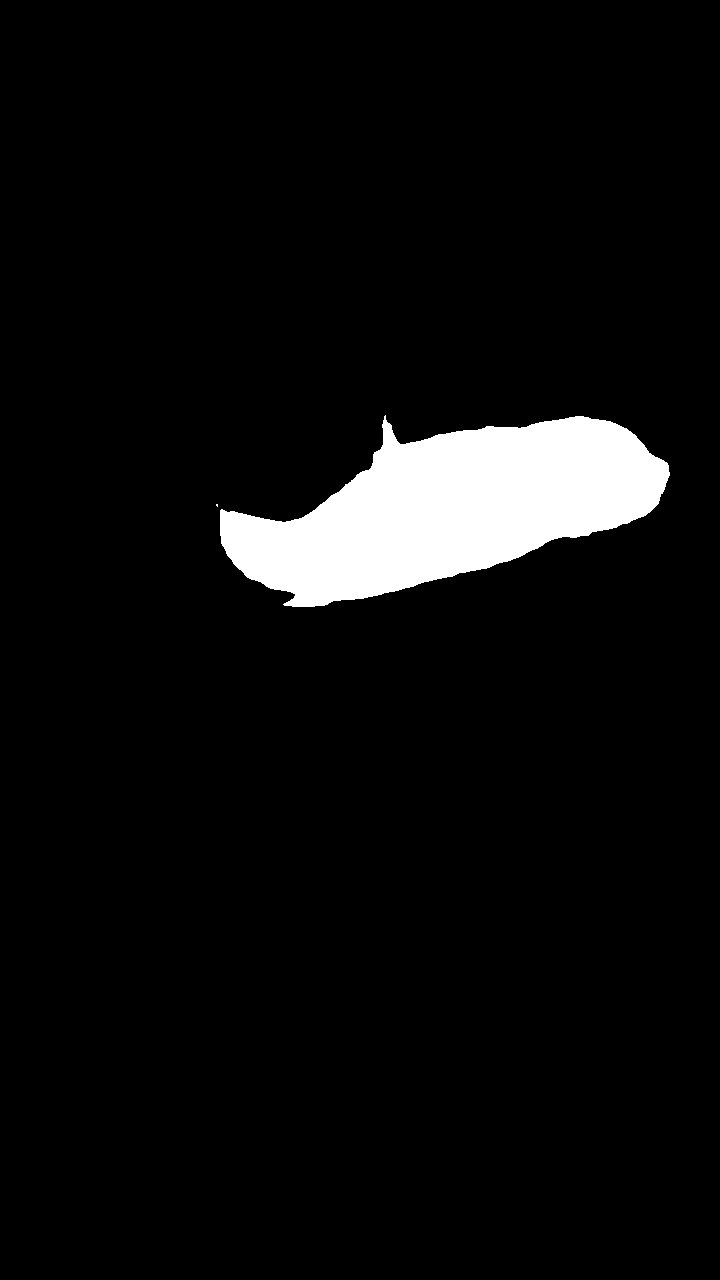}
\includegraphics[width=0.1\textwidth]{pictures/scotch&soda/cup.jpg}
\includegraphics[width=0.1\textwidth]{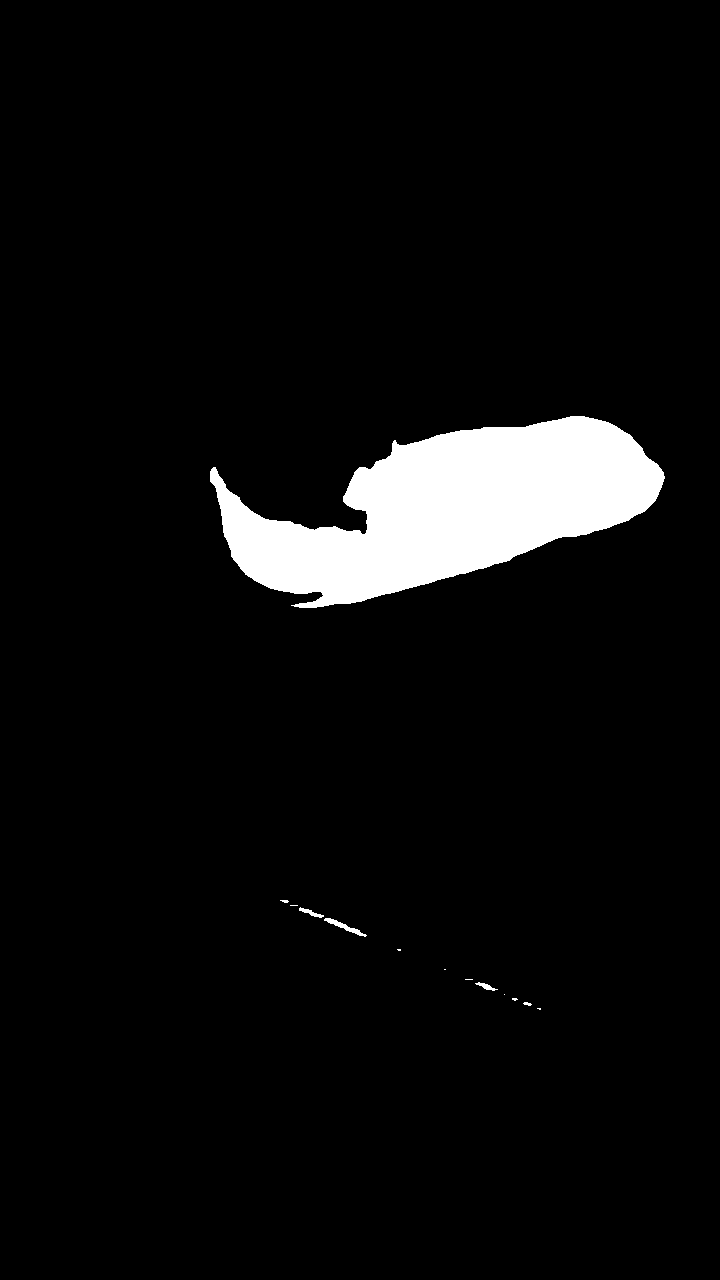}
\includegraphics[width=0.1\textwidth]{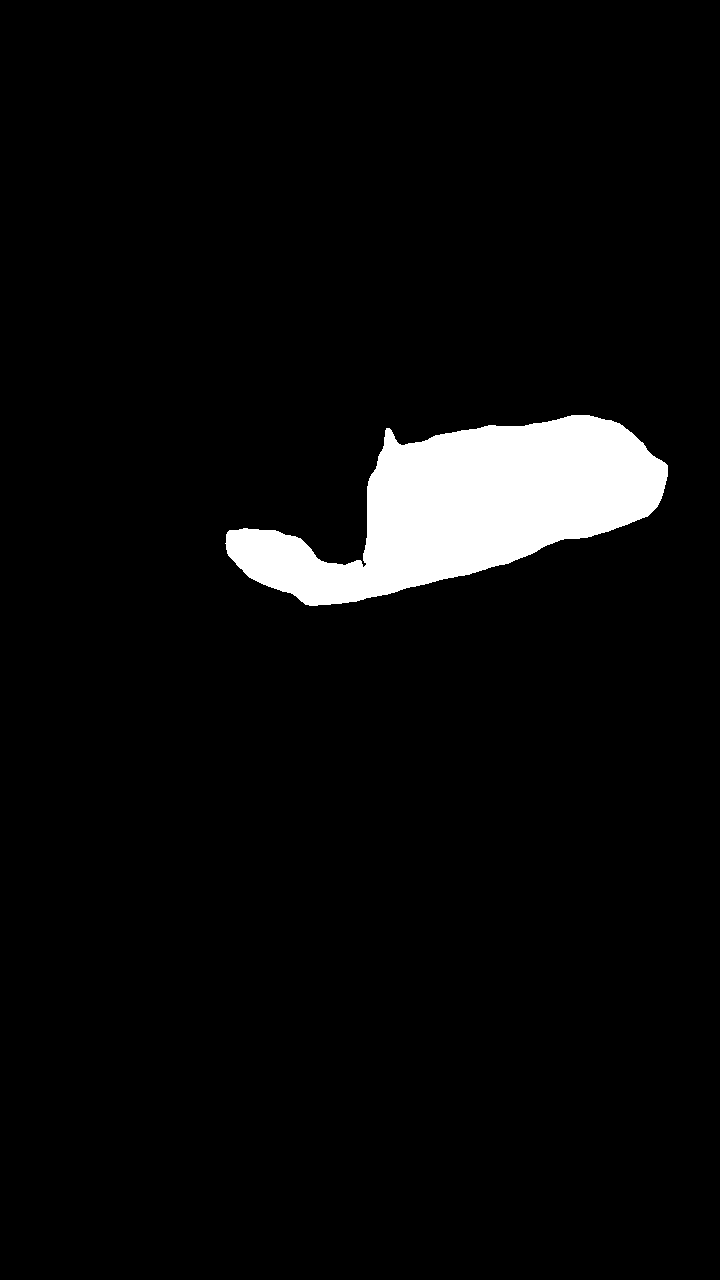}
\includegraphics[width=0.1\textwidth]{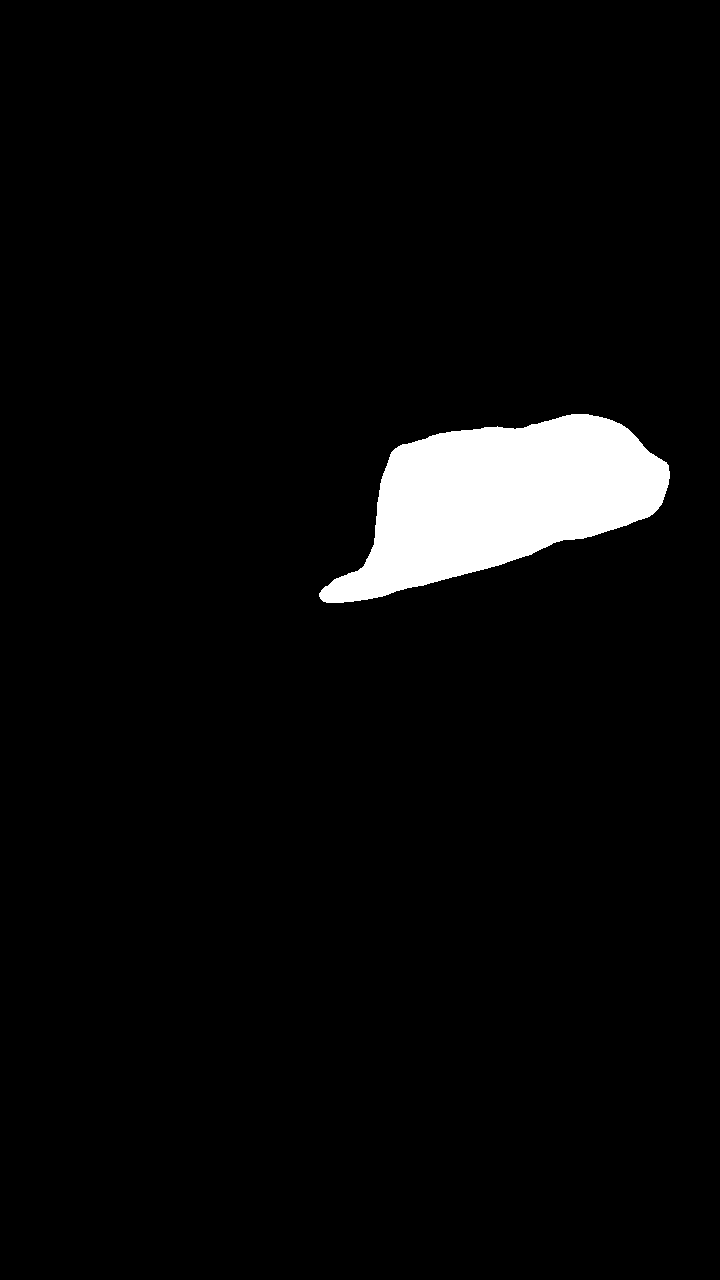}
\includegraphics[width=0.1\textwidth]{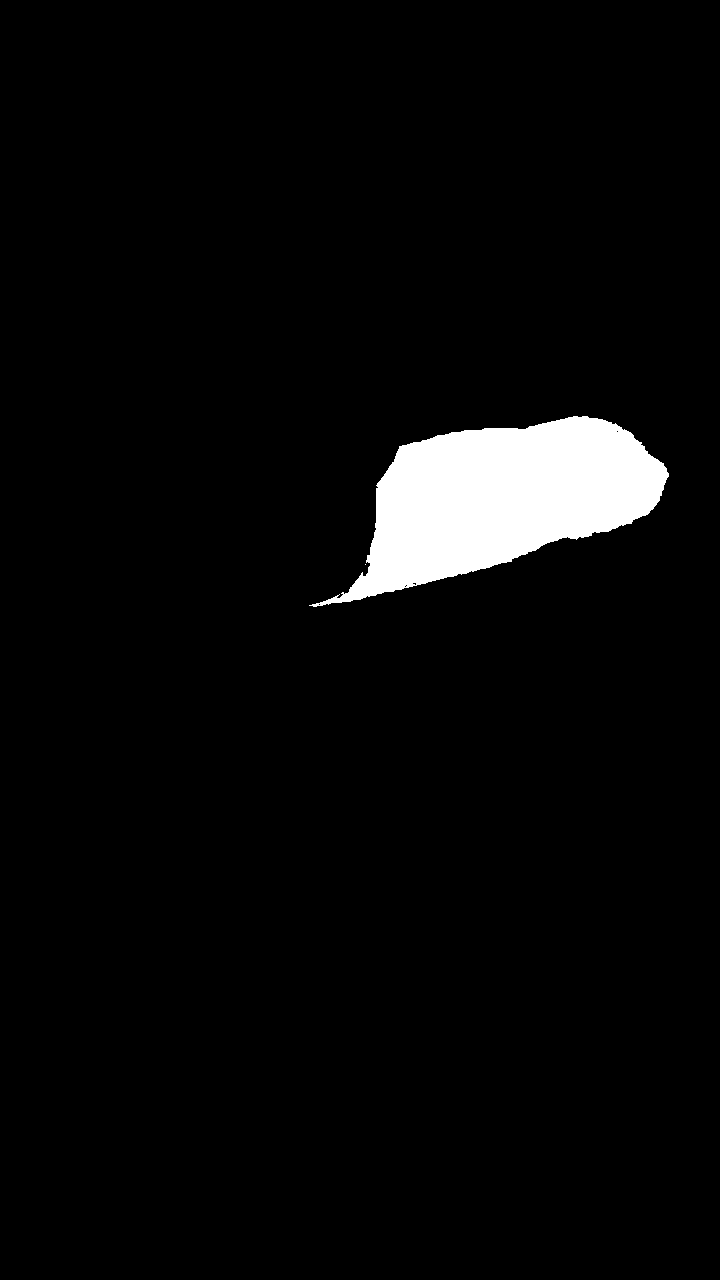}

\includegraphics[width=0.1\textwidth]{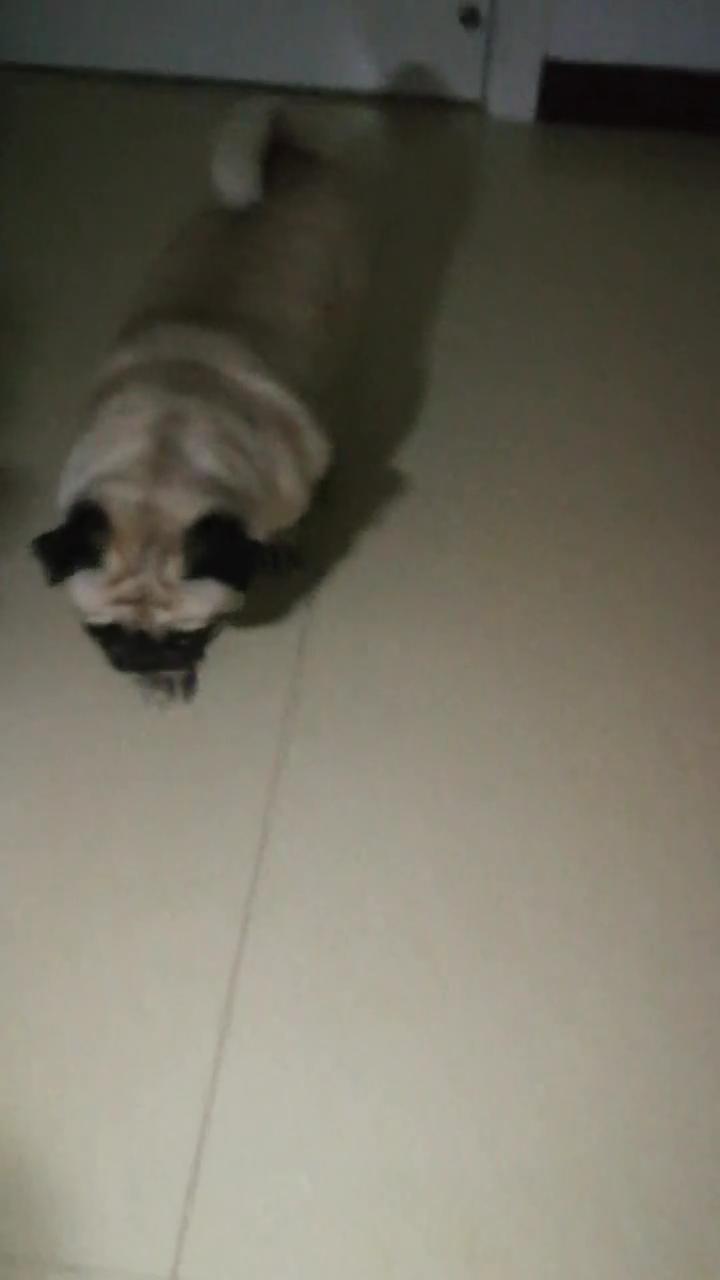}
\includegraphics[width=0.1\textwidth]{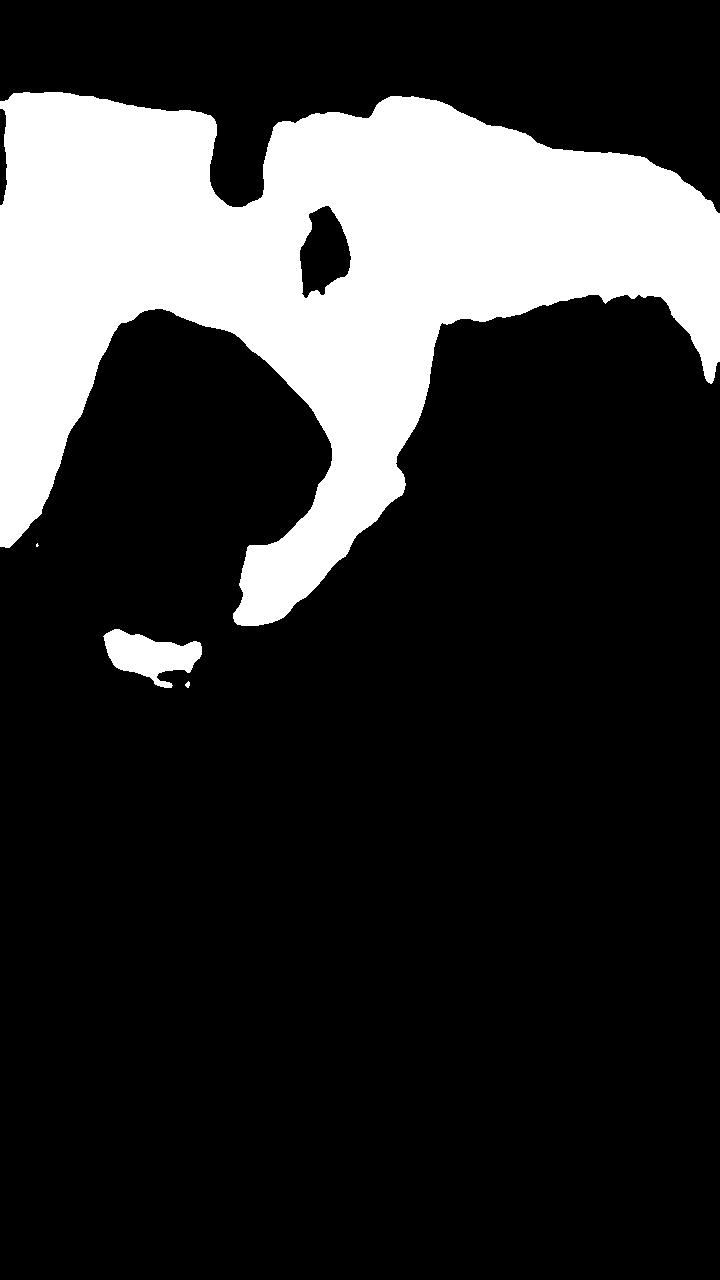}
\includegraphics[width=0.1\textwidth]{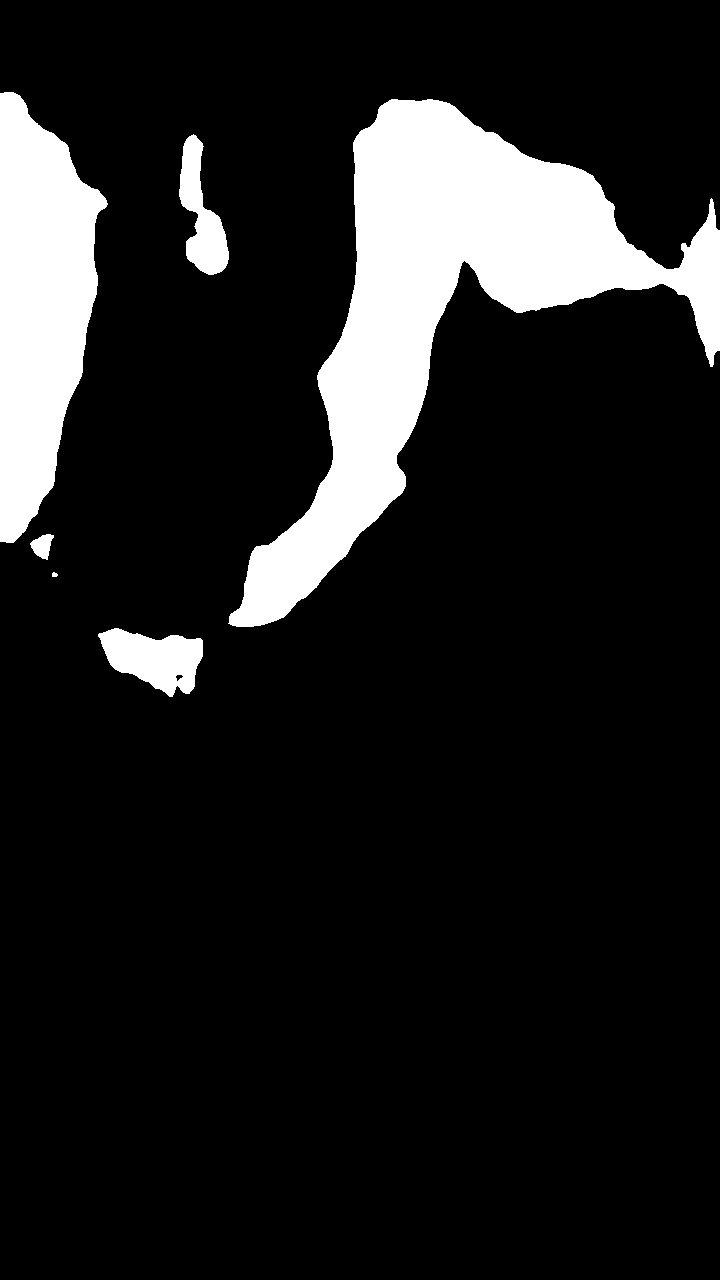}
\includegraphics[width=0.1\textwidth]{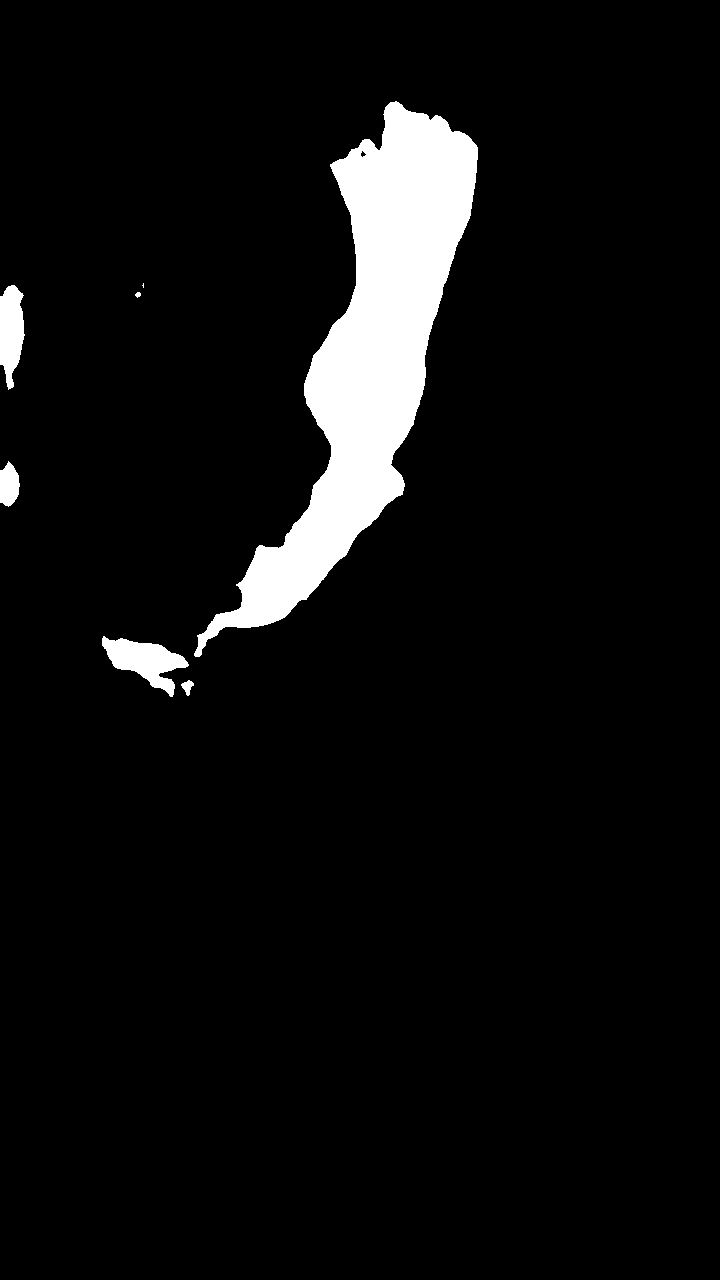}
\includegraphics[width=0.1\textwidth]{pictures/scotch&soda/dog2.jpg}
\includegraphics[width=0.1\textwidth]{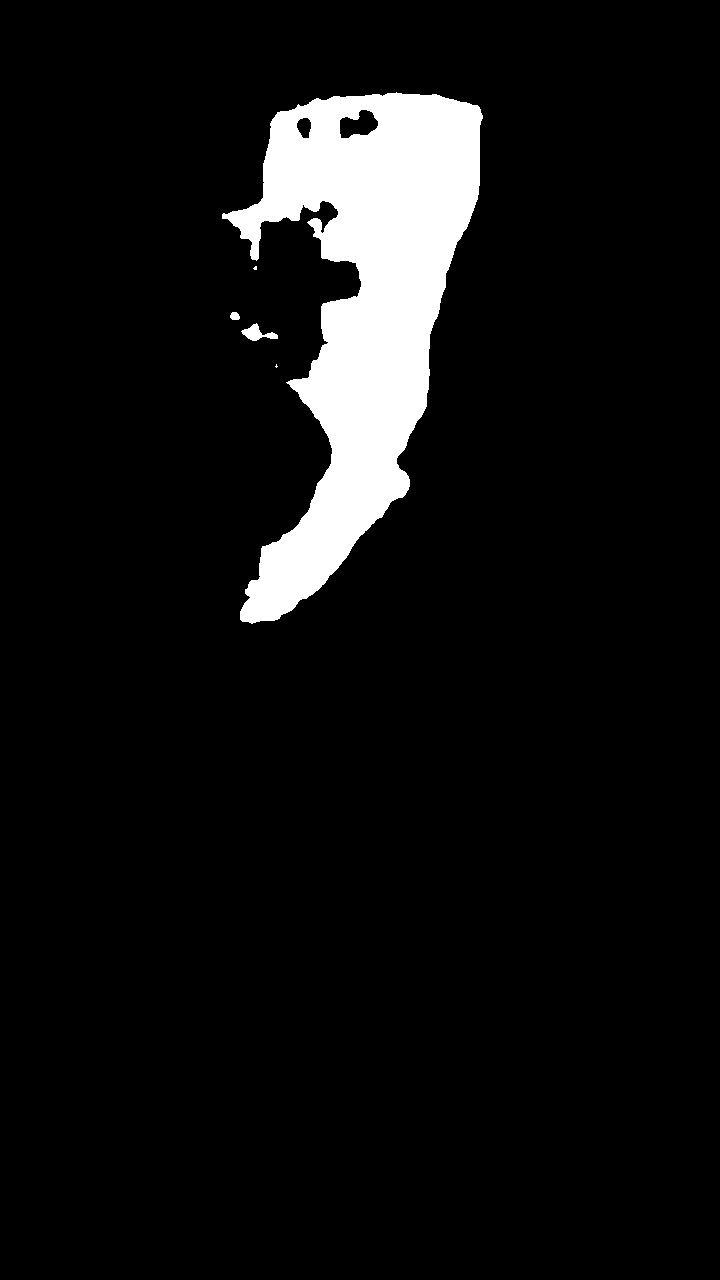}
\includegraphics[width=0.1\textwidth]{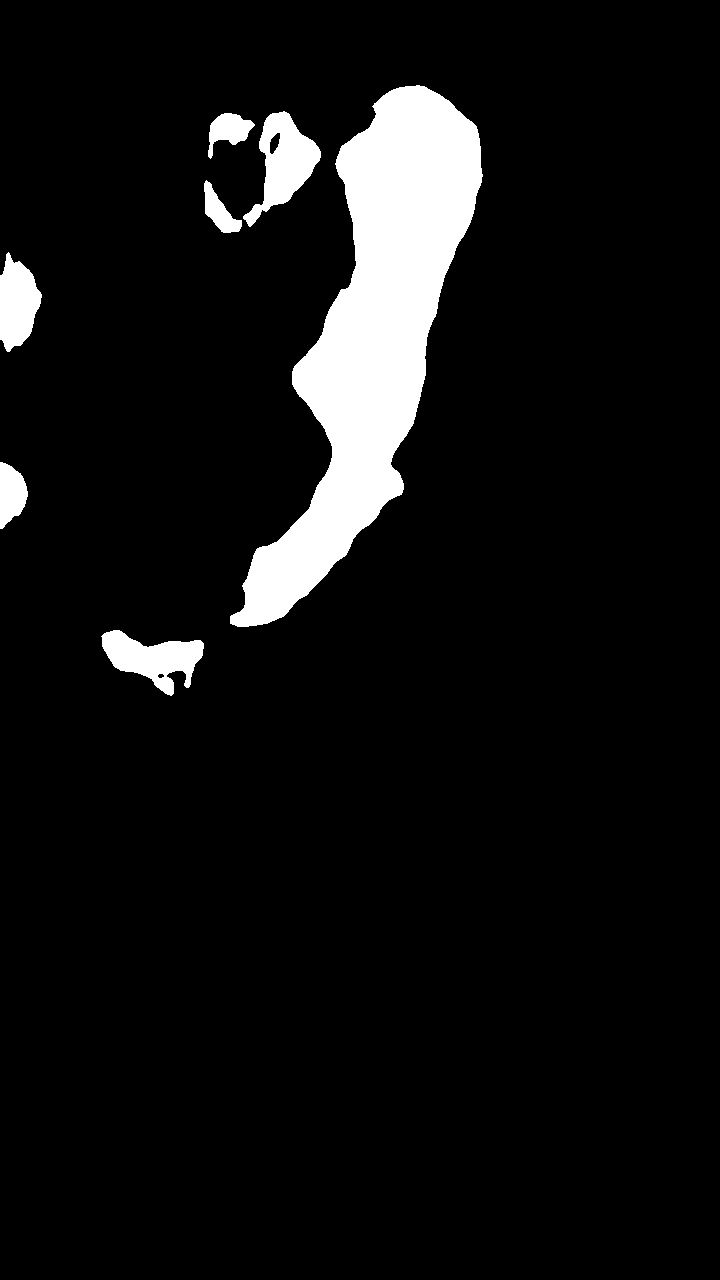}
\includegraphics[width=0.1\textwidth]{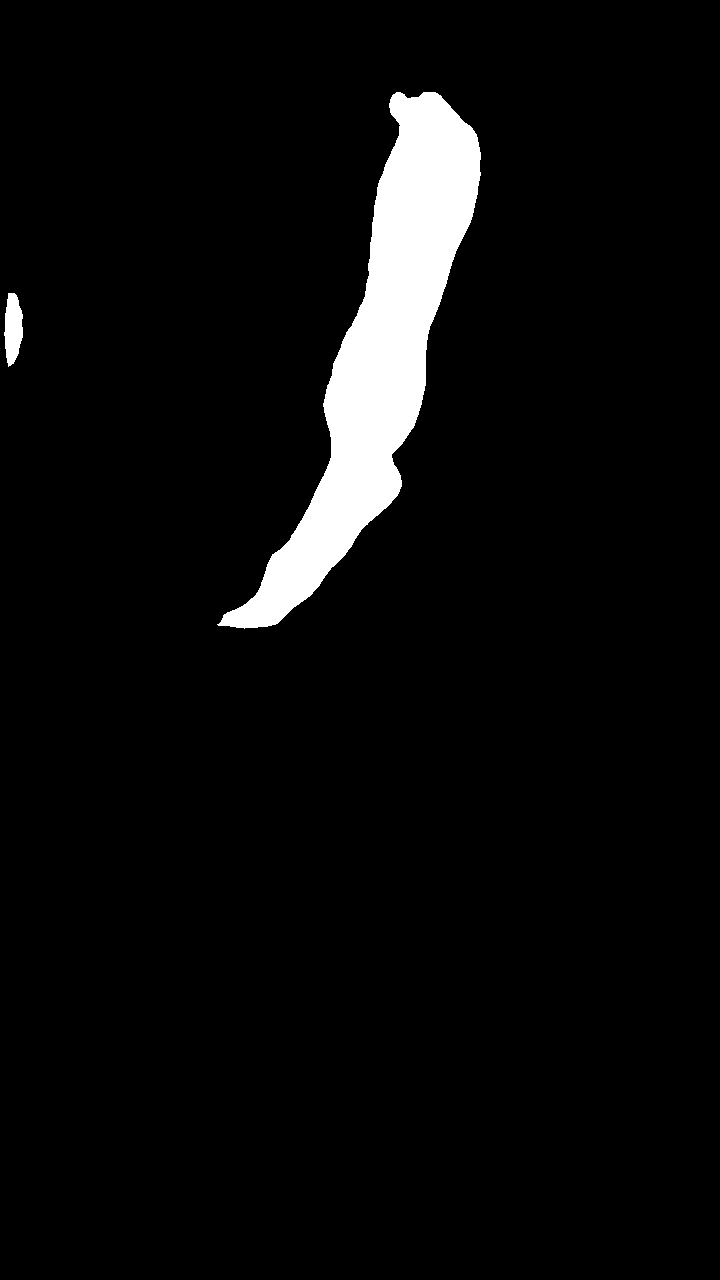}
\includegraphics[width=0.1\textwidth]{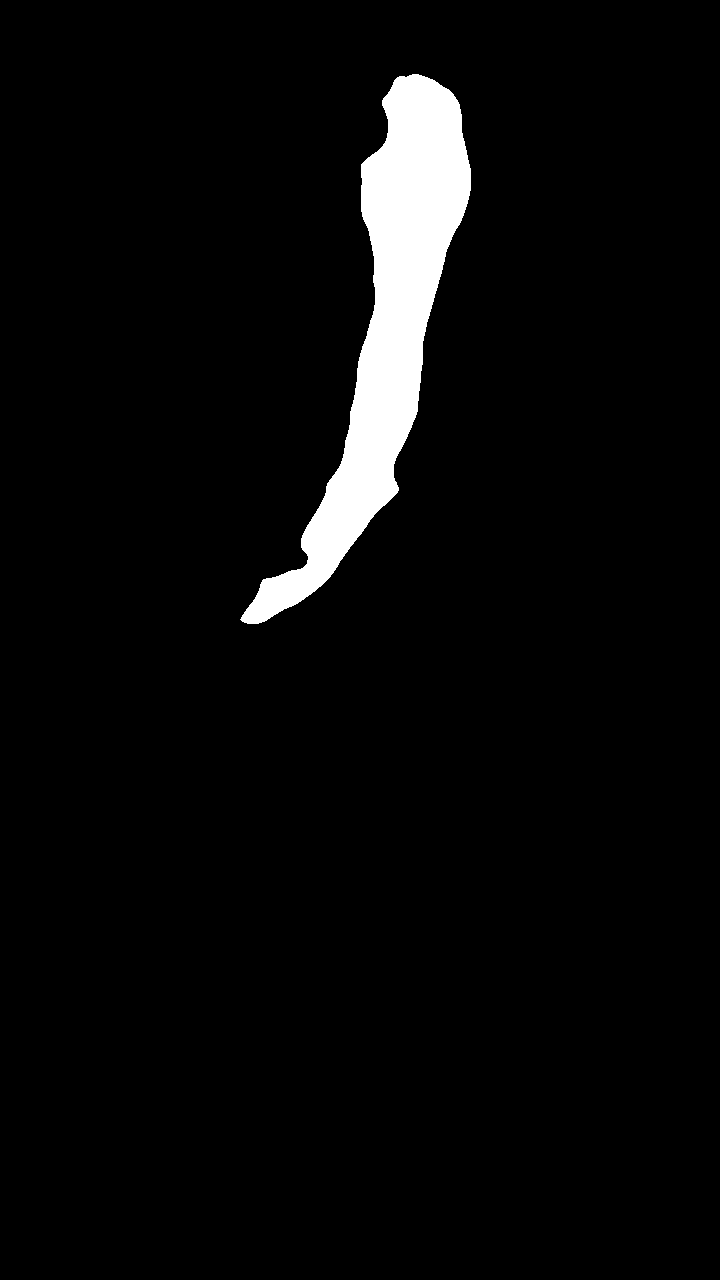}

\includegraphics[width=0.1\textwidth]{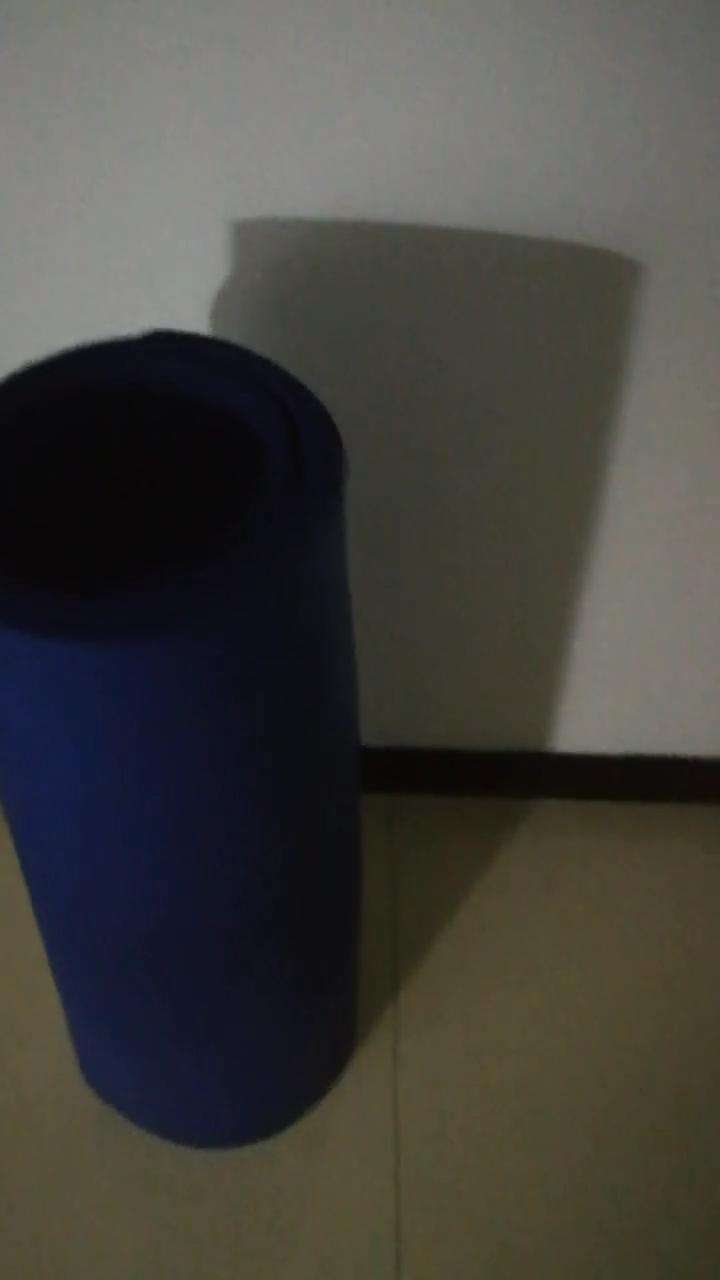}
\includegraphics[width=0.1\textwidth]{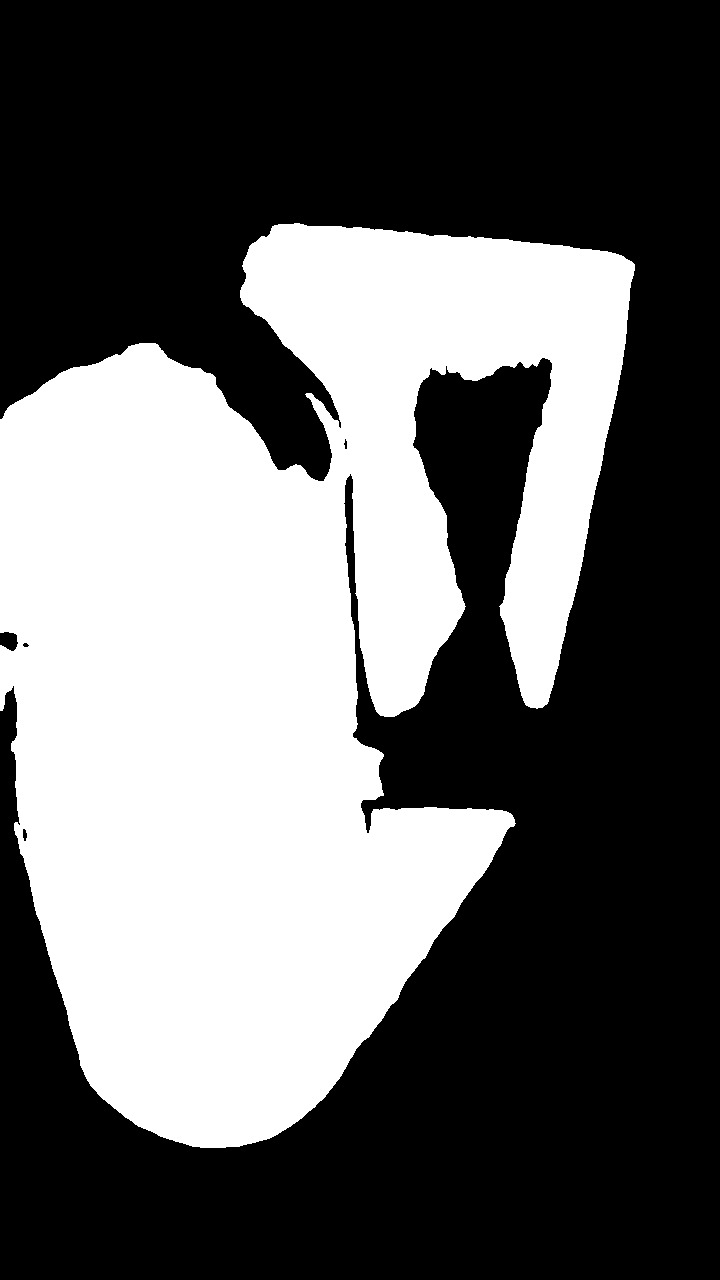}
\includegraphics[width=0.1\textwidth]{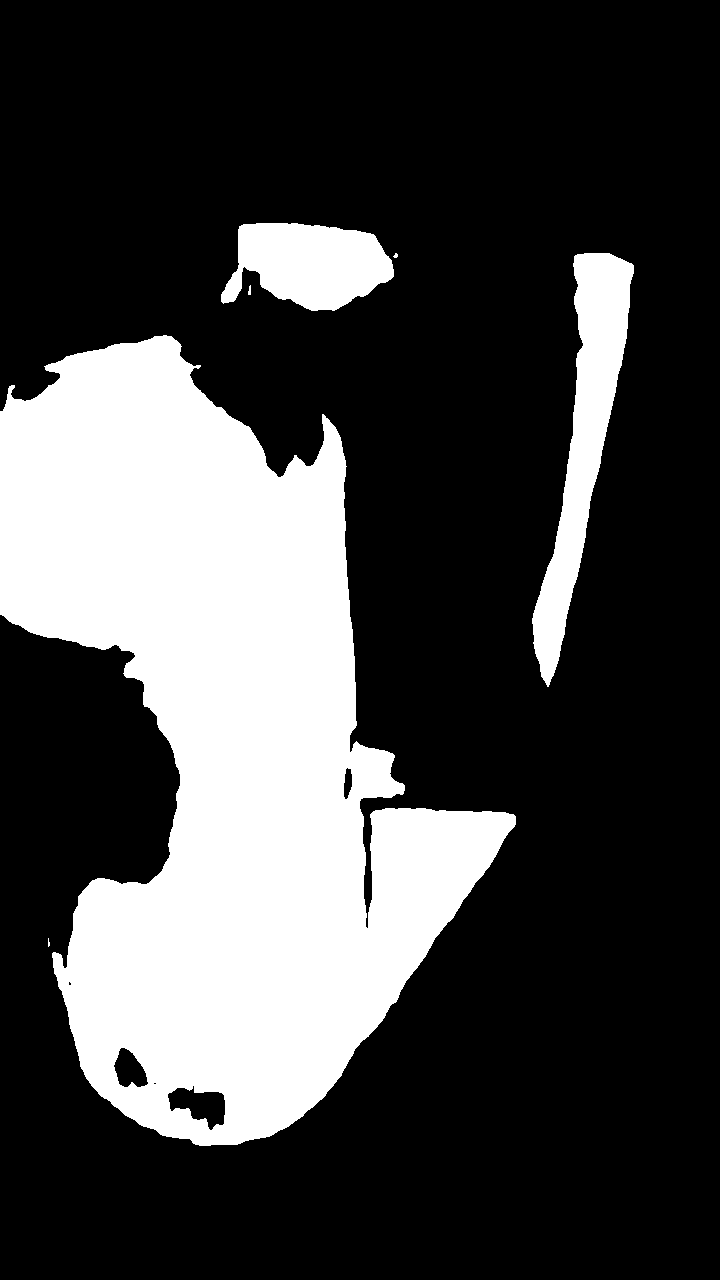}
\includegraphics[width=0.1\textwidth]{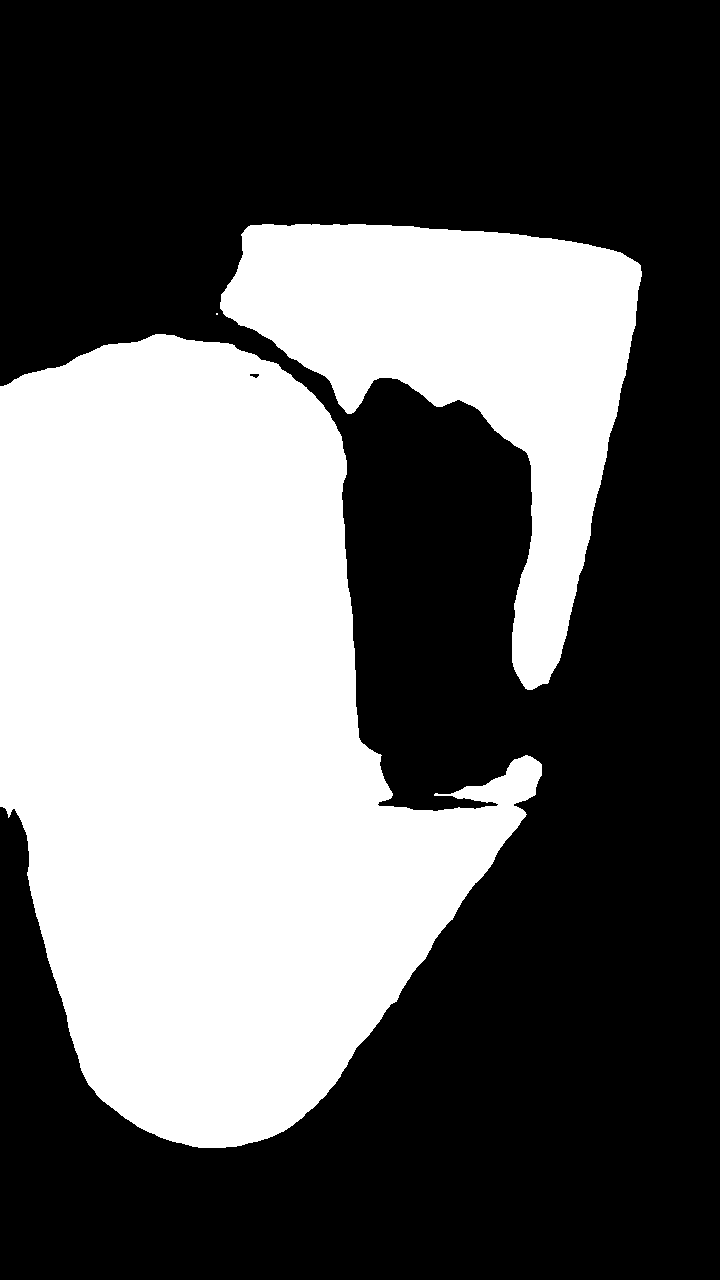}
\includegraphics[width=0.1\textwidth]{pictures/scotch&soda/cushion.jpg}
\includegraphics[width=0.1\textwidth]{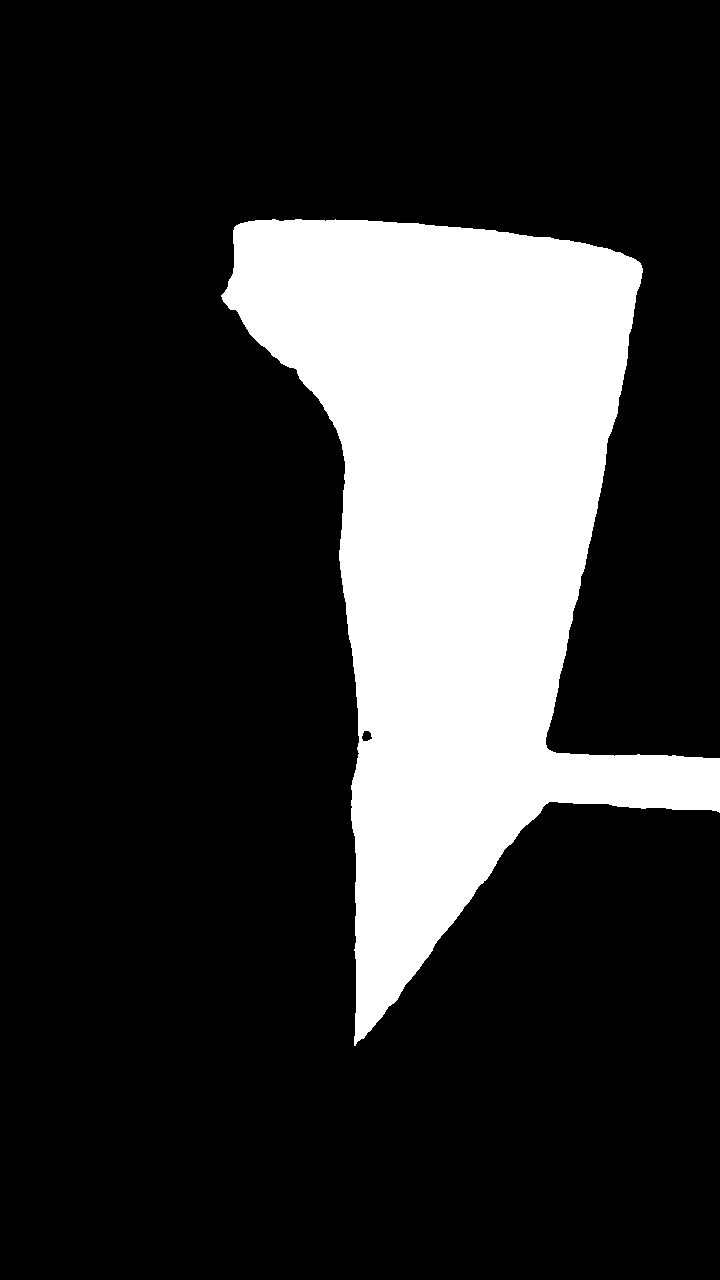}
\includegraphics[width=0.1\textwidth]{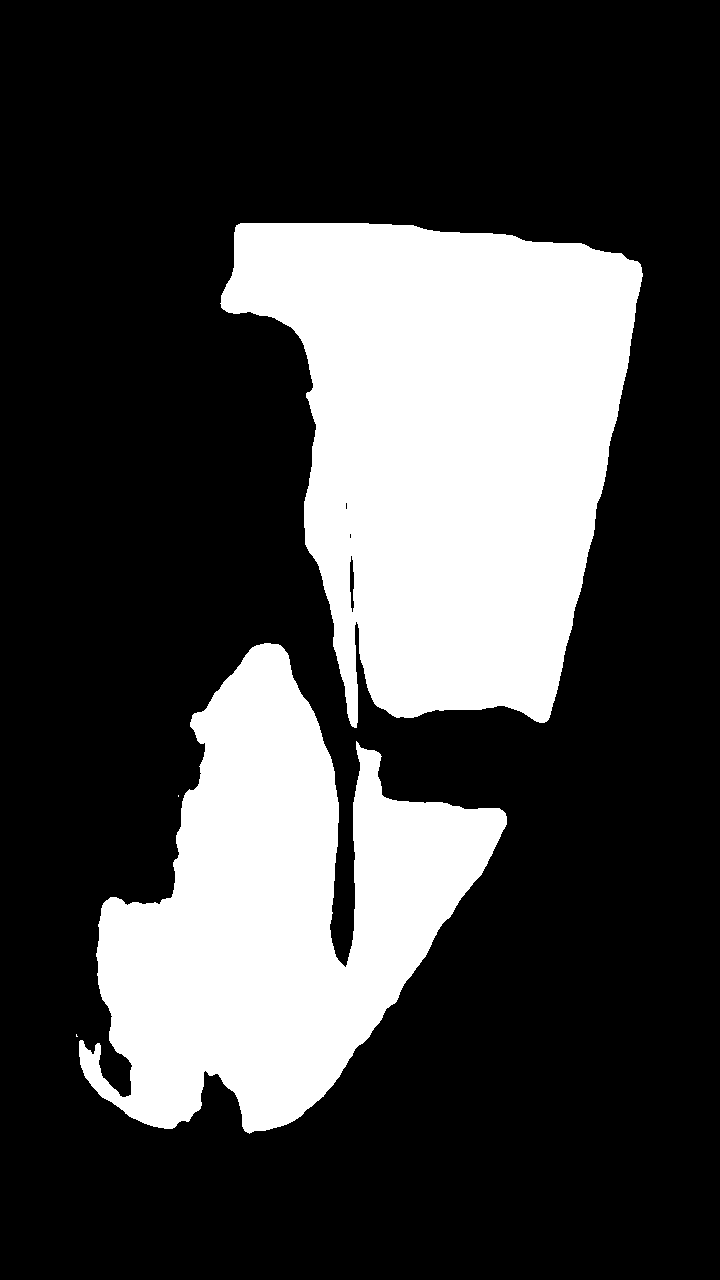}
\includegraphics[width=0.1\textwidth]{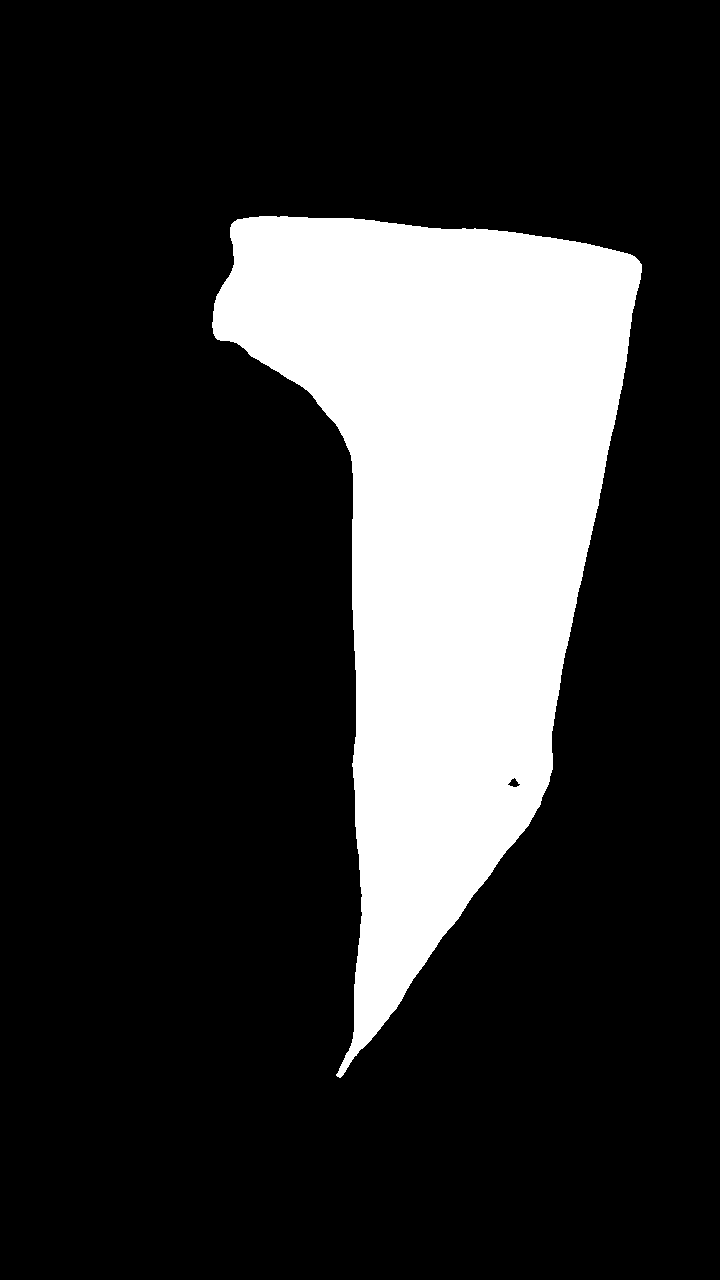}
\includegraphics[width=0.1\textwidth]{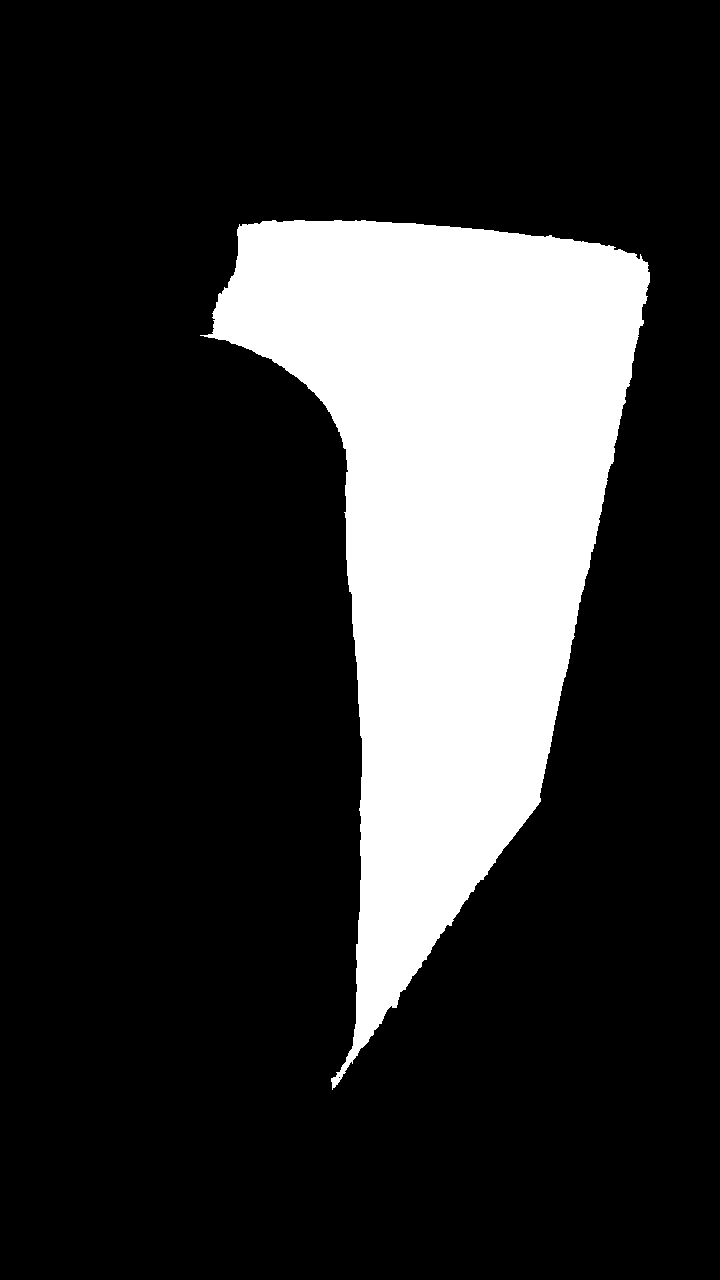}

\includegraphics[width=0.1\textwidth]{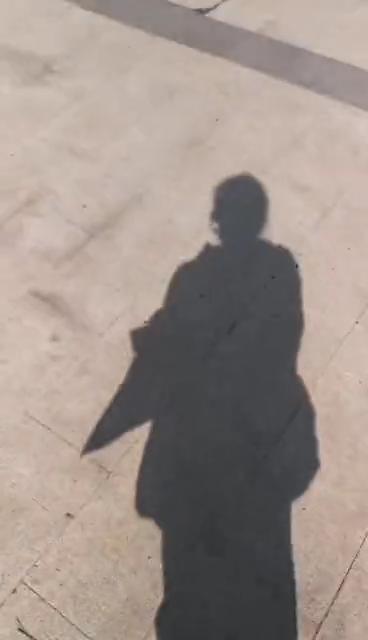}
\includegraphics[width=0.1\textwidth]{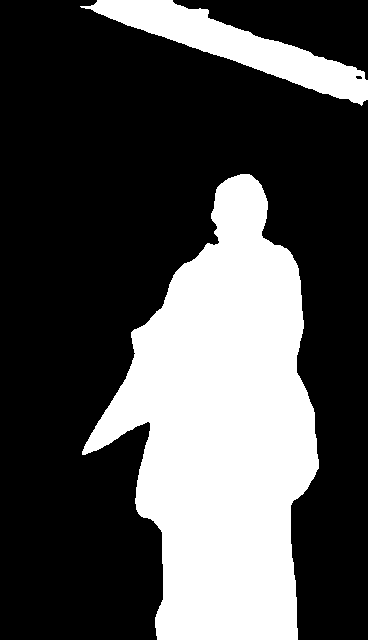}
\includegraphics[width=0.1\textwidth]{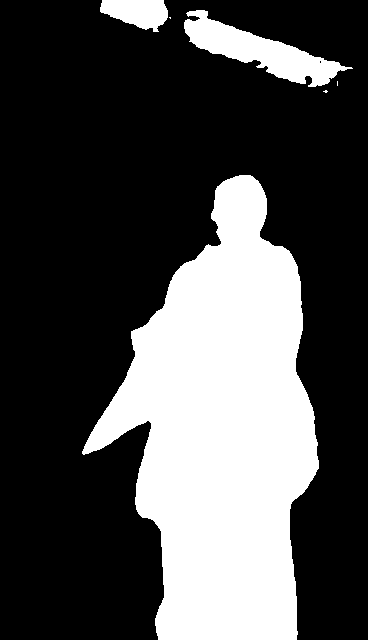}
\includegraphics[width=0.1\textwidth]{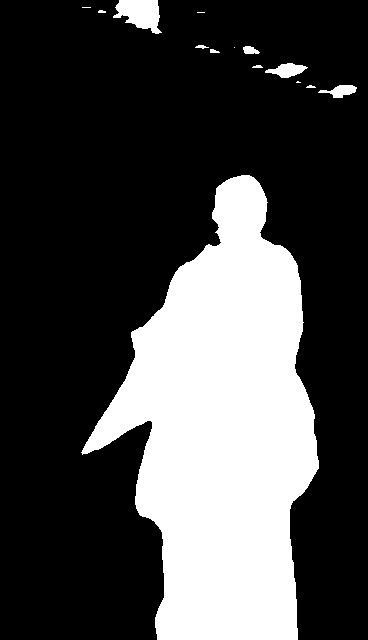}
\includegraphics[width=0.1\textwidth]{pictures/scotch&soda/walk4.jpg}
\includegraphics[width=0.1\textwidth]{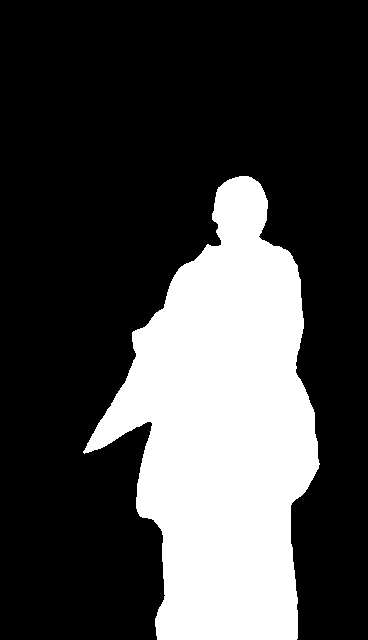}
\includegraphics[width=0.1\textwidth]{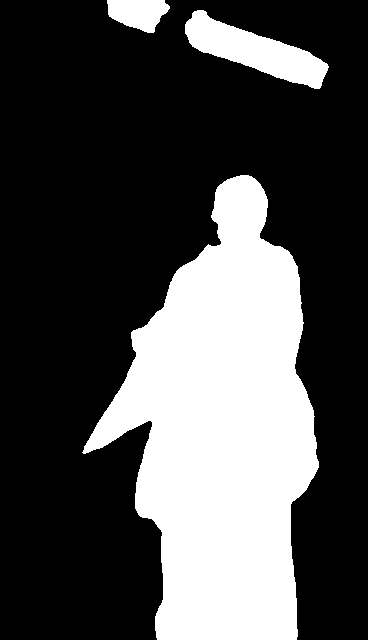}
\includegraphics[width=0.1\textwidth]{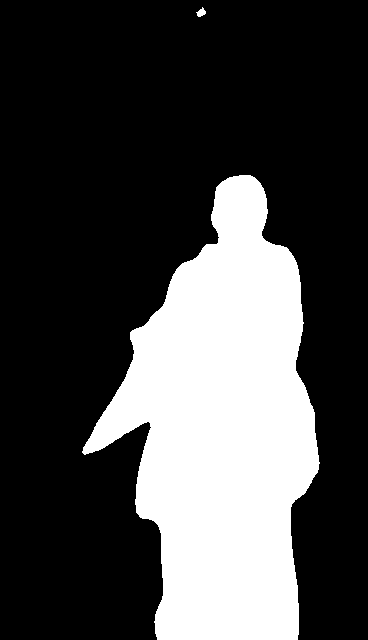}
\includegraphics[width=0.1\textwidth]{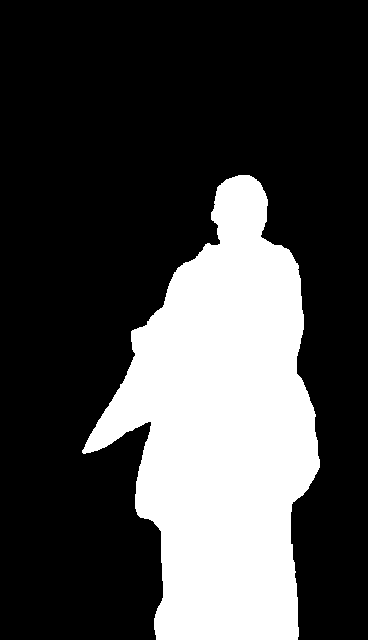}

\includegraphics[width=0.1\textwidth]{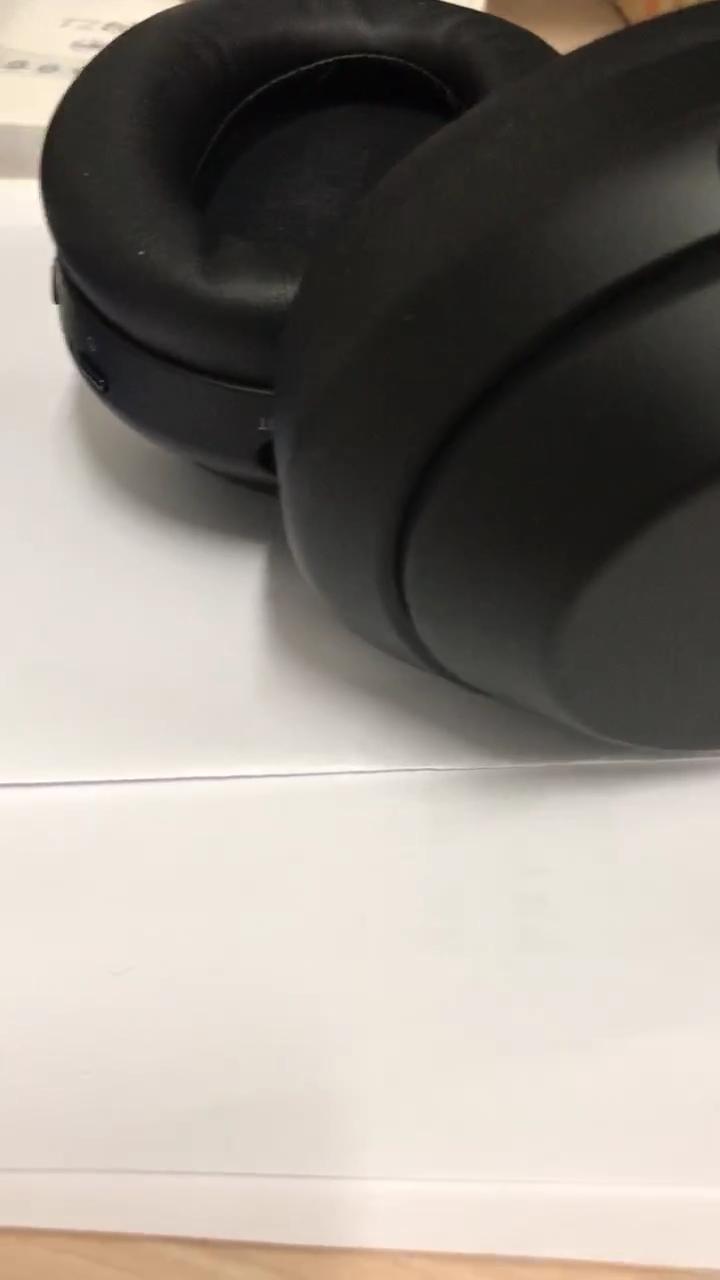}
\includegraphics[width=0.1\textwidth]{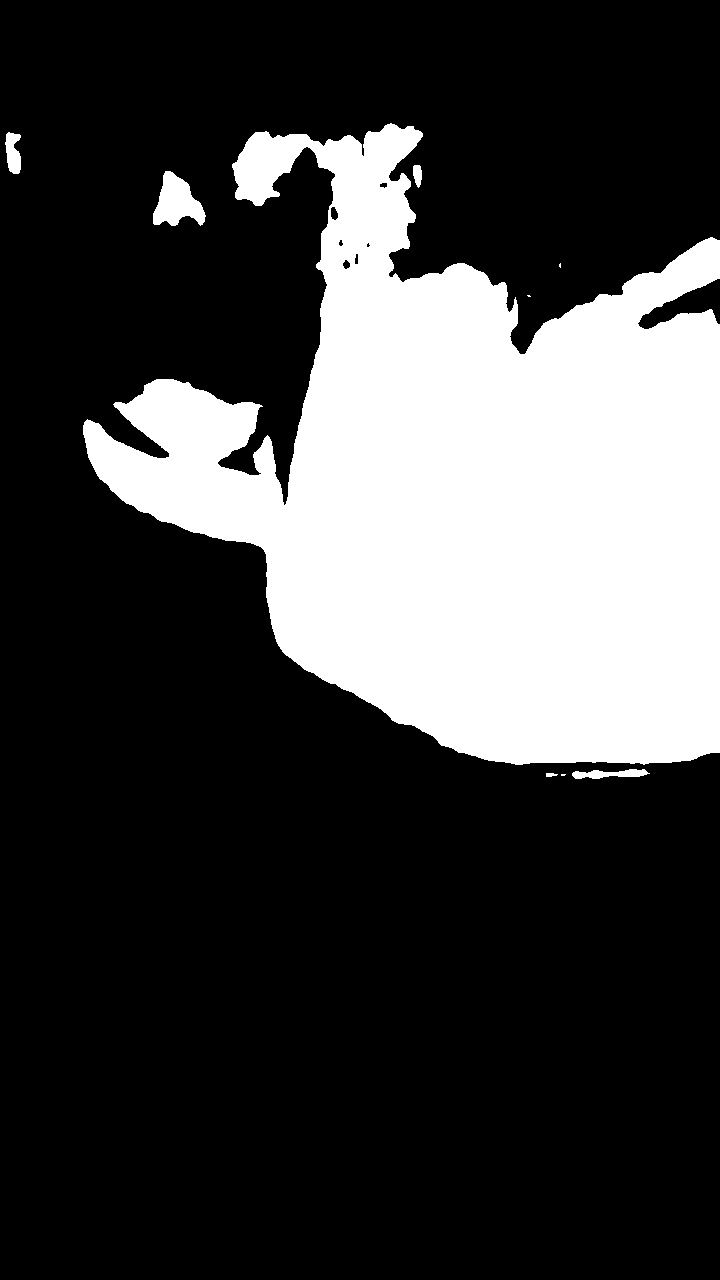}
\includegraphics[width=0.1\textwidth]{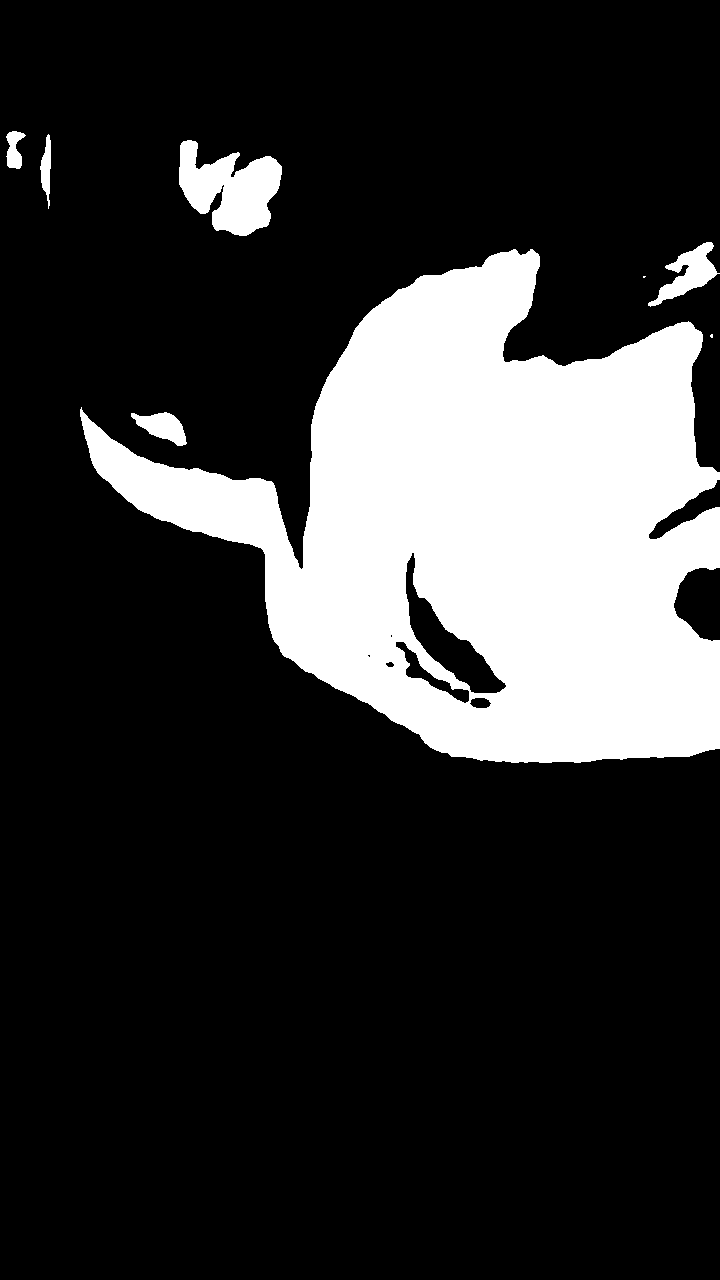}
\includegraphics[width=0.1\textwidth]{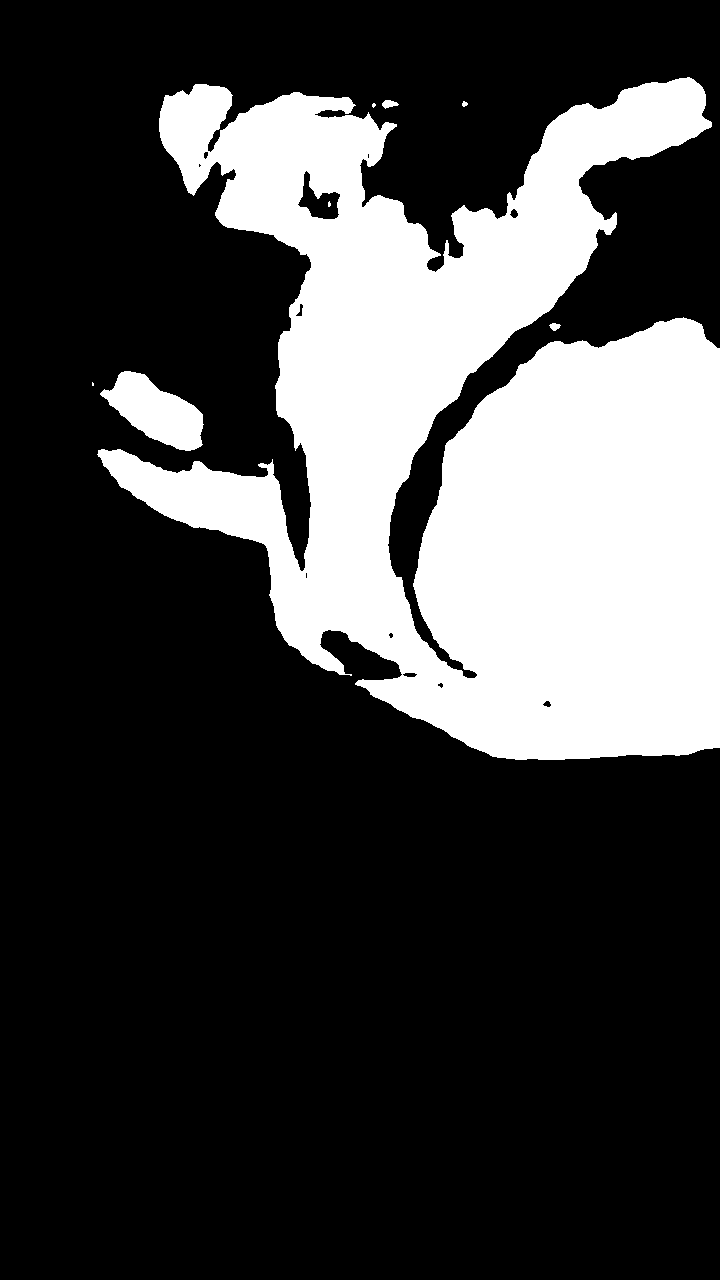}
\includegraphics[width=0.1\textwidth]{pictures/scotch&soda/table.jpg}
\includegraphics[width=0.1\textwidth]{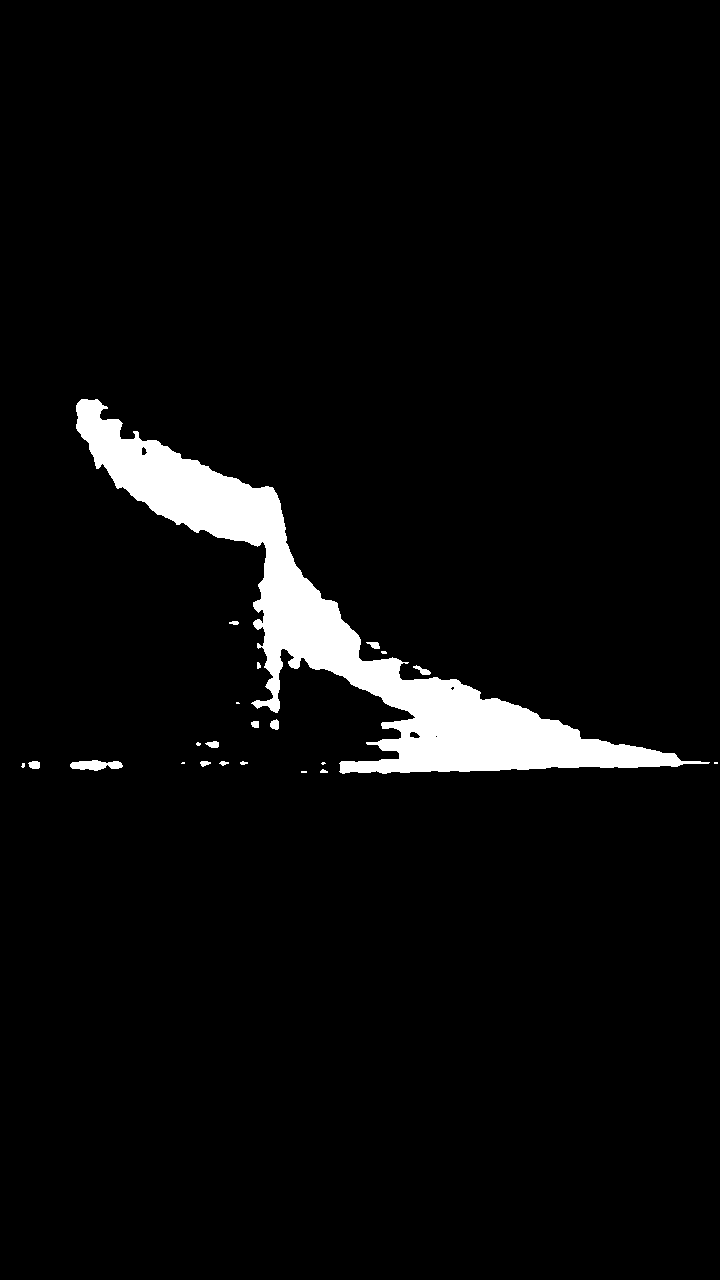}
\includegraphics[width=0.1\textwidth]{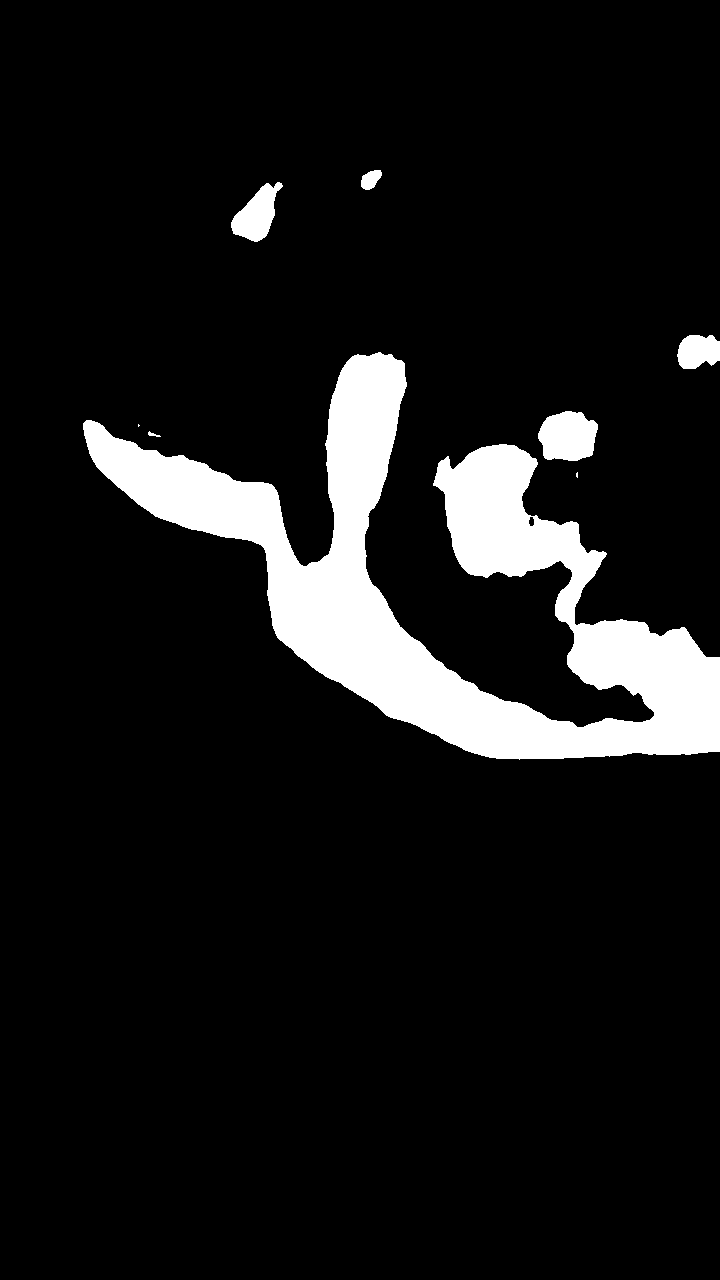}
\includegraphics[width=0.1\textwidth]{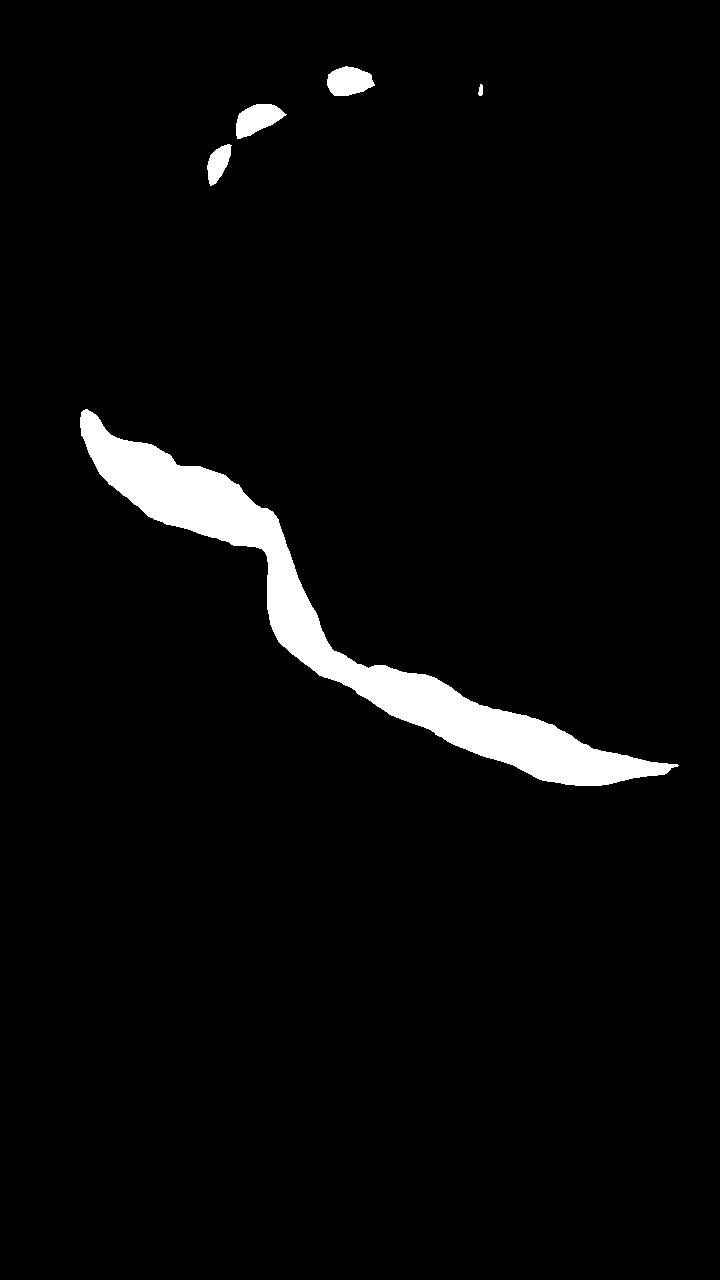}
\includegraphics[width=0.1\textwidth]{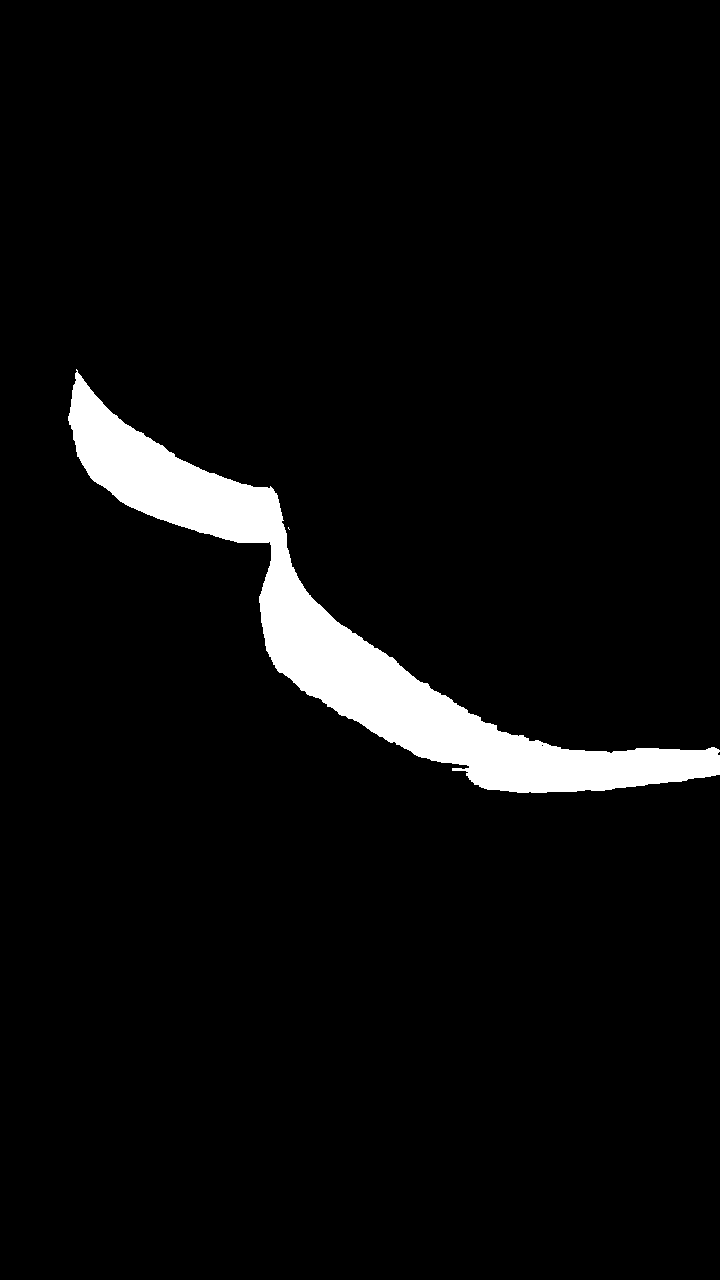}

\begin{minipage}{0.1\textwidth}
\centering
\footnotesize(a) Input
\end{minipage}
\begin{minipage}{0.1\textwidth}
\centering
\footnotesize(b) DDP
\end{minipage}
\begin{minipage}{0.1\textwidth}
\centering
\footnotesize(c) Pix2Seq
\end{minipage}
\begin{minipage}{0.1\textwidth}
\centering
\footnotesize(d) SILT
\end{minipage}
\begin{minipage}{0.1\textwidth}
\centering
\footnotesize(e) SCOTCH \& SODA
\end{minipage}
\begin{minipage}{0.1\textwidth}
\centering
\footnotesize(f) DAS
\end{minipage}
\begin{minipage}{0.1\textwidth}
\centering
\footnotesize(g) TBGDiff
\end{minipage}
\begin{minipage}{0.1\textwidth}
\centering
\footnotesize(h) Ours
\end{minipage}
\begin{minipage}{0.1\textwidth}
\centering
\footnotesize(i) GT
\end{minipage}
\vspace{-3mm}
\caption{Qualitative comparison results of state-of-the-art methods (DDP~\cite{ji2023ddp}, Pix2Seq~\cite{chen2023generalist}, SILT~\cite{yang2023silt}, SCOTCH \& SODA~\cite{liu2023scotch}, DAS~\cite{wang2023detect}, and TBGDiff~\cite{zhou2024timeline}). (b-d) are the recent methods in IOS, ISD, and VOS, and (e-g) are the latest networks in VSD.}
\label{fig:res2}
\vspace{-3mm}
\end{figure*}

Fig.~\ref{fig:res2} provides additional shadow segmentation masks of our method and comparative approaches on different images of ViSha~\cite{chen2021triple}. For images containing black objects, our method can still accurately detect shadow pixels despite interference from black objects. For instance, DTTNet accurately identifies shadow regions in cases involving the gray patterns on the bottle in the second row, the dog's nose in the third row, the dark cup in the fourth row, the gray streaks on the ground in the fifth row, and the black object in the last row. Furthermore, we present results of our method on the CVSD~\cite{duan2024two} dataset in Fig.~\ref{fig:longterm1} to demonstrate its detection performance in complex scenarios.

\section{Core Algorithm Implementation}
For the detailed core source code of dataset preprocessing
and DTTNet, please refer to the accompanying python files. Please follow the outlined in the README.md.

\section{Video of the Results}
Our method, DTTNet, demonstrates robust performance on a variety of challenging video shadow detection sequences. The result video showcases DTTNet’s ability to accurately distinguish dynamic shadows from complex backgrounds under varying lighting conditions and motion patterns. It highlights both qualitative improvements and temporal consistency across frames, especially in challenging scenes with occlusion, fast motion, and ambiguous dark regions. Please refer to the MP4 video included in the supplementary materials.

\bibliography{aaai2026}